\documentclass[conference]{IEEEtran}
\usepackage{times}
\usepackage{titletoc}
\usepackage[numbers]{natbib}
\usepackage{multicol}
\usepackage[bookmarks=true]{hyperref}
\usepackage{url}            
\usepackage[usenames,dvipsnames]{xcolor}
\usepackage{graphicx}
\usepackage{subcaption}
\usepackage{wrapfig}        
\usepackage{booktabs}       
\usepackage{amsfonts}       
\usepackage{microtype}      
\usepackage{amsmath,amssymb,bbm}
\usepackage{siunitx}        
\usepackage[capitalize,noabbrev]{cleveref}  
\usepackage{soul}

\crefformat{footnote}{#2\textsuperscript{\normalfont #1}#3}  

\usepackage{fontawesome}
\usepackage{multirow}

\definecolor{forestgreen}{rgb}{0.13, 0.55, 0.13}
\definecolor{indiagreen}{rgb}{0.07, 0.53, 0.03}

\hypersetup{
    colorlinks=true,
    linkcolor=black,
    urlcolor=brown,
    citecolor=black,
}

\begin{document}

\title{HumanoidBench: Simulated Humanoid Benchmark for Whole-Body Locomotion and Manipulation}

\author{
  Carmelo Sferrazza$^1$ \quad Dun-Ming Huang$^1$ \quad Xingyu Lin$^1$ \quad Youngwoon Lee$^{1,2}$ \quad Pieter Abbeel$^1$ \\
  $^1$UC Berkeley \qquad $^2$Yonsei University 
}


\makeatletter
\let\@oldmaketitle\@maketitle
\renewcommand{\@maketitle}{\@oldmaketitle
    \centering
    \vspace{1em}
    \includegraphics[width=1.0\linewidth,height=0.25\linewidth]{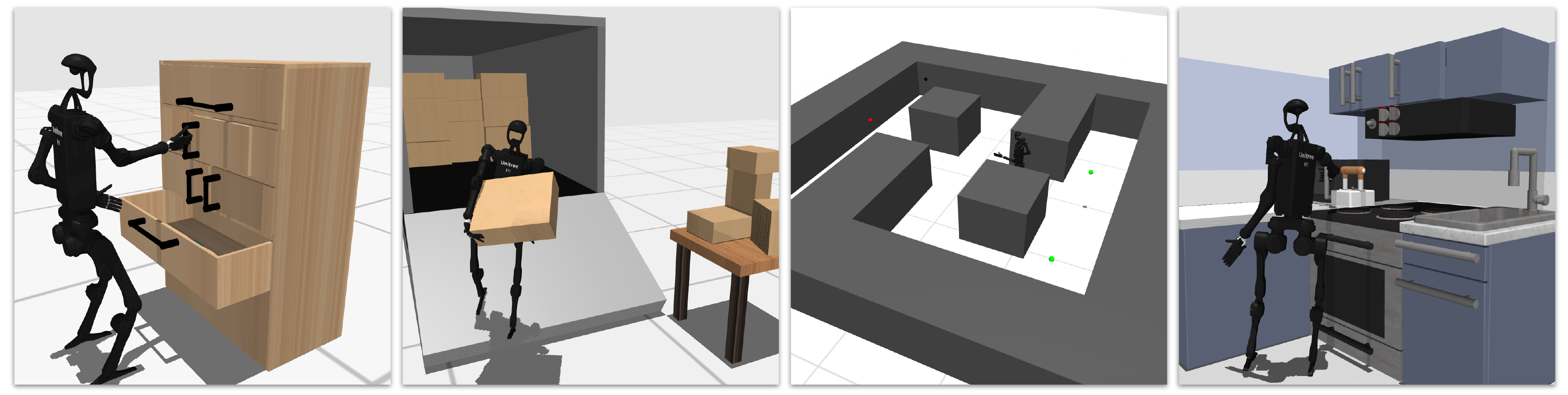}
    \captionof{figure}{
        Humanoid robots equipped with dexterous hands hold immense promise for integration into real-world human environments. 
        Nonetheless, harnessing the full potential of humanoid robots presents numerous challenges, such as the intricate control of robots with complex dynamics, sophisticated coordination among various body parts, and addressing long-horizon complex tasks envisioned for these robots. We present \textbf{HumanoidBench}, a simulated humanoid robot benchmark consisting of $15$ whole-body manipulation and $12$ locomotion tasks, such as shelf rearrangement, package unloading, and maze navigation.
    }
    \label{fig:teaser}
    \smallskip}
\makeatother

\maketitle
\addtocounter{figure}{-1}

\begin{abstract}
Humanoid robots hold great promise in assisting humans in diverse environments and tasks, due to their flexibility and adaptability leveraging human-like morphology. However, research in humanoid robots is often bottlenecked by the costly and fragile hardware setups. To accelerate algorithmic research in humanoid robots, we present a high-dimensional, simulated robot learning benchmark, HumanoidBench, featuring a humanoid robot equipped with dexterous hands and a variety of challenging whole-body manipulation and locomotion tasks. Our findings reveal that state-of-the-art reinforcement learning algorithms struggle with most tasks, whereas a hierarchical learning approach achieves superior performance when supported by robust low-level policies, such as walking or reaching. With HumanoidBench, we provide the robotics community with a platform to identify the challenges arising when solving diverse tasks with humanoid robots, facilitating prompt verification of algorithms and ideas. 
The open-source code is available at \linebreak \url{https://humanoid-bench.github.io}.
\end{abstract}

\IEEEpeerreviewmaketitle

\begin{table*}[t]
    \centering
    \begin{tabular}{lcccccc}
        \toprule
        Benchmark                                      & Dexterous hands & Action dim. & DoF & Task horizon & \# Tasks & Skills\footnotemark[1] \\
        \midrule
        MyoHand~\citep{caggiano2022myosuite}           & \textcolor{indiagreen}\faCheck    & $39$    & $23$D  & $50$-$2000$      & $9$      & PnP, R, Po, IR, H, Ro   \\
        Adroit~\citep{plappert2018robotics}            & \textcolor{indiagreen}\faCheck    & $24$     & $24$D  & $200$      & $4$      & PnP, P, R, Po, IR, H, L, Ro   \\
        MyoLeg~\citep{caggiano2022myosuite}                        & \textcolor{red}\faTimes            & $80$    & $20$D  & $1000$      & $1$      & Lo, St   \\
        LocoMujoco~\citep{alhafez2023locomujoco} (Unitree-H1)       & \textcolor{red}\faTimes            & $19$    & $6$D  & $100$-$500$      & $27$      & L, Lo, St, BM   \\
        DMControl~\citep{tassa2018deepmind} (Humanoid)                     & \textcolor{red}\faTimes            & $24$-$56$    & $22$D  & $1000$      & $6$      & Lo, St   \\
        FurnitureSim~\citep{heo2023furniturebench}     & \textcolor{red}\faTimes           & $8$     & $6$D   & $2300$   & $8$      & PnP, P, I, IR, H, L, Ro   \\
        robosuite~\citep{zhu2020robosuite}             & \textcolor{red}\faTimes           & $6$-$24$     & $6$-$7$D   & $500$      & $9$      & PnP, P, I, R, IR, H, L, Ro   \\
        rlbench~\citep{james2019rlbench}               & \textcolor{red}\faTimes           & $6$-$7$     & $6$-$7$D   & $100$-$1000$      & $106$      & PnP, P, I, R, Po, IR, H, L, Ro   \\
        metaworld~\citep{yu2019meta}                   & \textcolor{red}\faTimes           & $6$     & $7$D   & $500$      & $50$     & PnP, P, I, R, Po, IR, H, L, Ro   \\
        \midrule
        \textbf{HumanoidBench (Ours)}                  & \textcolor{indiagreen}\faCheck    & $61$    & $75$D   & $500$-$1000$   & $27$     & PnP, P, I, R, Po, IR, H, L, Ro, Lo, BM, St  \\
        \bottomrule
    \end{tabular}
    \begin{flushleft}\scriptsize
        \textsuperscript{1}PnP: Pick-and-place / P: Push / I: Insert / R: Reach / Po: Pose / IR: In-hand re-orientation / H: Hold / L: Lift / Ro: Rotate / Lo: Locomotion / BM: Whole-body (humanoid) Manipulation / St: Stabilization
    \end{flushleft}
    \vspace{-0.5em}
    \caption{
        \textbf{Comparison of simulated robot benchmarks.} Our humanoid robot benchmark tests a variety of complex, long-horizon task with a large action space.
    }  
    \label{tab:benchmarks}
    \vspace{-1em}
\end{table*}

\section{Introduction}

Humanoid robots have long held promise to be seamlessly deployed in our daily lives. Despite the rapid progress in humanoid robots' hardware (e.g., Boston Dynamics Atlas, Tesla Optimus, Unitree H1), their controllers are fully or partially hand-designed for specific tasks, which requires significant engineering efforts for each new task and environment, and often demonstrates only limited whole-body control capabilities. 

In recent years, robot learning has shown steady progress in both robotic manipulation~\citep{chi2023diffusion, zhao2023learning, fu2024mobile} and locomotion~\citep{kumar2021rma, zhuang2023robot}. However, scaling learning algorithms to humanoid robots is still challenging and has been delayed mainly due to such robots' costly and unsafe real-world experimental setups.

To accelerate the progress of research for humanoid robots, we present the first-of-its-kind humanoid robot benchmark, \textbf{HumanoidBench}, with a diverse set of locomotion and manipulation tasks, providing an accessible, fast, safe, and inexpensive testbed to robot learning researchers. Our simulated humanoid benchmark demonstrates a variety of challenges in learning for autonomous humanoid robots, such as the intricate control of robots with complex dynamics, sophisticated coordination among various body parts, and long-horizon complex tasks.

HumanoidBench provides (1) a simulation environment comprising a humanoid robot with two dexterous hands, as illustrated in \Cref{fig:teaser}; (2) a variety of tasks, spanning locomotion, manipulation, and whole-body control, incorporating humans' everyday tasks; (3) a standardized benchmark to evaluate the progress of the community on high-dimensional humanoid robot learning and control. In fact, HumanoidBench supports generic controller structures, including both learning and model-based approaches \citep{feng2014optimization, kuindersma2016optimization}. In this paper, we present extensive benchmarking results of the state-of-the-art reinforcement learning (RL) algorithms, which do not require extensive domain knowledge, and a hierarchical RL approach.

The simulation environment of HumanoidBench uses the MuJoCo~\citep{todorov2012mujoco} physics engine. For the simulated humanoid robot, we mainly opt for a Unitree H1 humanoid robot\footnote{\label{footnote:h1}\url{https://www.unitree.com/h1}}, which is relatively affordable and offers accurate simulation models~\citep{menagerie2022github}, with two dexterous Shadow Hands\footnote{\label{footnote:shadowhand}\url{https://www.shadowrobot.com/dexterous-hand-series/}} attached to its arms. Our environment can easily incorporate any humanoid robots and end effectors; thus, we provide other models, including Unitree G1\footnote{\label{footnote:g1}\url{https://www.unitree.com/g1}}, Agility Robotics Digit\footnote{\label{footnote:digit}\url{https://agilityrobotics.com/robots}}, the Robotiq 2F-85 gripper, and the Unitree H1 hand.

The HumanoidBench task suite includes $15$ distinct whole-body manipulation tasks involving a variety of interactions, e.g., unloading packages from a truck, wiping windows using a tool, catching and shooting a basketball. In addition, we provide $12$ locomotion tasks (not requiring hands' dexterity), which can serve as primitive skills for whole-body manipulation tasks and provide a set of easier tasks to verify algorithms. The benchmarking results on this task suite show how the state-of-the-art RL algorithms struggle with controlling the complex humanoid robot dynamics and solving the most challenging tasks, illustrating ample opportunities for future research.

\section{Related Work}
\label{sec:related_work}

Deep reinforcement learning (RL) has made rapid progress with the advent of standardized, simulated benchmarks, such as Atari~\citep{bellemare13arcade} and continuous control~\citep{brockman2016openai, tassa2018deepmind} benchmarks. In robotic manipulation, most existing simulated environments are limited to quasi-static, short-horizon skills, having focused on tasks like picking and placing~\citep{brockman2016openai, james2019rlbench, zhu2020robosuite, yu2019meta, mandlekar2021robomimic}, in-hand manipulation~\citep{plappert2018robotics, andrychowicz2020learning, caggiano2022myosuite}, and screwing~\citep{narang2022factory}.

Complex manipulation tasks, such as block stacking~\citep{duan2017one-shot}, kitchen tasks~\citep{gupta2019relay}, and table-top manipulation~\citep{kannan2021robodesk, mees2022calvin, corl2020softgym}, have been introduced but are still limited to a combination of pushing, picking, and placing. On the other hand, the IKEA furniture assembly environment~\citep{lee2021ikea}, BEHAVIOR~\citep{srivastava2021behavior, li2022behavior}, and Habitat~\citep{szot2021habitat} present diverse long-horizon (mobile) manipulation tasks, with their main focus being on high-level planning by abstracting complex low-level control problems, while FurnitureBench~\citep{heo2023furniturebench} introduces a simulated benchmark for complex long-horizon furniture assembly tasks with sophisticated low-level control. 

However, most of these benchmarks use a single-arm manipulation setup with either a parallel gripper or a dexterous hand \cite{caggiano2023myodex, plappert2018robotics}, limiting the types of object interactions and not addressing the challenges of coordinating multiple parts of a body~\citep{lee2020learning}, e.g., multiple fingers, arms, and legs. Robosuite~\cite{zhu2020robosuite}  includes a handful of bimanual manipulation tasks, while more recently \citet{chen2023bi} and \citet{zakka2023robopianist} have introduced additional benchmarks that require coordinating two floating robot hands, i.e., not attached to any arm base. 

While bimanual manipulation is one of the key objectives of humanoid robots, most benchmarks in humanoid research have so far focused on the locomotion challenges~\citep{caggiano2022myosuite, lee2019scalable, peng2018deepmimic, alhafez2023locomujoco}. In this regard, such simulations have accelerated research on control algorithms \cite{berg2023sar, peng2018sfv, peng2021amp, merel2020catch}, ultimately leading to achieve robust humanoid locomotion in the real world \cite{adu2023exploring, radosavovic2023learning, cheng2024expressive}. 

Recent works have extended humanoid simulations to different domains involving a certain degree of manipulation, i.e., tennis \cite{zhang2023learning}, soccer \cite{haarnoja2023learning}, ball manipulation \cite{wang2023physhoi} and catching \cite{mattern2024mimo}, and box moving \cite{xie2023hierarchical}. However, all these works focus on demonstrating their approaches on specific humanoid tasks and lack a diversity of tasks. In addition, most of the previous work focuses on simplistic humanoid models \cite{mattern2024mimo, wang2023physhoi}, leading to inaccurate physics and collision handling. This motivates us to implement a \emph{comprehensive} simulated humanoid benchmark based on real-world hardware and consisting of a diverse set of whole-body control tasks with careful design choices for diversity and usability. 

In contrast to prior robotic simulation benchmarks, HumanoidBench presents a broader set of challenges, featuring high-dimensional action spaces and DoFs, resulting from humanoid robots and dexterous hands, and a variety of long-horizon tasks, which cover a comprehensive set of robotic locomotion and manipulation skills, as summarized in \Cref{tab:benchmarks}.

Finally, we note how in the literature, tasks that require long-term planning with a high-dimensional action space have been addressed with hierarchical reinforcement learning (HRL), which decouples low-level and high-level planning in a reinforcement learning paradigm~\citep{lin1993hierarchical, bacon2017option-critic, nachum2018data, lee2019composing, gupta2019relay, lee2020learning, pertsch2020spirl}. In the context of humanoids, we propose an HRL paradigm to show how a specific set of low-level skills (e.g., standing, walking) facilitates learning of higher-level tasks.

\begin{figure}[t]
    \centering
    \includegraphics[width=1.0\linewidth]{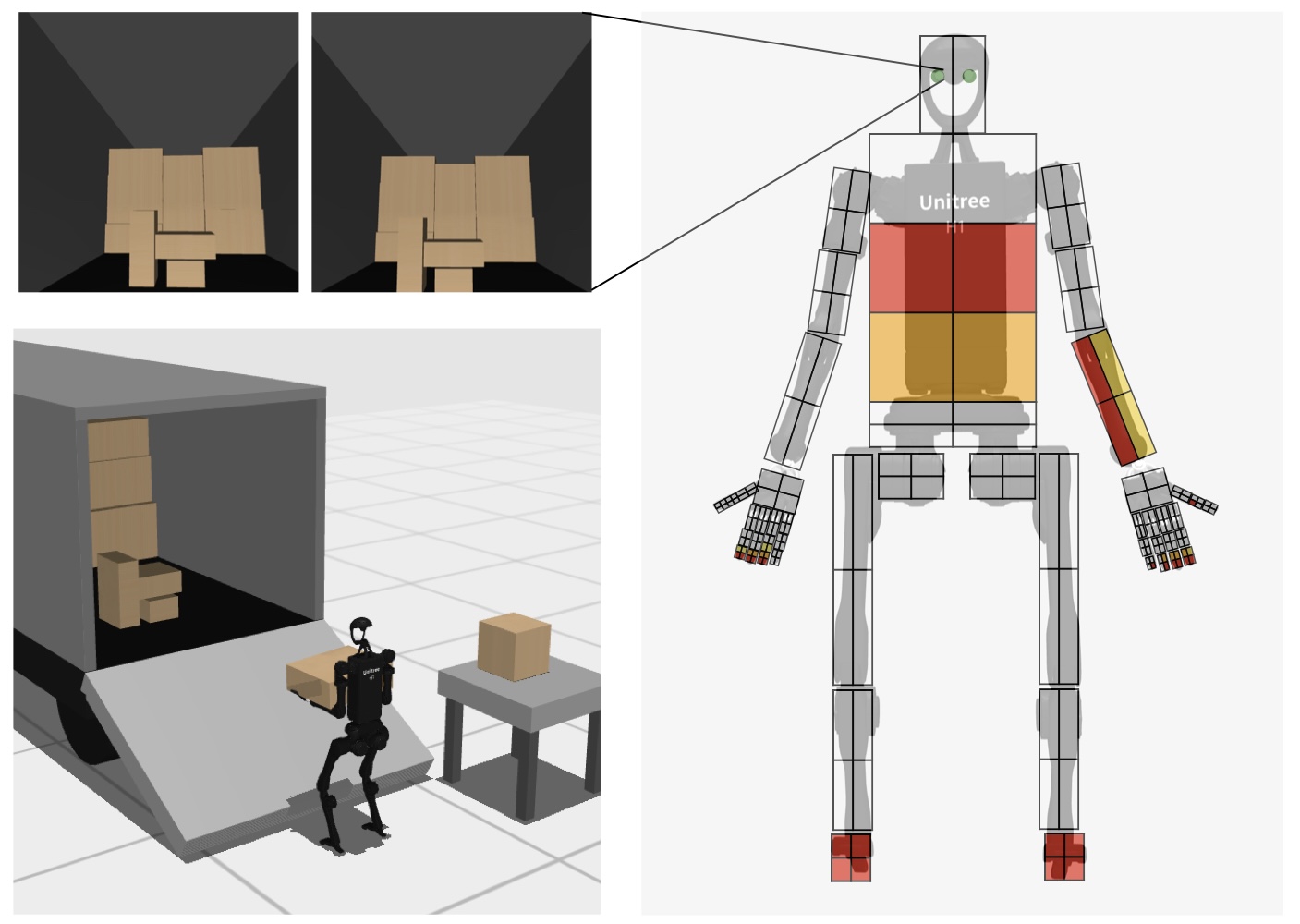}
    \caption{Example egocentric visual (top-left) and whole-body tactile (right) observations when the humanoid interacts with a package in the \texttt{truck} environment. In the right figure, the two cameras on the robot head are highlighted in green, while continuous tactile pressure readings are indicated in shades of red (strong pressure) and yellow (mild pressure). Note that for ease of visualization, we are not showing shear forces and tactile readings on the back of the robot, which are also implemented in our environment.}
    \label{fig:humanoid_env}
\end{figure}

\begin{figure*}[ht!]
    \centering
    \captionsetup[subfigure]{labelformat=empty}
    \begin{subfigure}[ht]{0.325\textwidth}
        \centering
        \includegraphics[width=0.49\textwidth]{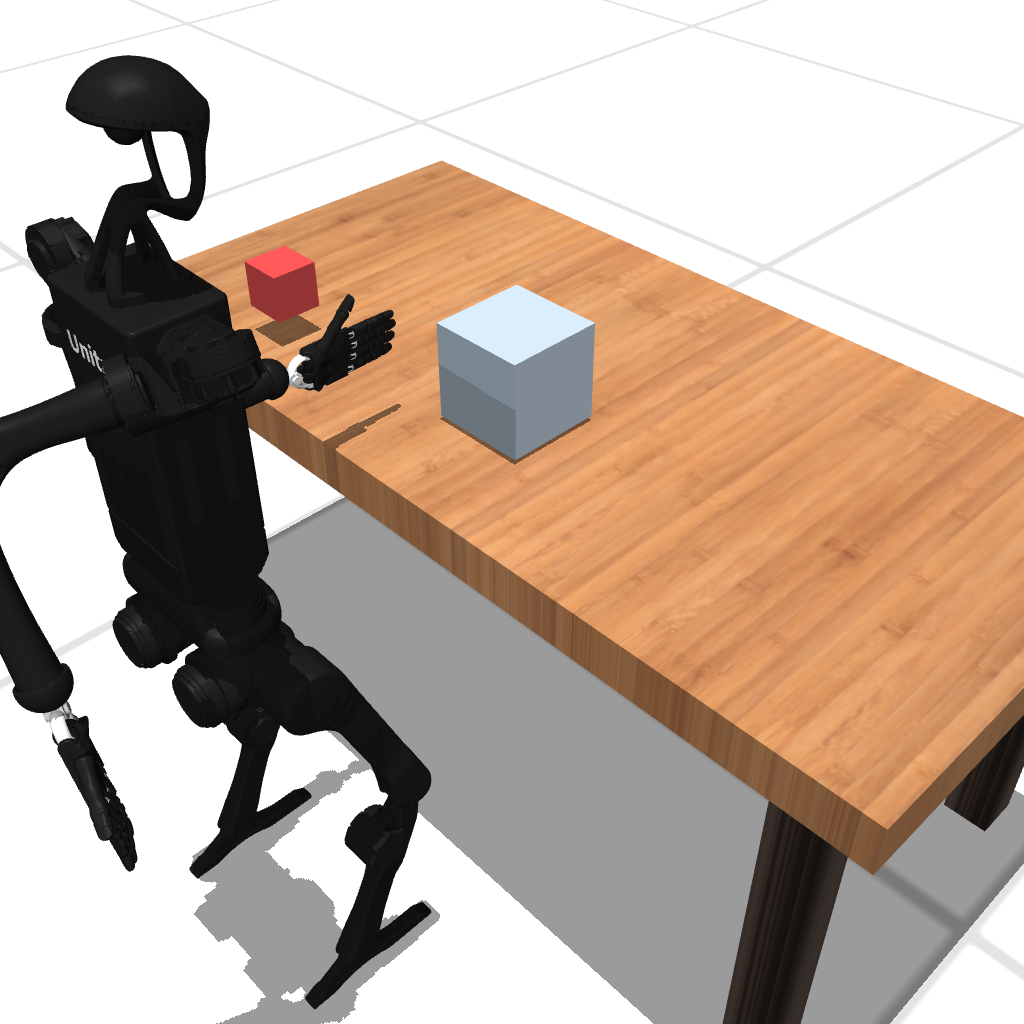}
        \includegraphics[width=0.49\textwidth]{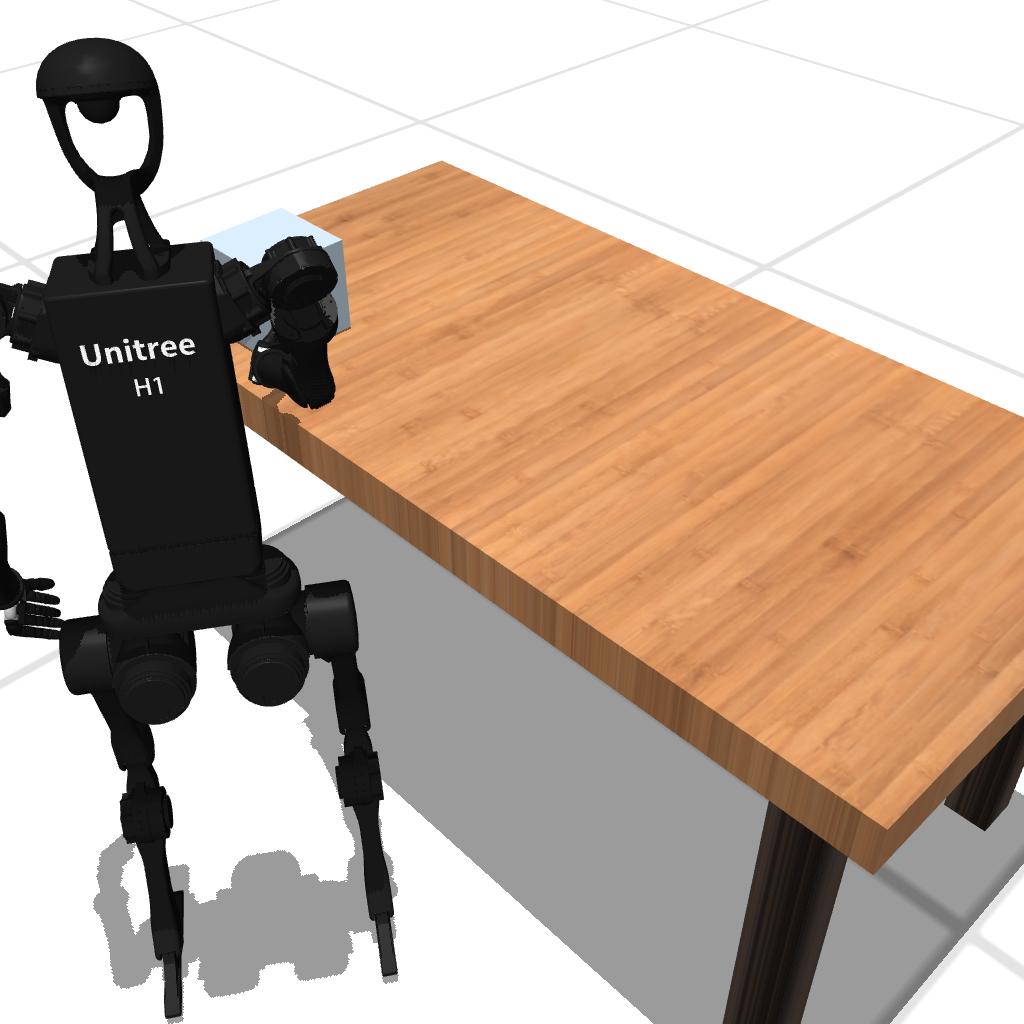}
        \vspace{-1.7em}
        \caption{\texttt{push}}
    \end{subfigure}
    \hfill
    \begin{subfigure}[ht]{0.325\textwidth}
        \centering
        \includegraphics[width=0.49\textwidth]{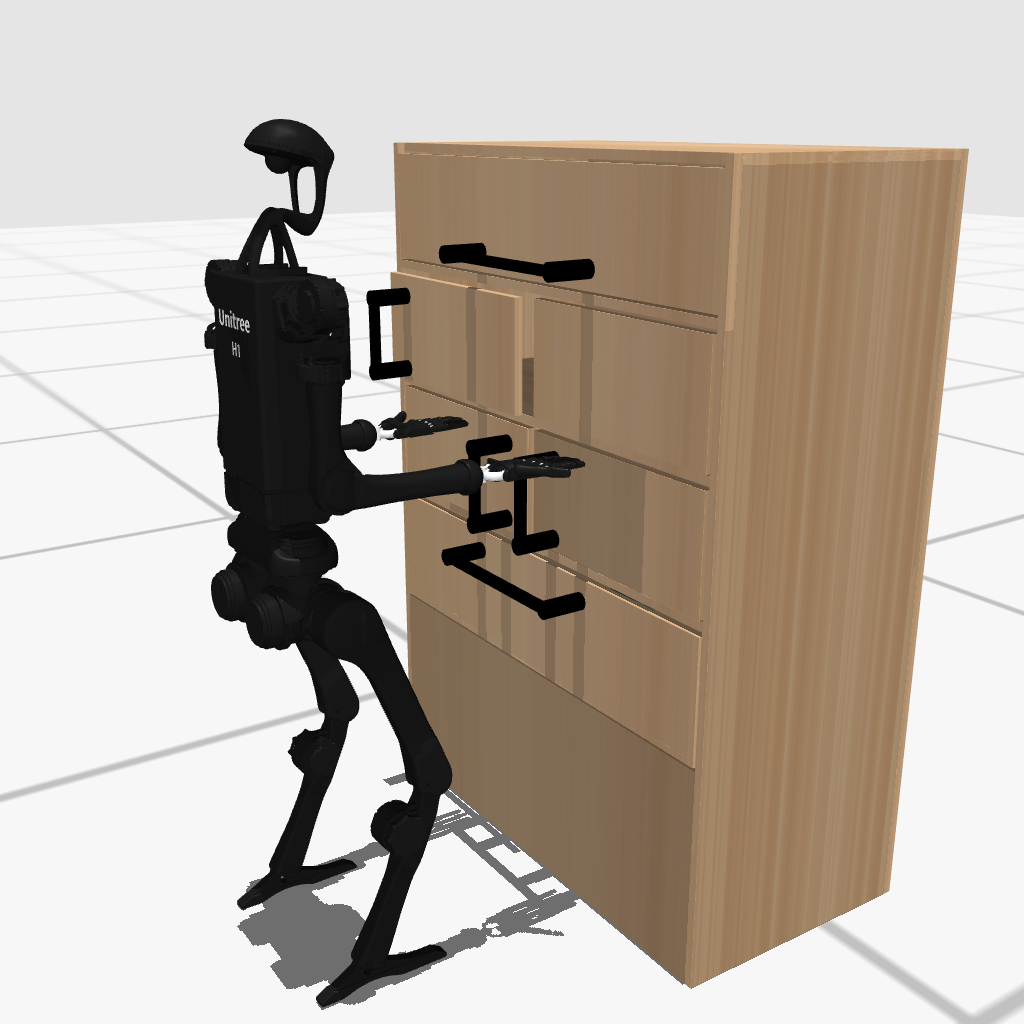}
        \includegraphics[width=0.49\textwidth]{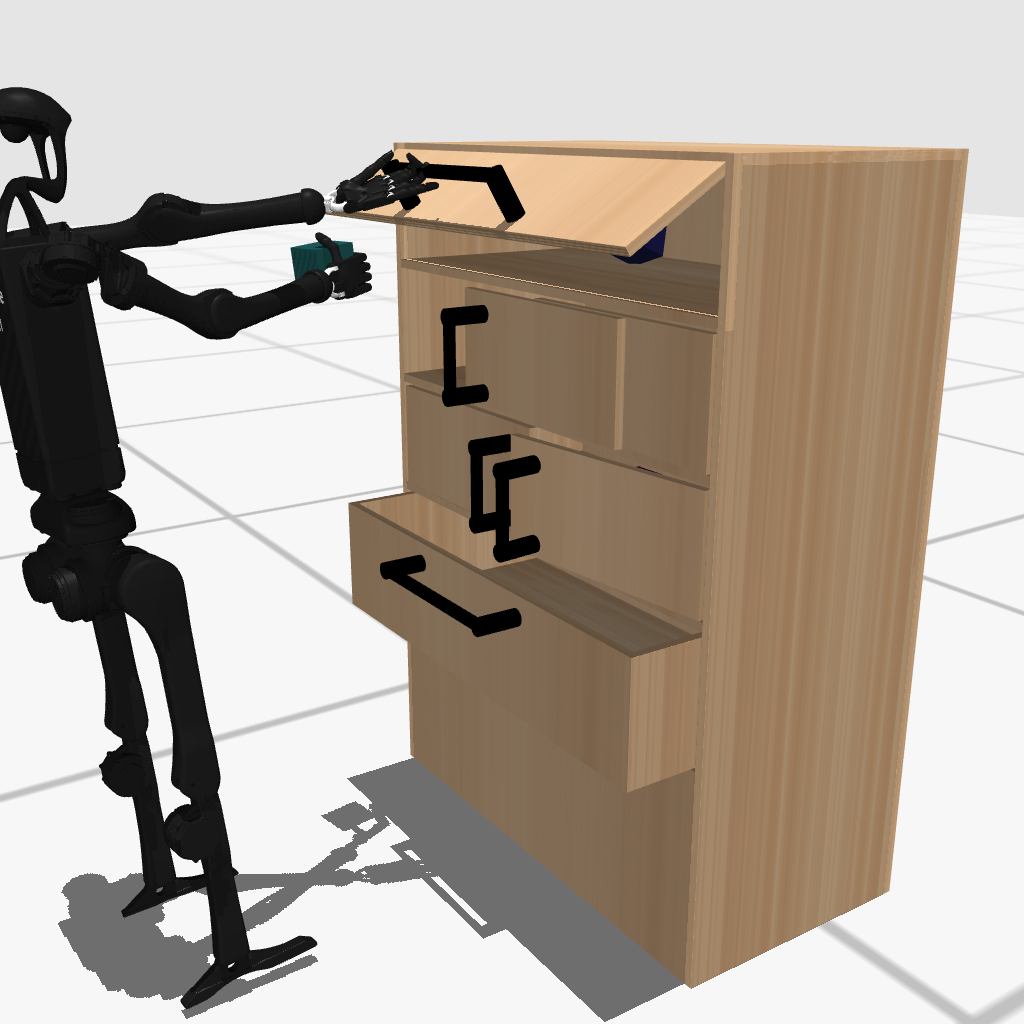}
        \vspace{-1.7em}
        \caption{\texttt{cabinets}}
    \end{subfigure}
    \hfill
    \begin{subfigure}[ht]{0.325\textwidth}
        \centering
        \includegraphics[width=0.49\textwidth]{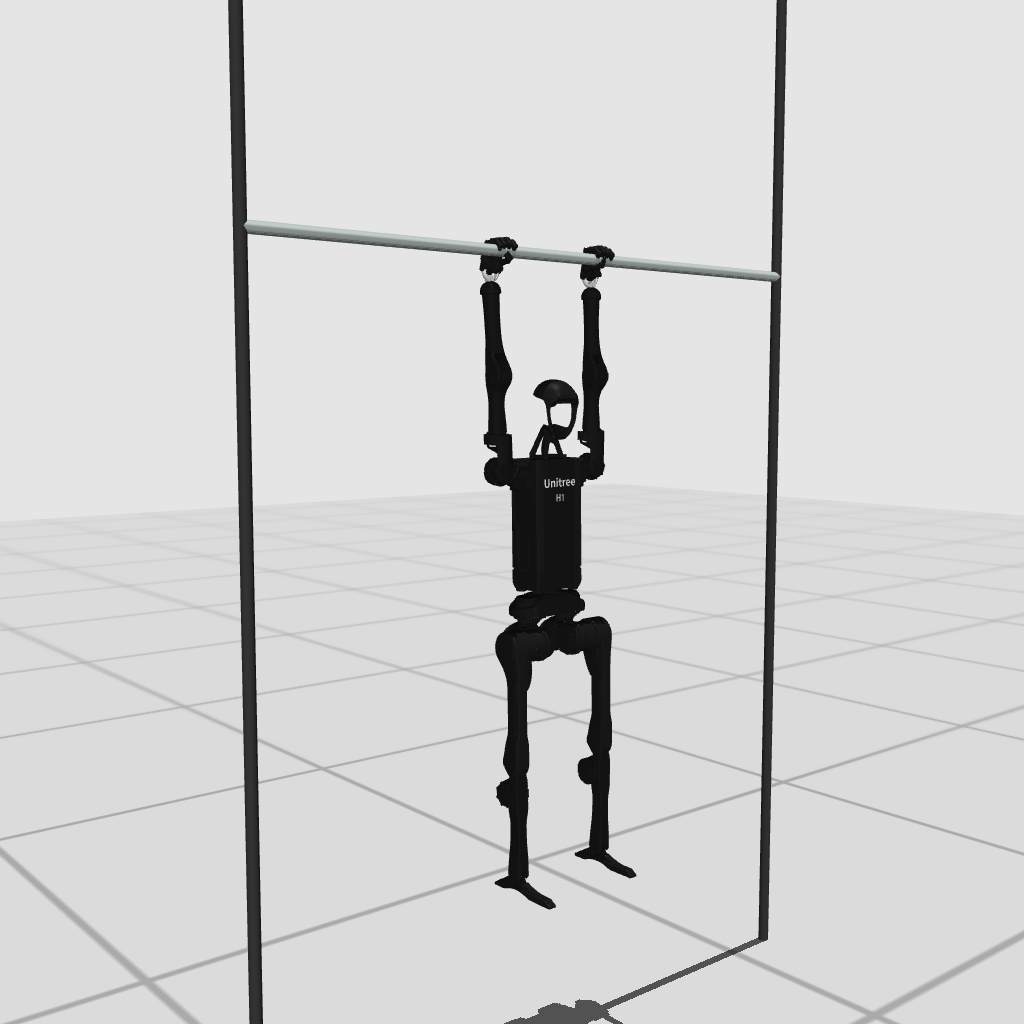}
        \includegraphics[width=0.49\textwidth]{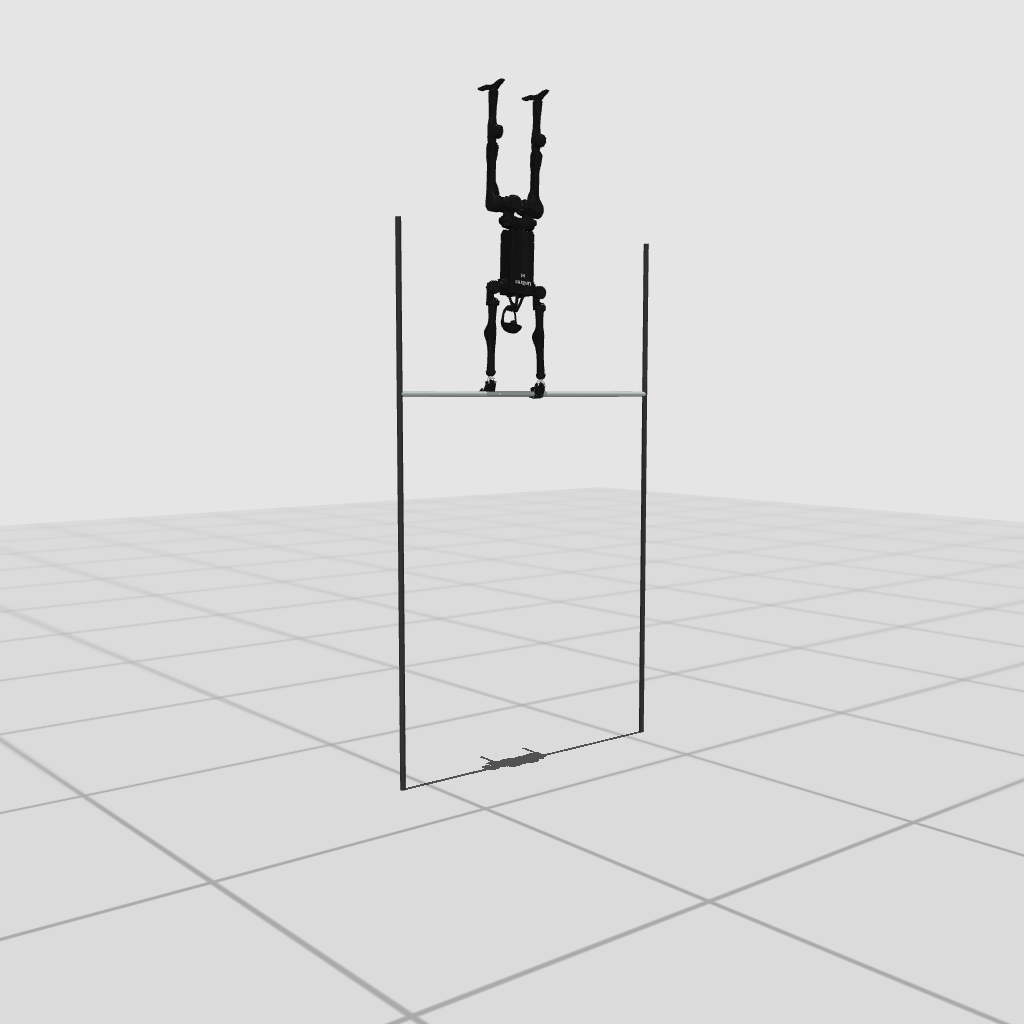}
        \vspace{-1.5em}
        \caption{\texttt{high\_bar}}
    \end{subfigure}
    \vspace{0.6em}
    \\
    \begin{subfigure}[ht]{0.325\textwidth}
        \centering
        \includegraphics[width=0.49\textwidth]{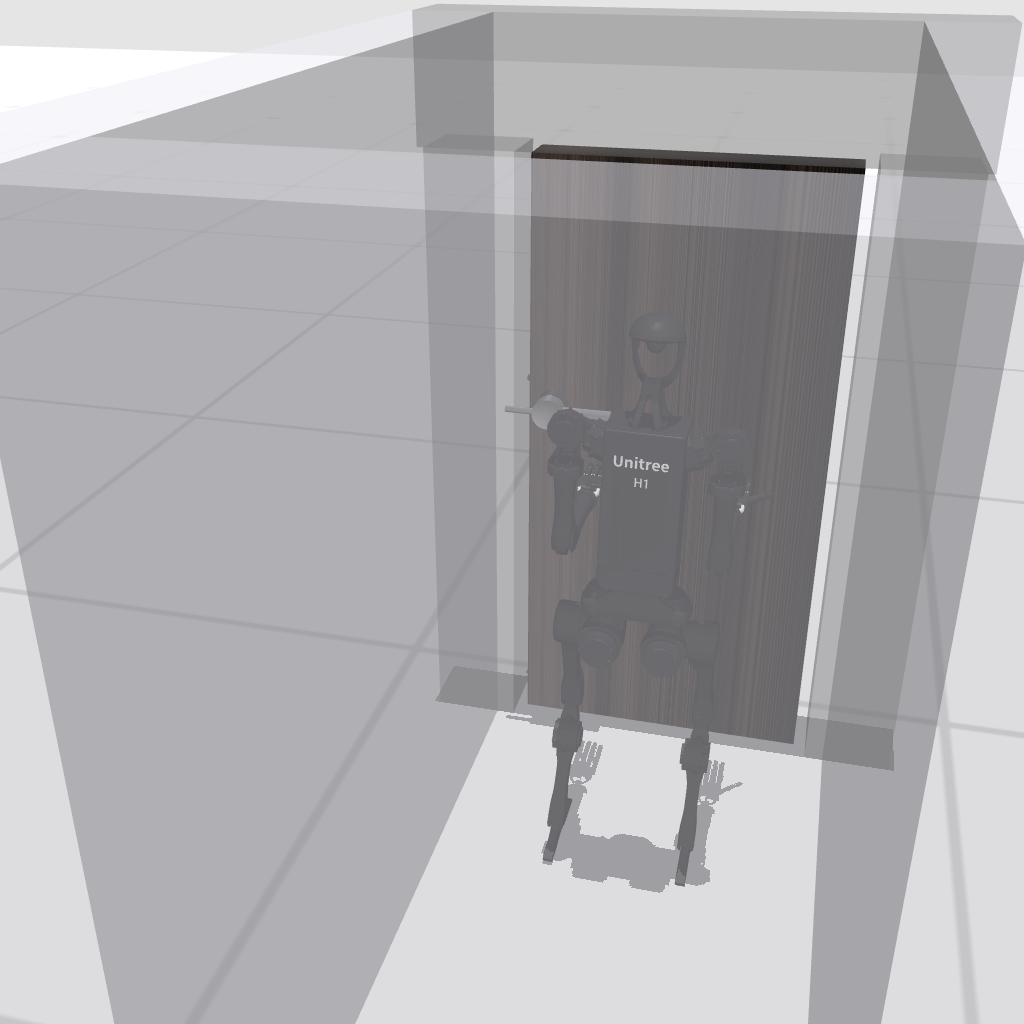}
        \includegraphics[width=0.49\textwidth]{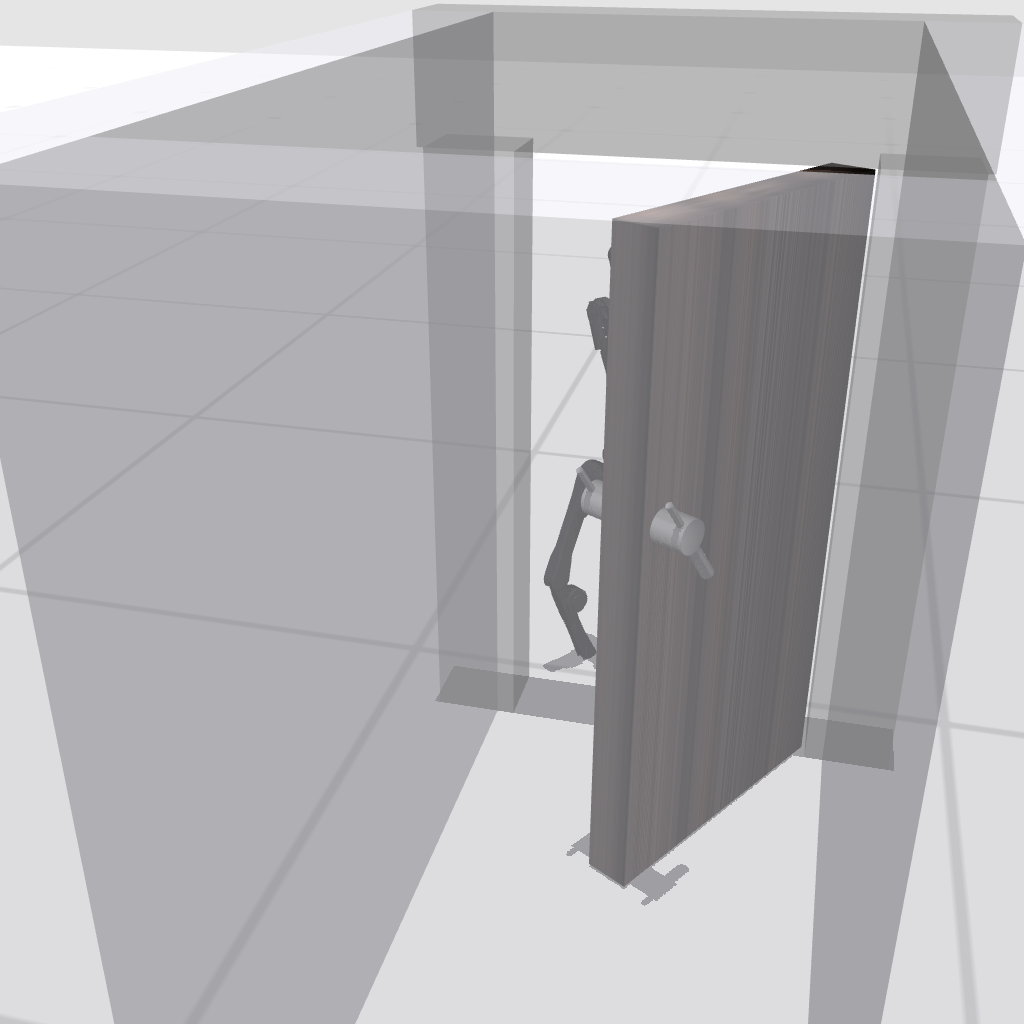}
        \vspace{-1.7em}
        \caption{\texttt{door}}
    \end{subfigure}
    \hfill
    \begin{subfigure}[ht]{0.325\textwidth}
        \centering
        \includegraphics[width=0.49\textwidth]{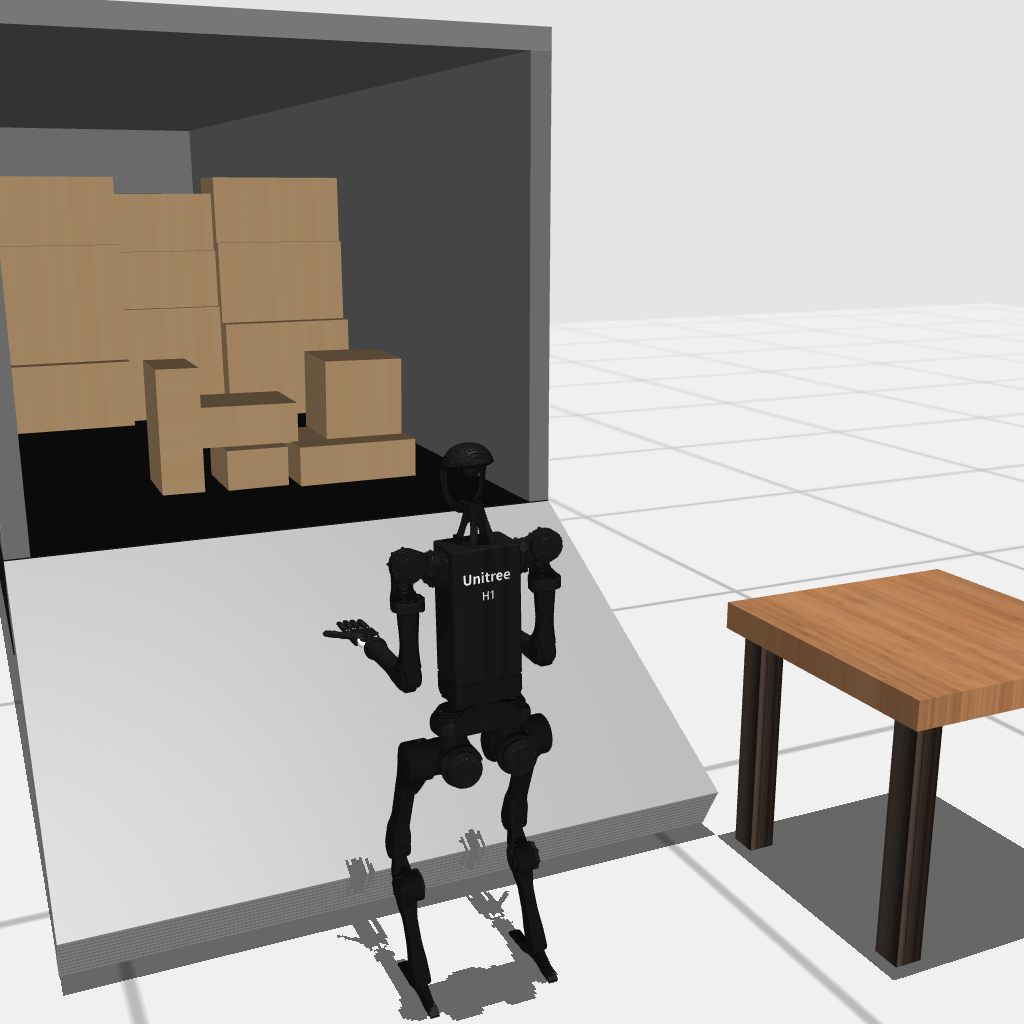}
        \includegraphics[width=0.49\textwidth]{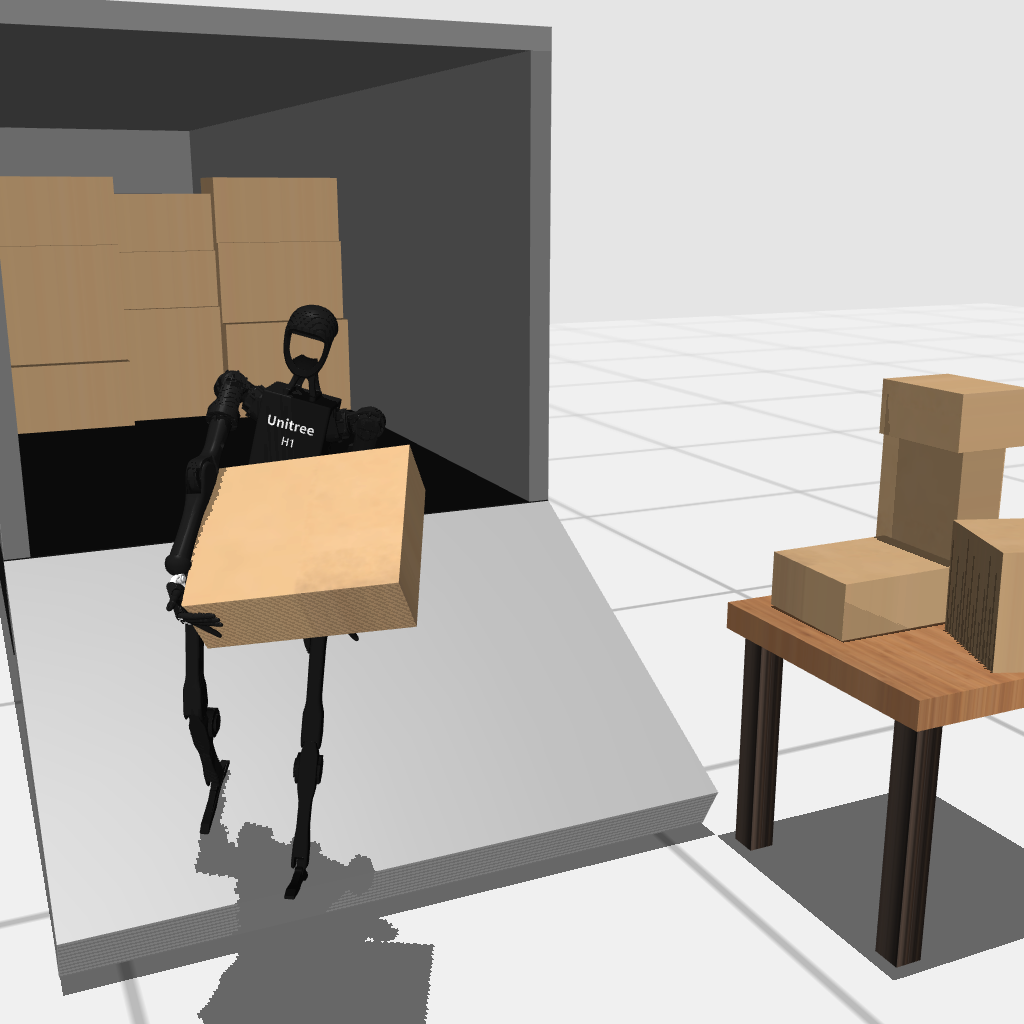}
        \vspace{-1.7em}
        \caption{\texttt{truck}}
    \end{subfigure}
    \hfill
    \begin{subfigure}[ht]{0.325\textwidth}
        \centering
        \includegraphics[width=0.49\textwidth]{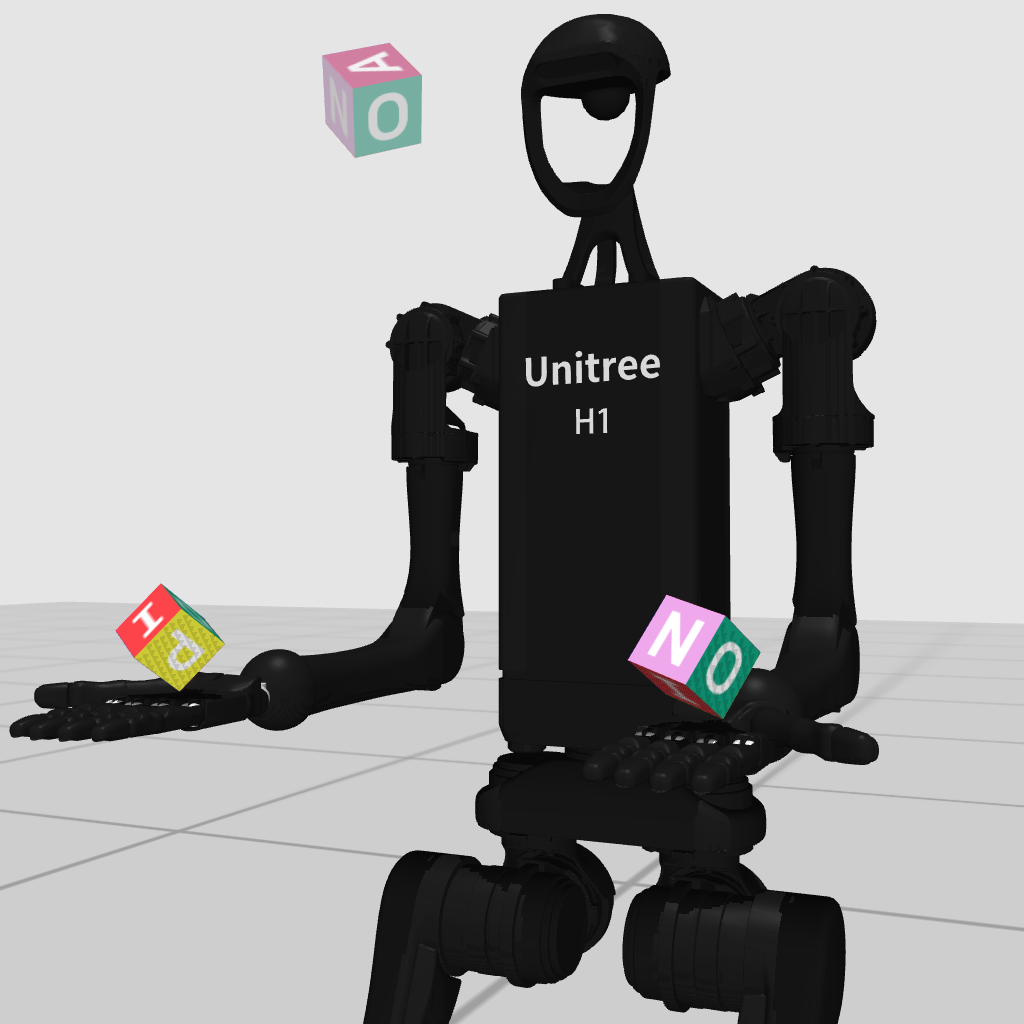}
        \includegraphics[width=0.49\textwidth]{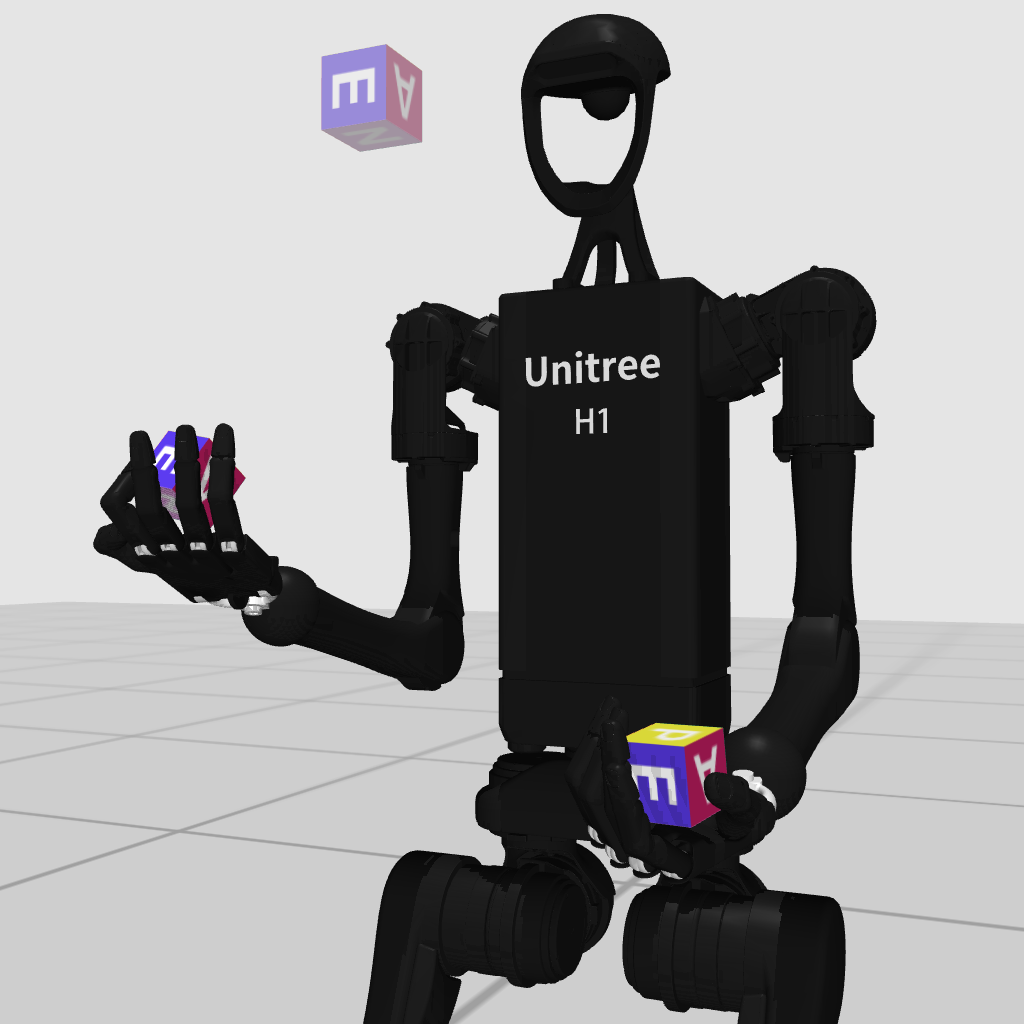}
        \vspace{-1.5em}
        \caption{\texttt{cubes}}
    \end{subfigure}
    \vspace{0.6em}
    \\
    \begin{subfigure}[ht]{0.325\textwidth}
        \centering
        \includegraphics[width=0.49\textwidth]{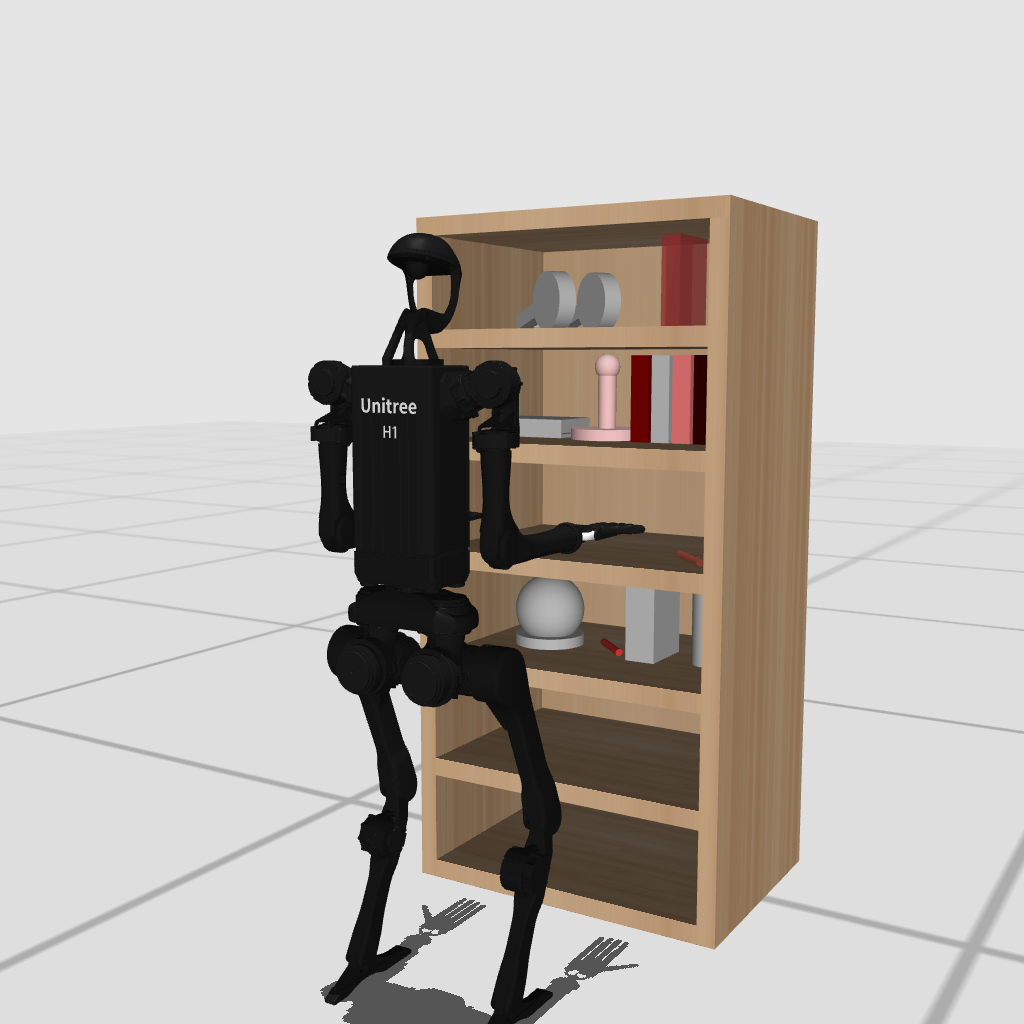}
        \includegraphics[width=0.49\textwidth]{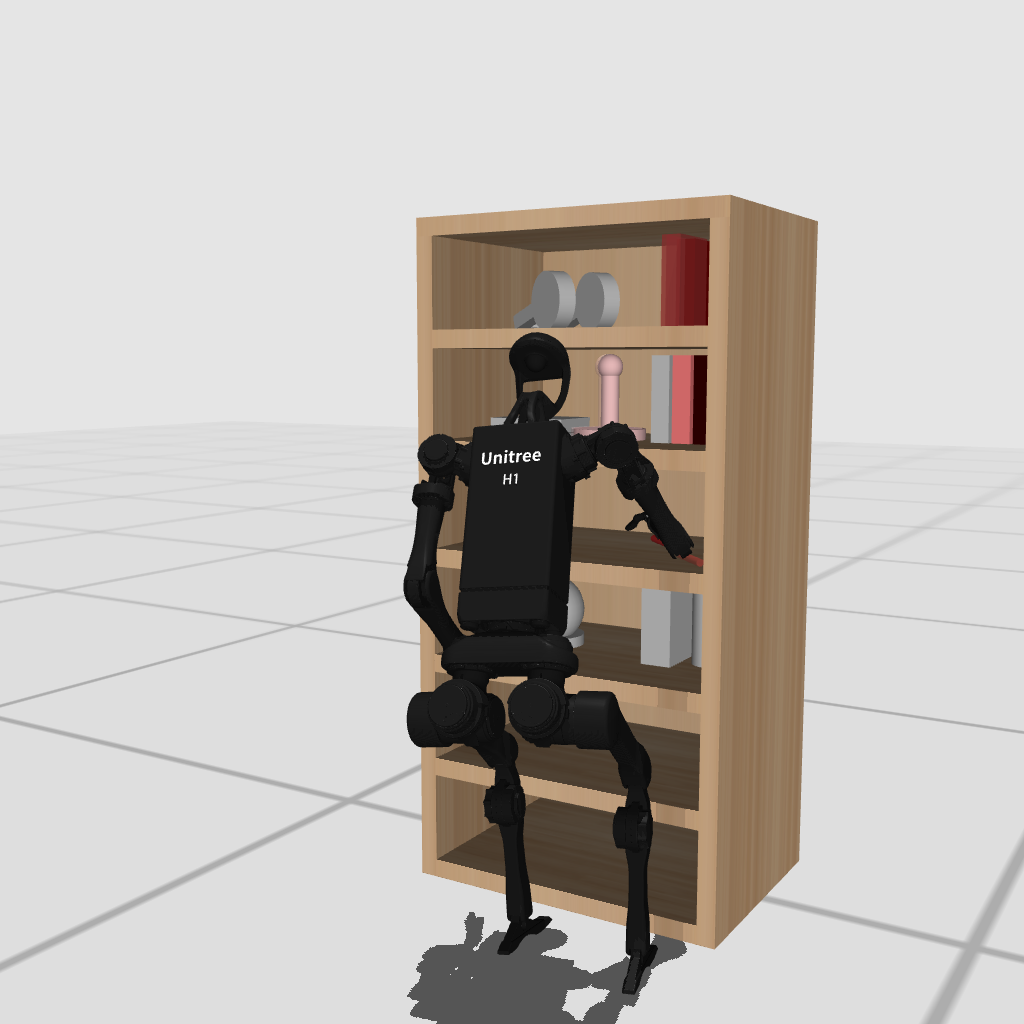}
        \vspace{-1.7em}
        \caption{\texttt{bookshelf}}
    \end{subfigure}
    \hfill
    \begin{subfigure}[ht]{0.325\textwidth}
        \centering
        \includegraphics[width=0.49\textwidth]{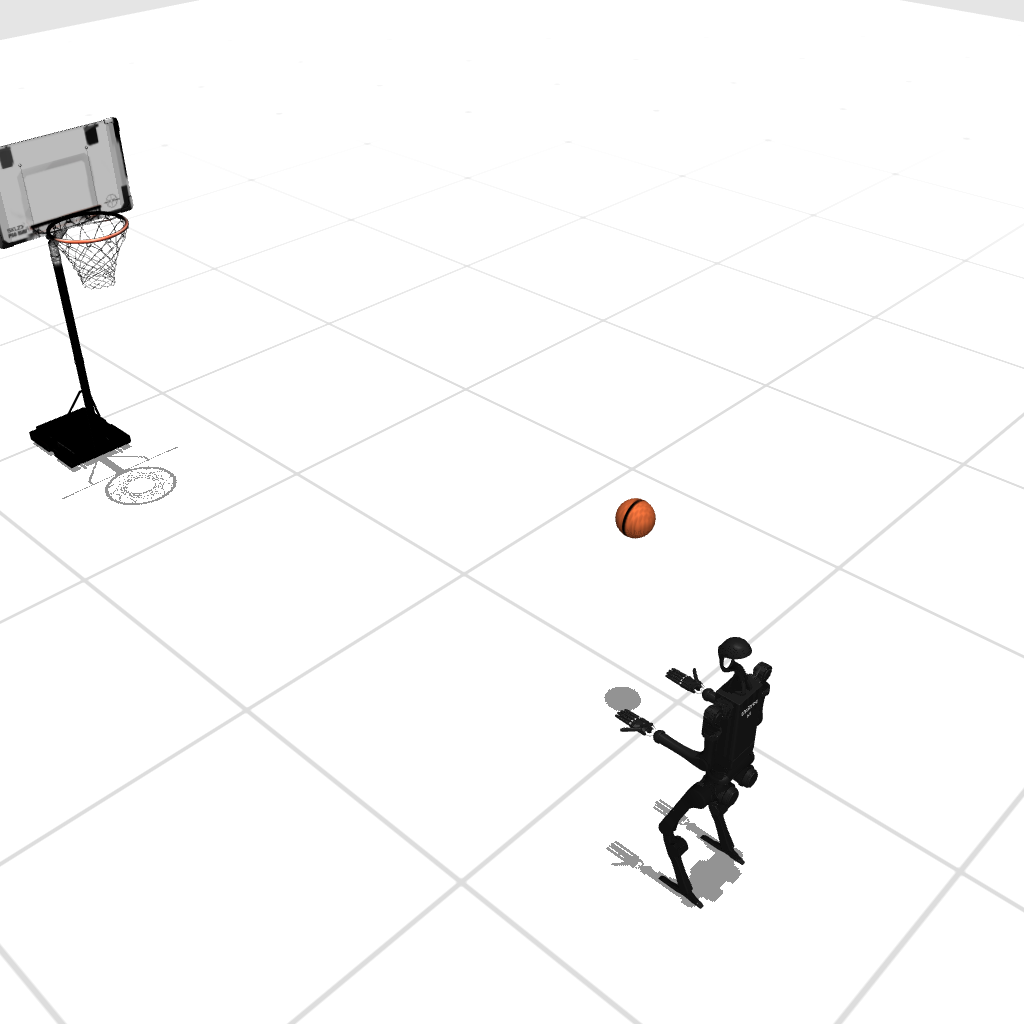}
        \includegraphics[width=0.49\textwidth]{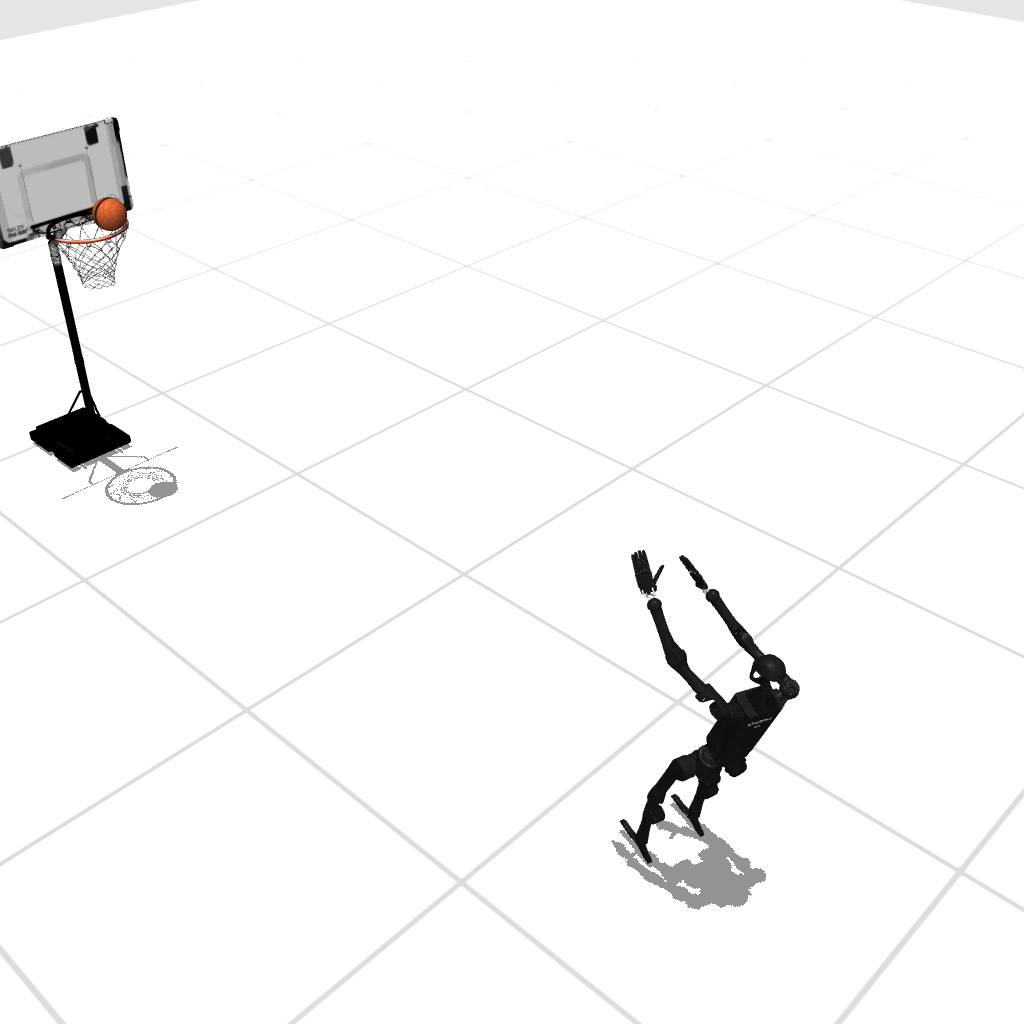}
        \vspace{-1.7em}
        \caption{\texttt{basketball}}
    \end{subfigure}
    \hfill
    \begin{subfigure}[ht]{0.325\textwidth}
        \centering
        \includegraphics[width=0.49\textwidth]{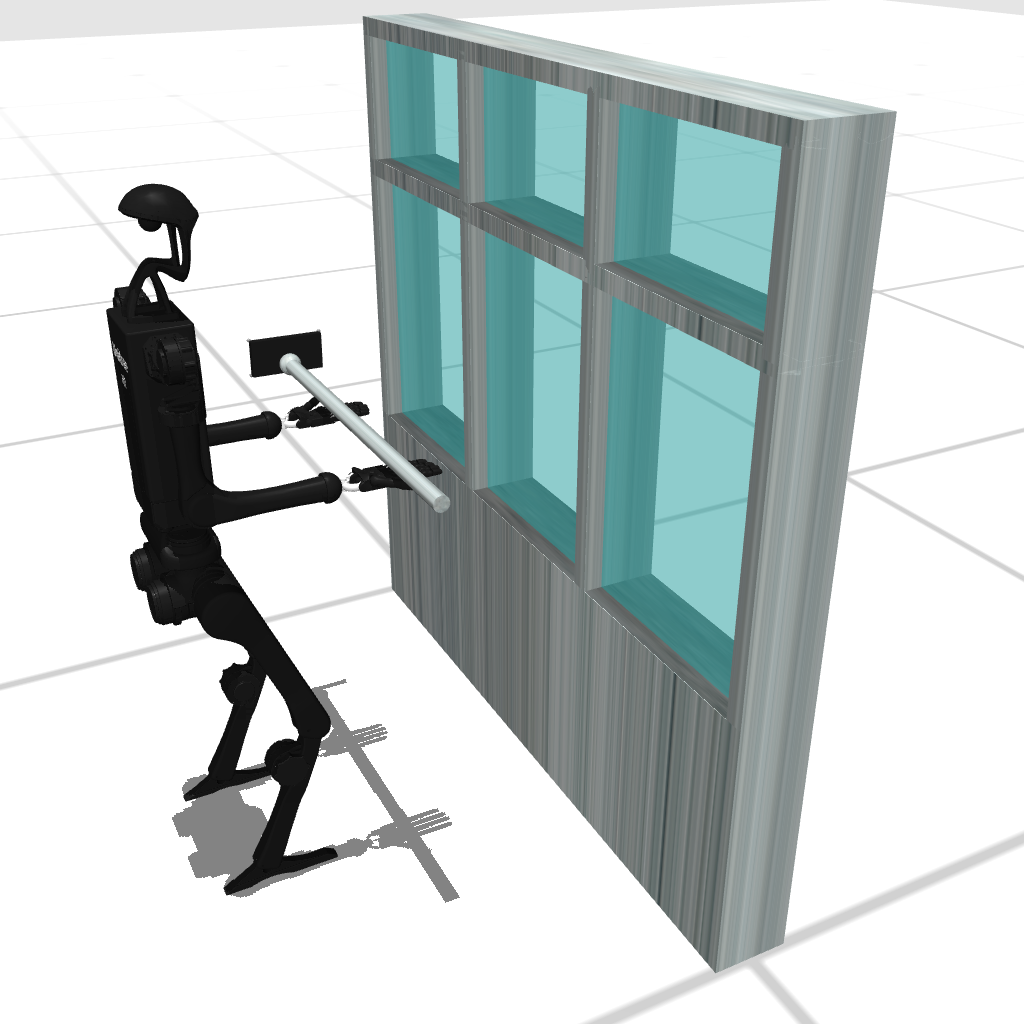}
        \includegraphics[width=0.49\textwidth]{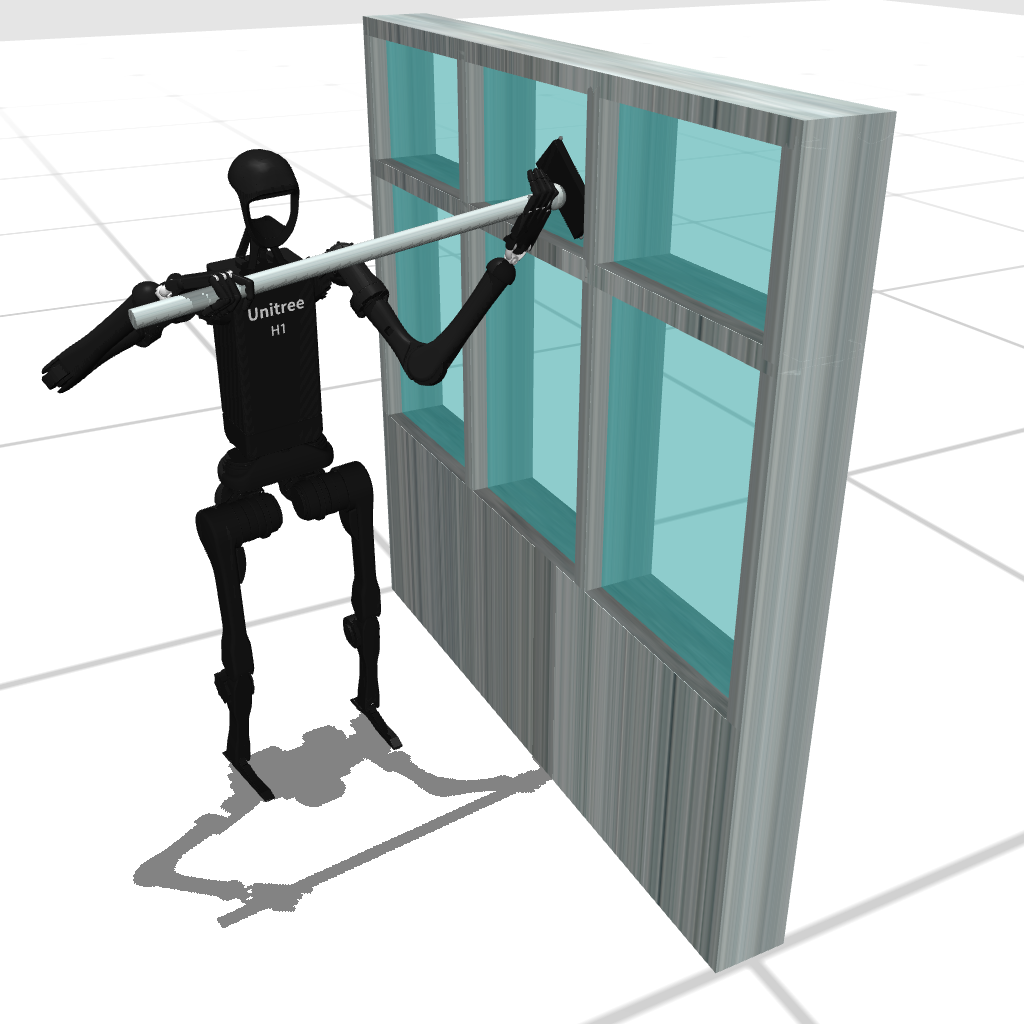}
        \vspace{-1.5em}
        \caption{\texttt{window}}
    \end{subfigure}
    \vspace{0.6em}
    \\
    \begin{subfigure}[ht]{0.325\textwidth}
        \centering
        \includegraphics[width=0.49\textwidth]{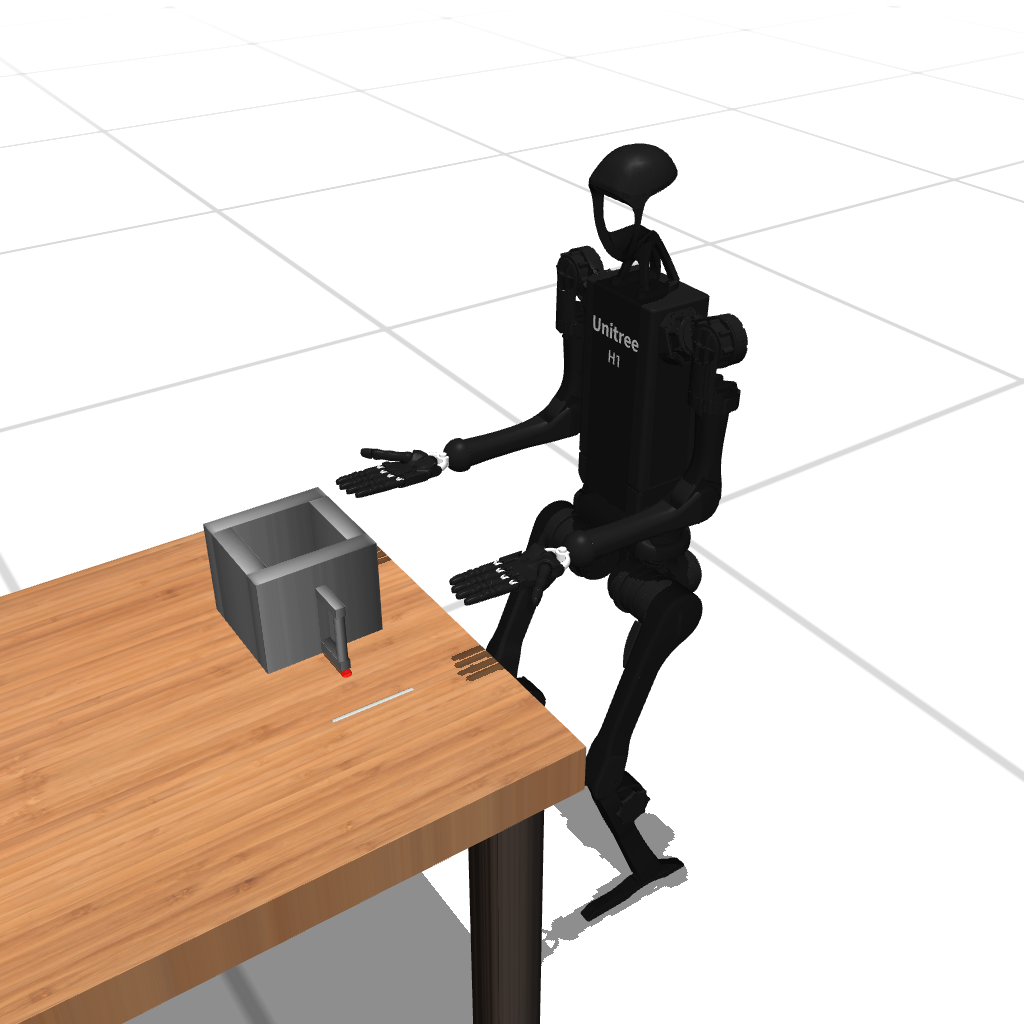}
        \includegraphics[width=0.49\textwidth]{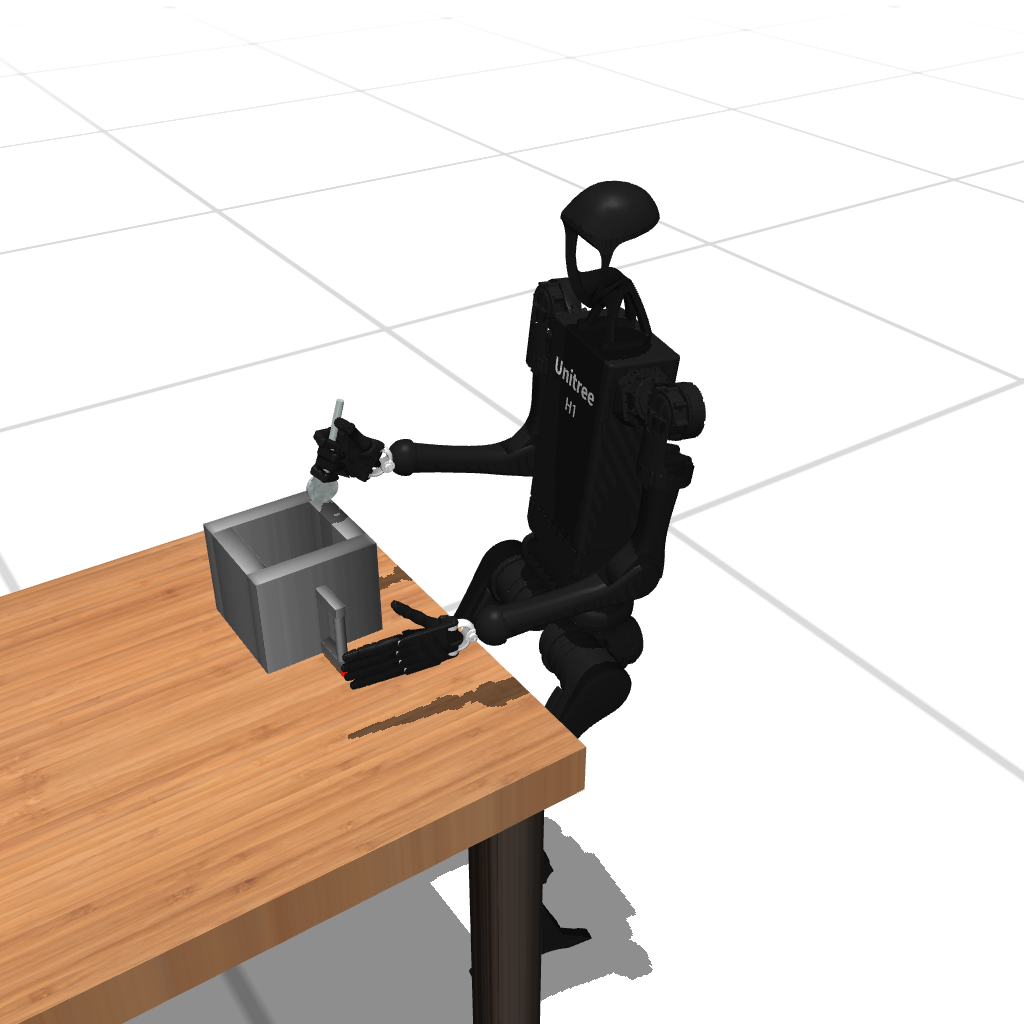}
        \vspace{-1.7em}
        \caption{\texttt{spoon}}
    \end{subfigure}
    \hfill
    \begin{subfigure}[ht]{0.325\textwidth}
        \centering
        \includegraphics[width=0.49\textwidth]{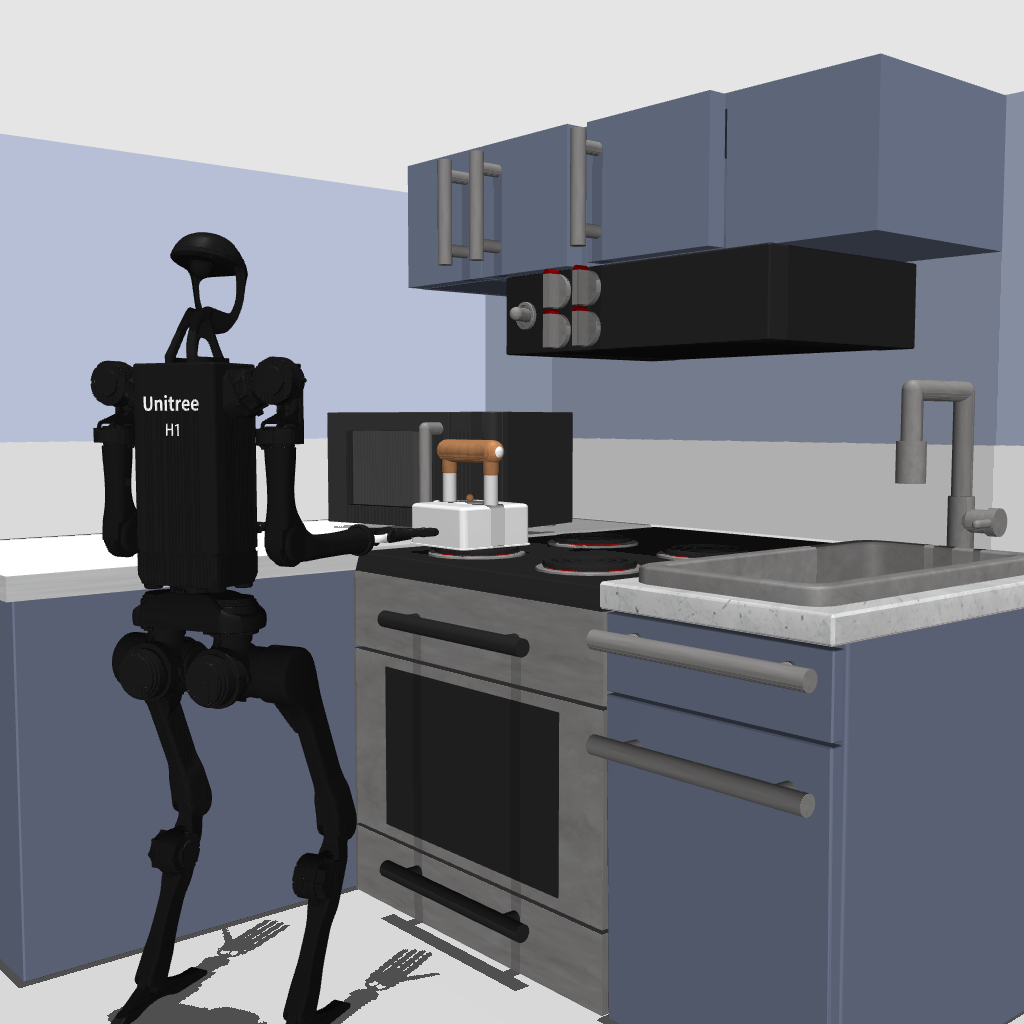}
        \includegraphics[width=0.49\textwidth]{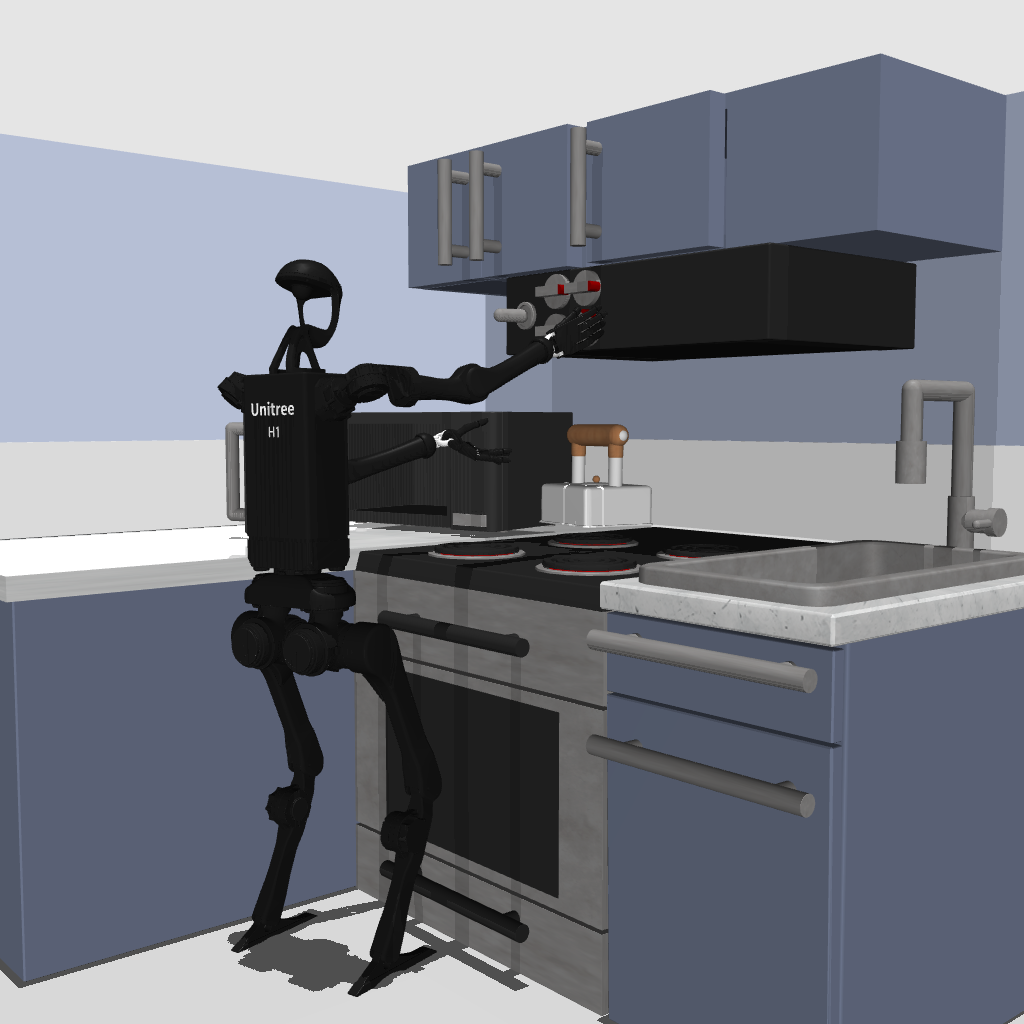}
        \vspace{-1.7em}
        \caption{\texttt{kitchen}}
    \end{subfigure}
    \hfill
    \begin{subfigure}[ht]{0.325\textwidth}
        \centering
        \includegraphics[width=0.49\textwidth]{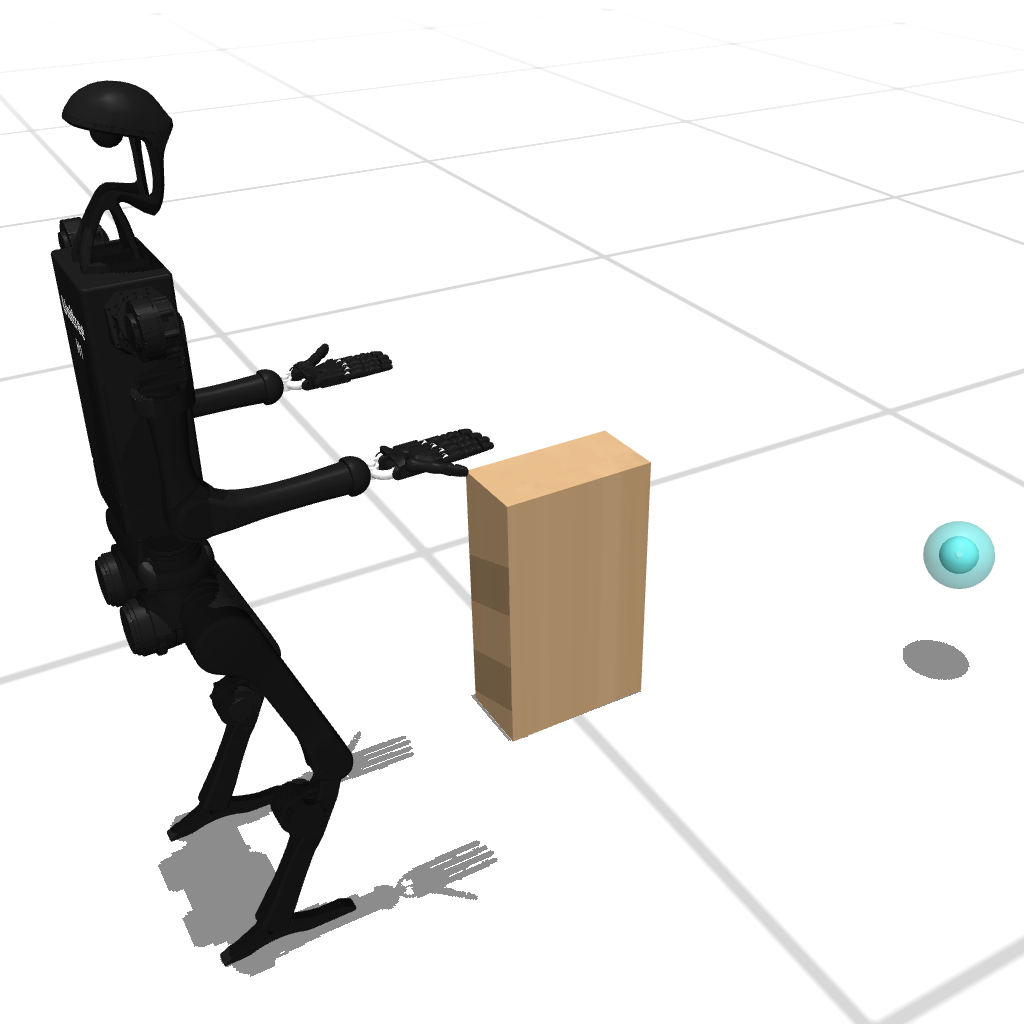}
        \includegraphics[width=0.49\textwidth]{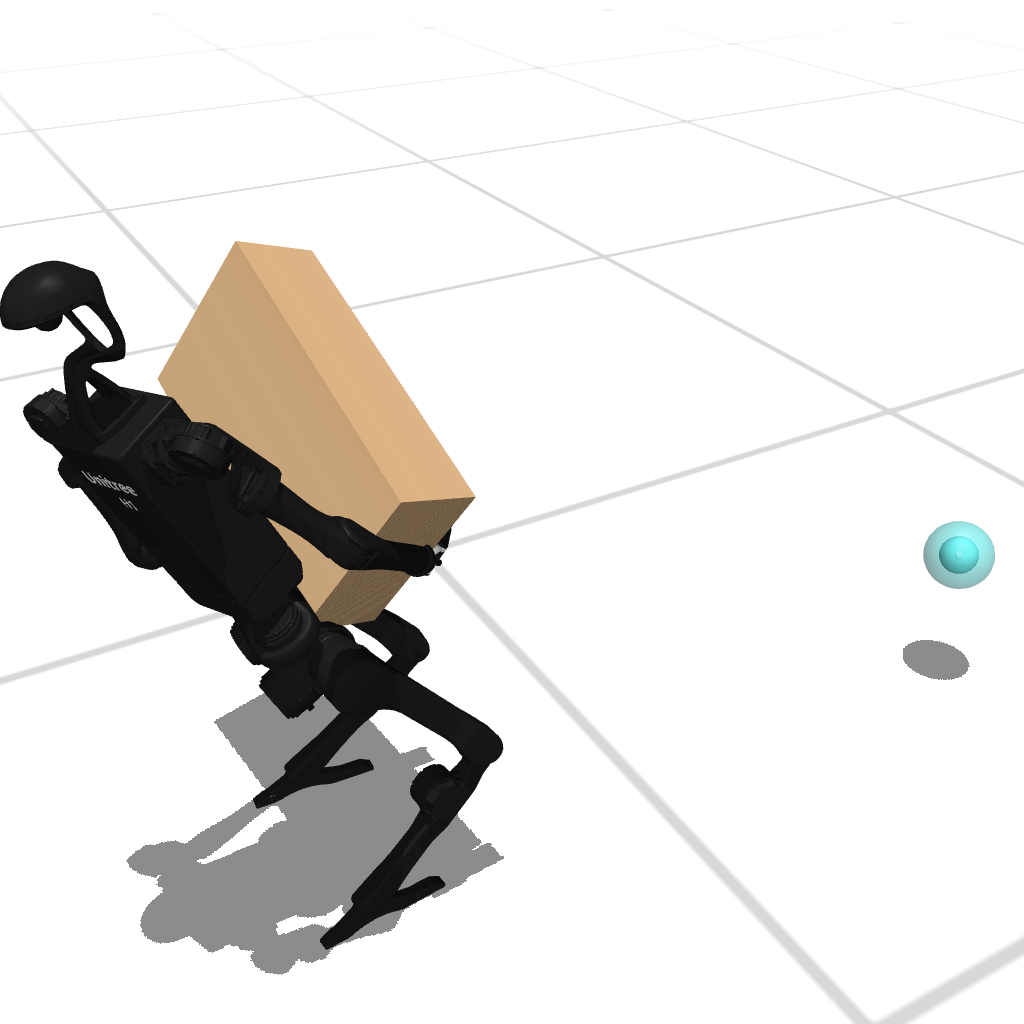}
        \vspace{-1.5em}
        \caption{\texttt{package}}
    \end{subfigure}
    \vspace{0.6em}
    \\
    \begin{subfigure}[ht]{0.325\textwidth}
        \centering
        \includegraphics[width=0.49\textwidth]{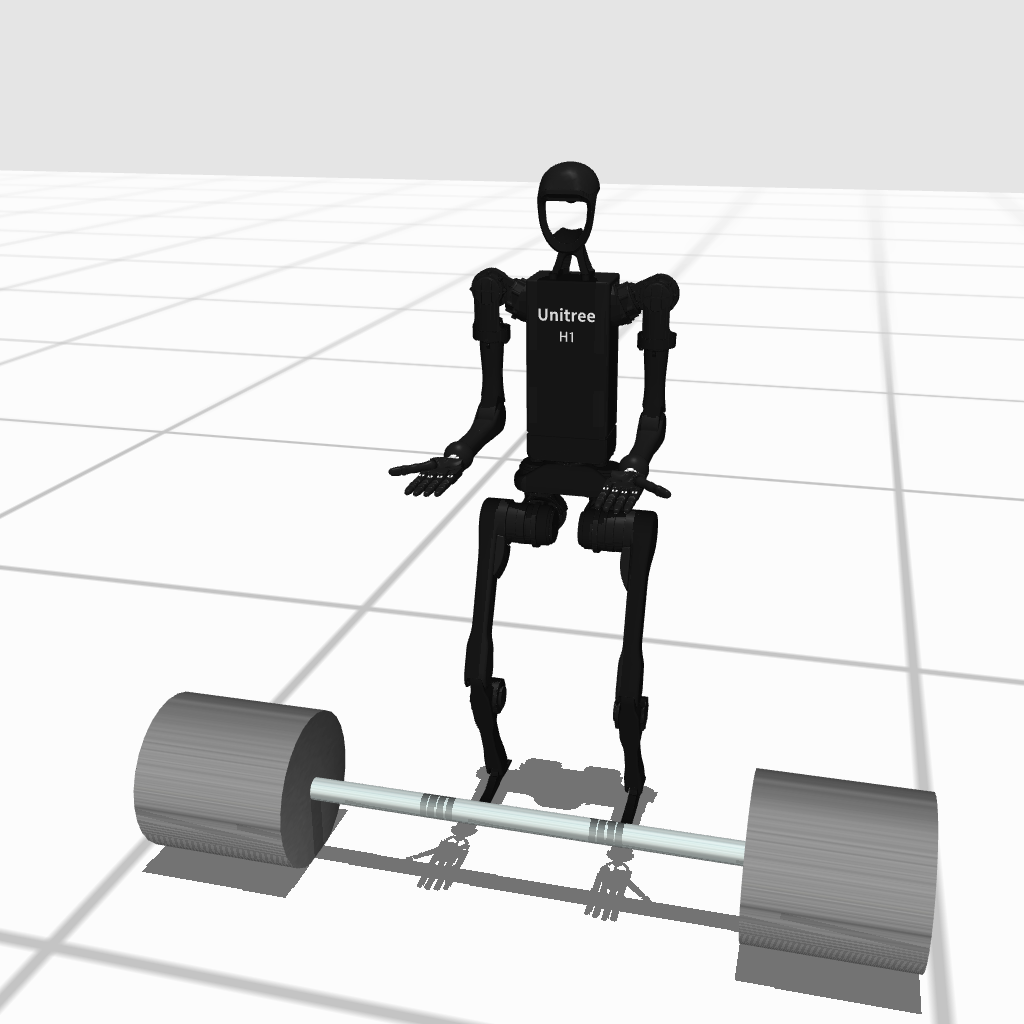}
        \includegraphics[width=0.49\textwidth]{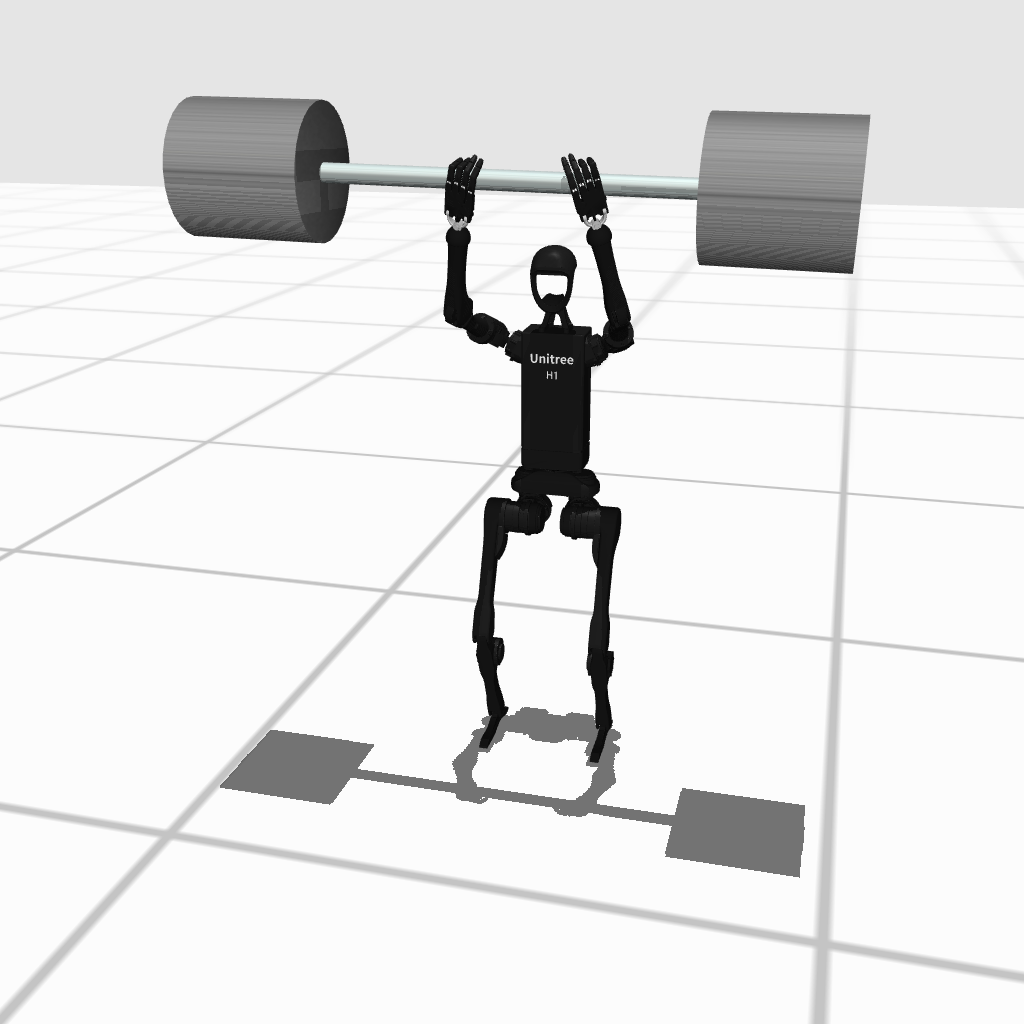}
        \vspace{-1.7em}
        \caption{\texttt{powerlift}}
    \end{subfigure}
    \hfill
    \begin{subfigure}[ht]{0.325\textwidth}
        \centering
        \includegraphics[width=0.49\textwidth]{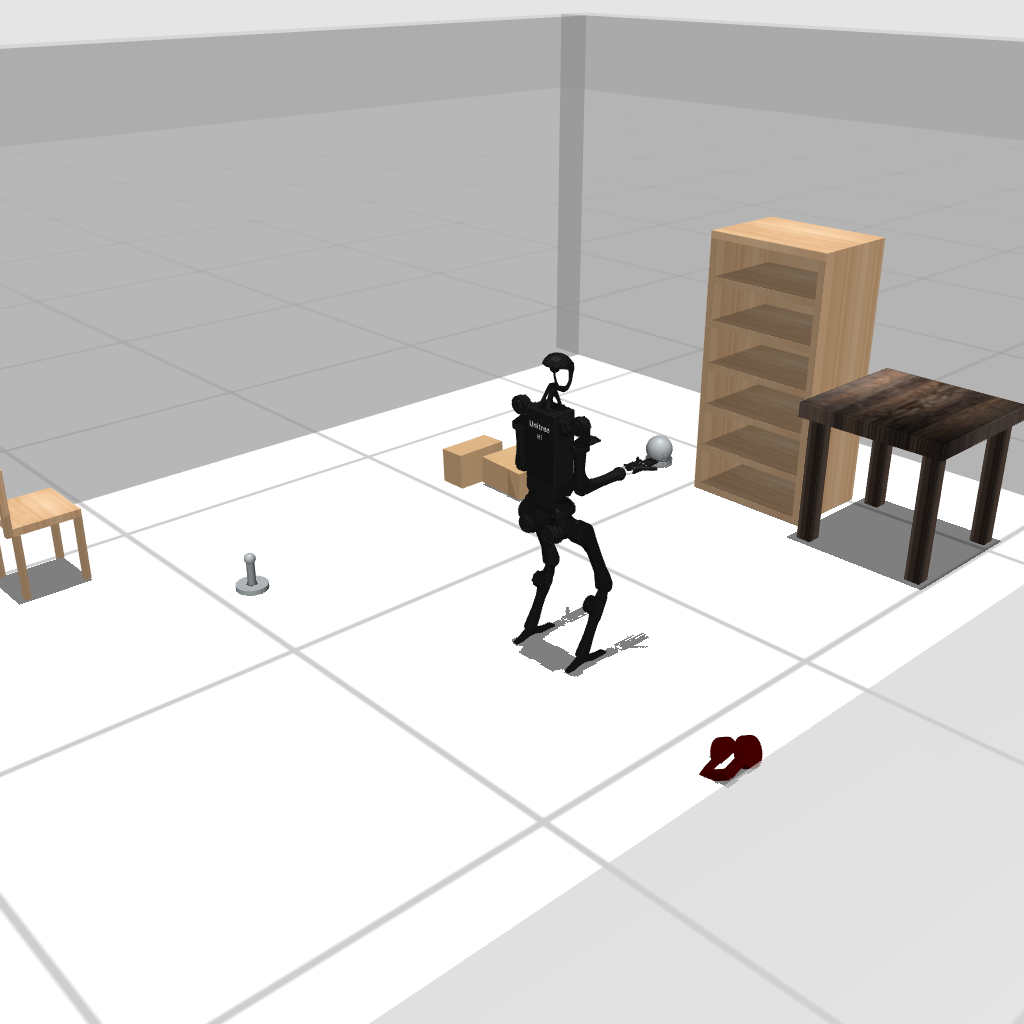}
        \includegraphics[width=0.49\textwidth]{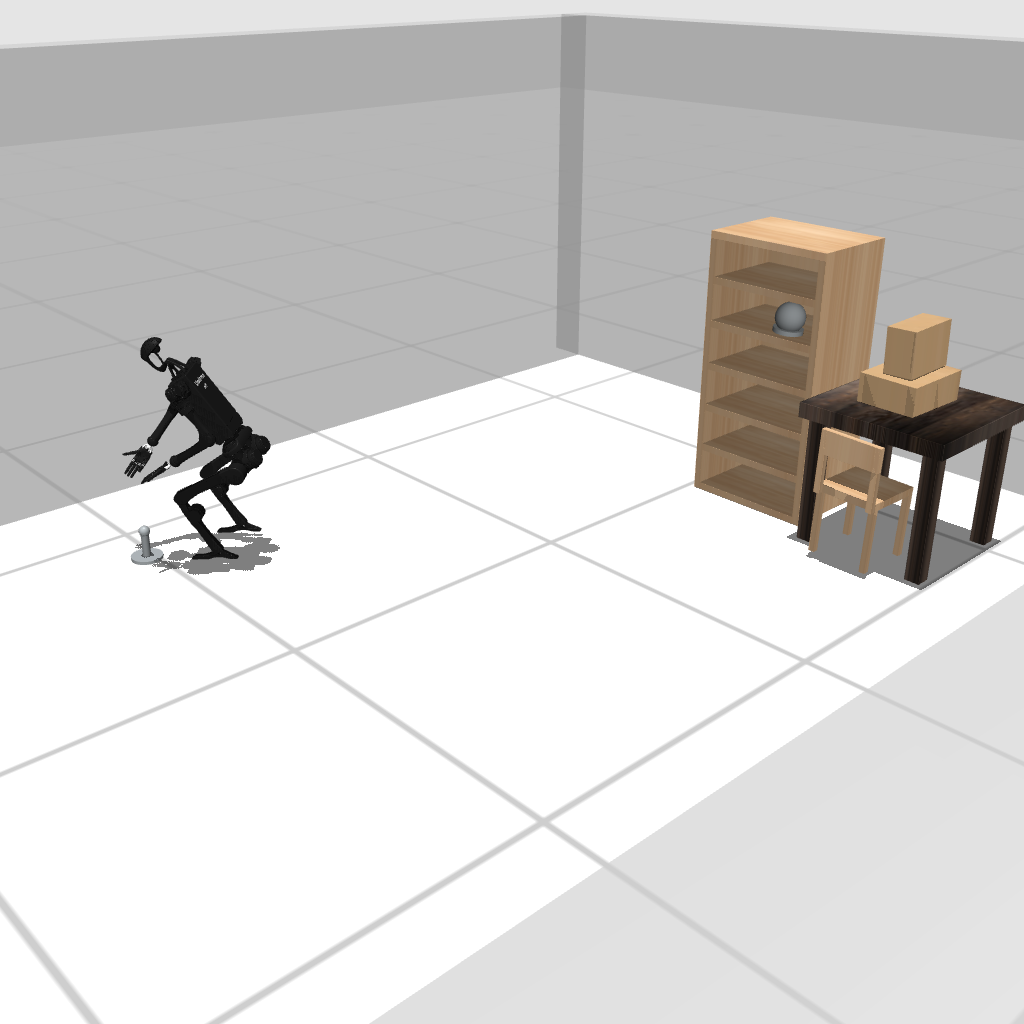}
        \vspace{-1.7em}
        \caption{\texttt{room}}
    \end{subfigure}
    \hfill
    \begin{subfigure}[ht]{0.325\textwidth}
        \centering
        \includegraphics[width=0.49\textwidth]{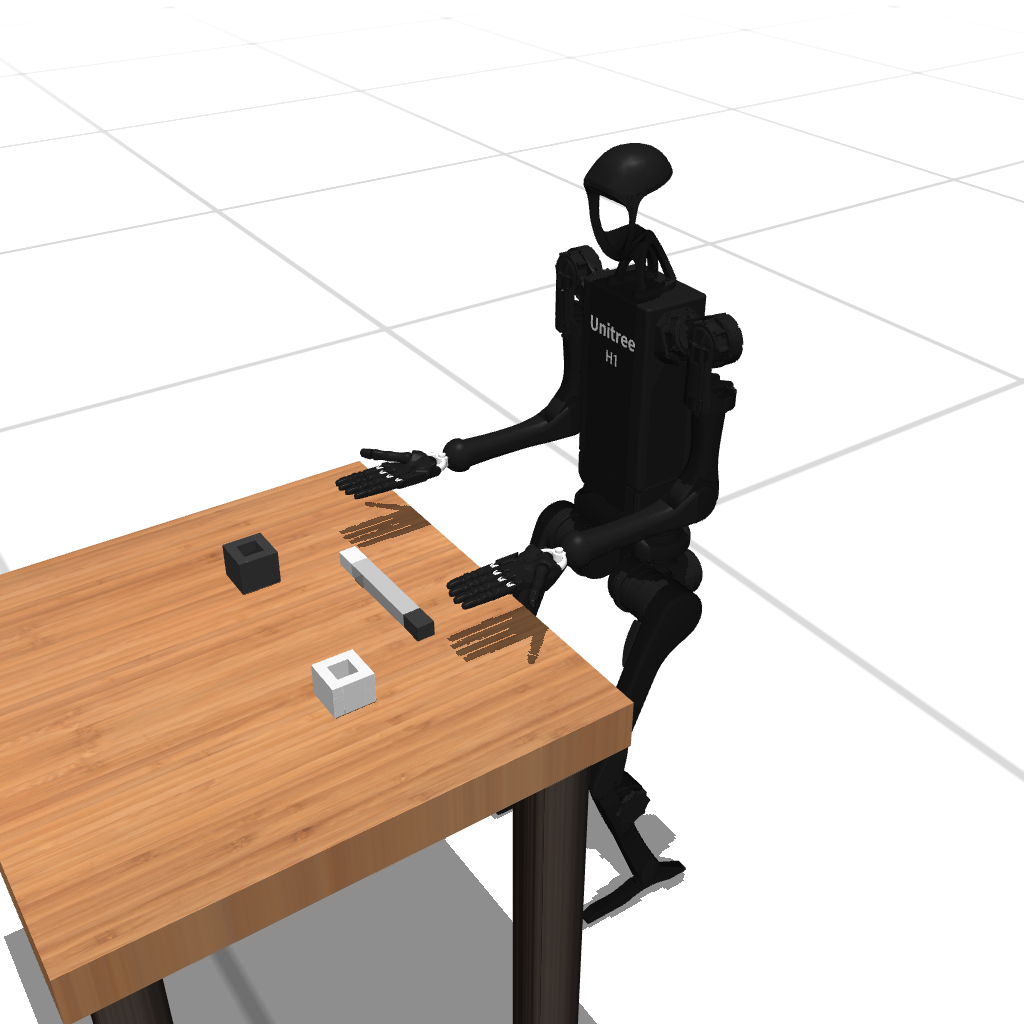}
        \includegraphics[width=0.49\textwidth]{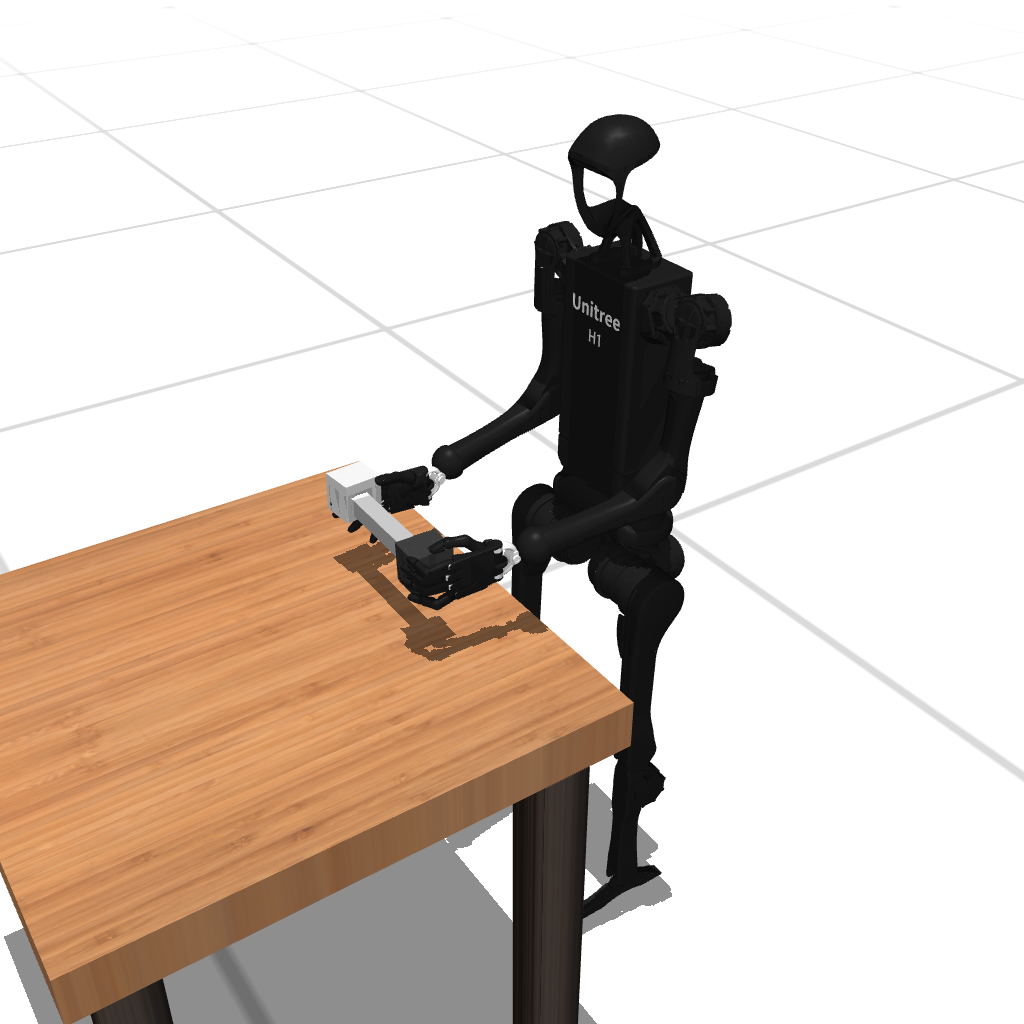}
        \vspace{-1.5em}
        \caption{\texttt{insert}}
    \end{subfigure}
    \caption{
        \textbf{HumanoidBench manipulation task suite.}
        We devise $15$ benchmarking whole-body manipulation tasks that cover a wide variety of interactions and difficulties. This figure illustrates an initial state for each task (left) and examples of the robot performing such tasks (right).
    }
    \label{fig:manipulation_task_suite}
\end{figure*}

\begin{figure*}[t]
    \centering
    \captionsetup[subfigure]{labelformat=empty}
    \begin{subfigure}[ht]{0.325\textwidth}
        \centering
        \includegraphics[width=0.49\textwidth]{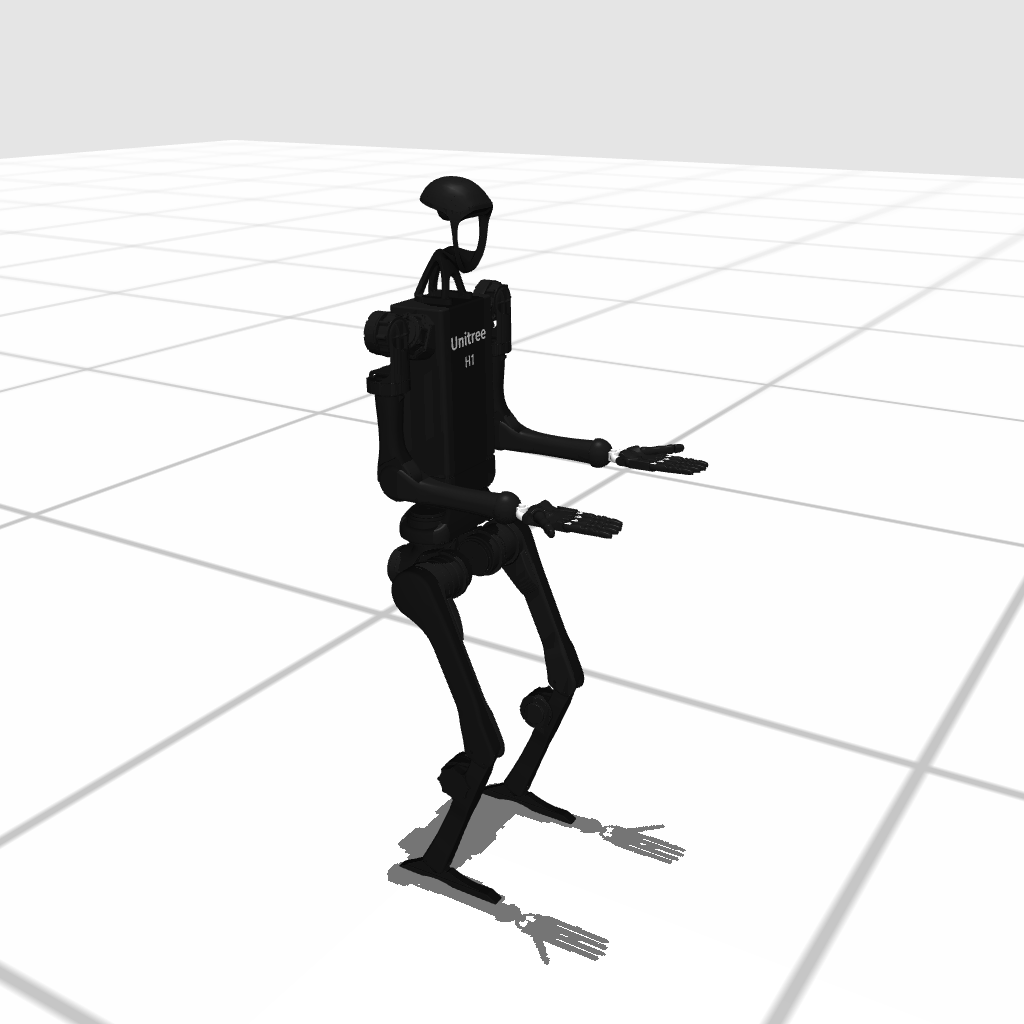}
        \includegraphics[width=0.49\textwidth]{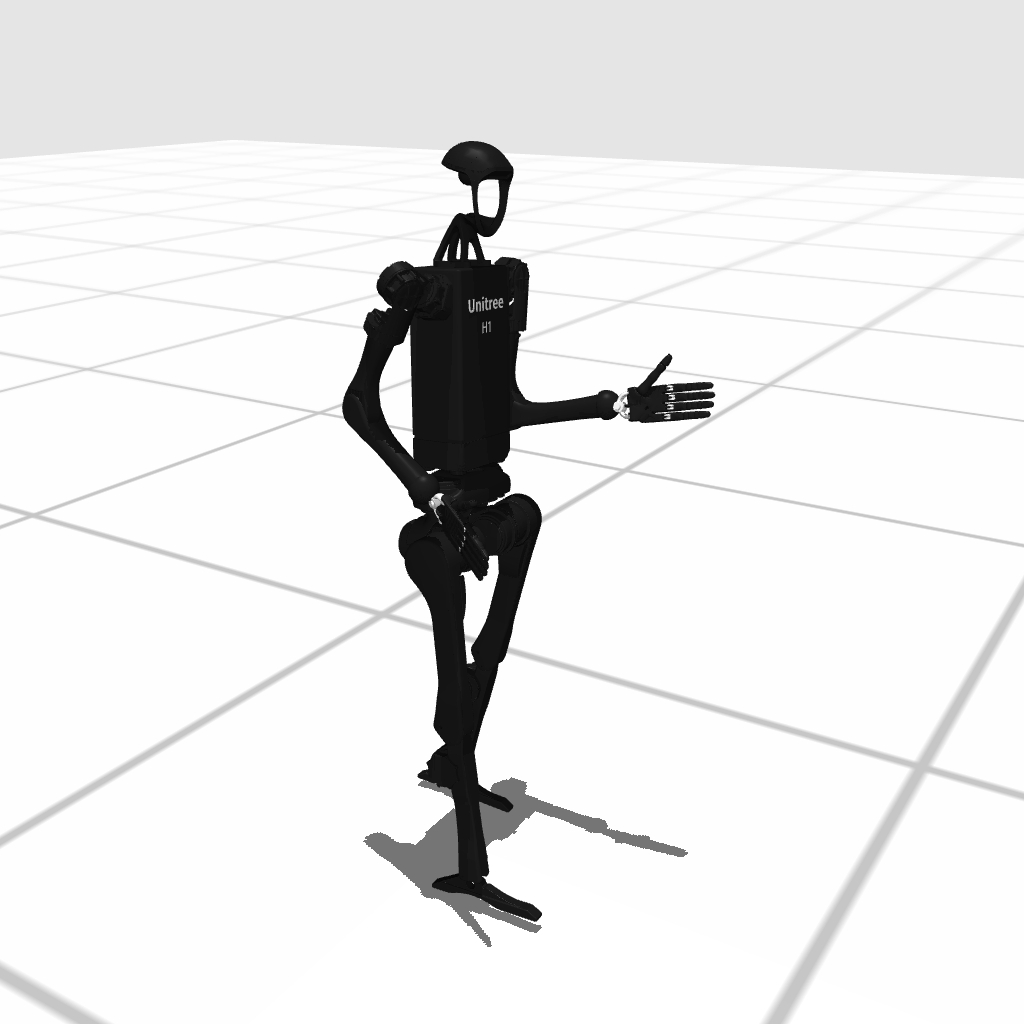}
        \vspace{-1.7em}
        \caption{\texttt{walk}}
    \end{subfigure}
    \hfill
    \begin{subfigure}[ht]{0.325\textwidth}
        \centering
        \includegraphics[width=0.49\textwidth]{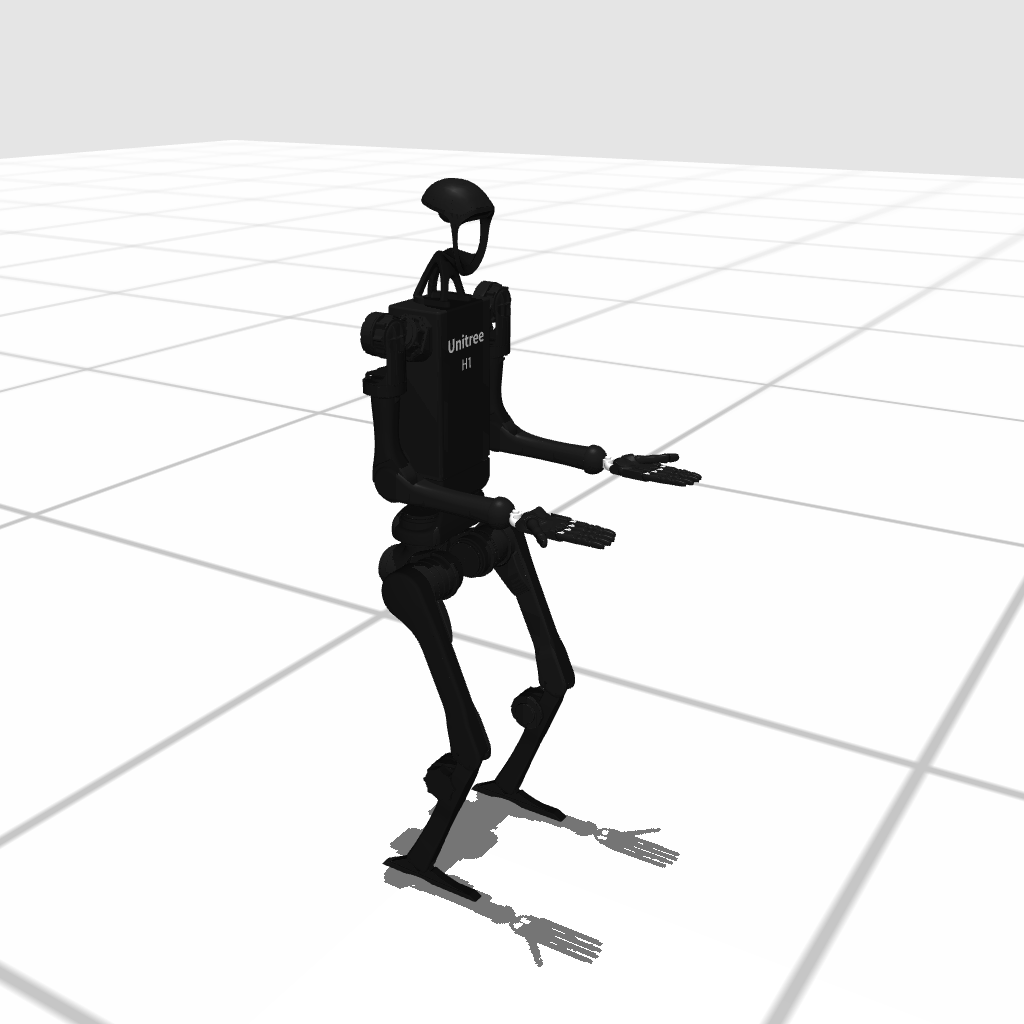}
        \includegraphics[width=0.49\textwidth]{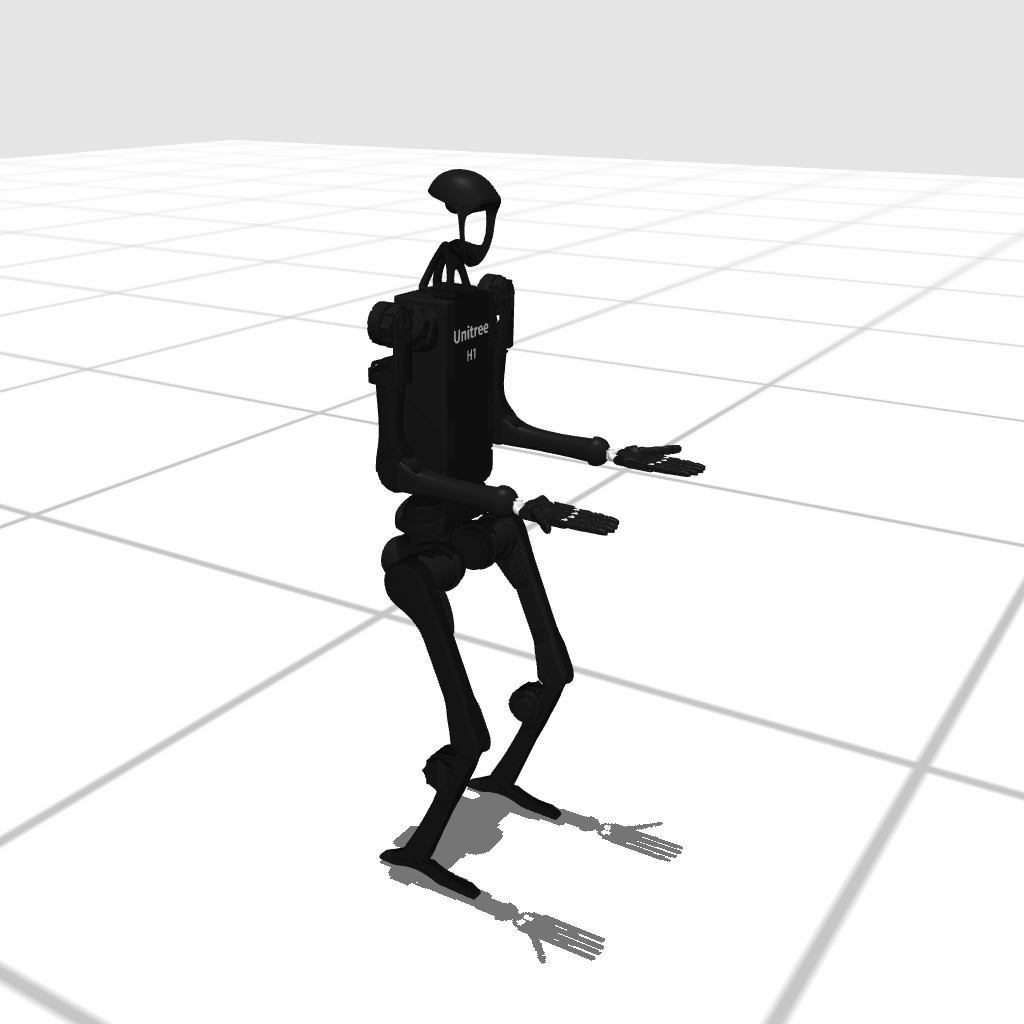}
        \vspace{-1.7em}
        \caption{\texttt{stand}}
    \end{subfigure}
    \hfill
    \begin{subfigure}[ht]{0.325\textwidth}
        \centering
        \includegraphics[width=0.49\textwidth]{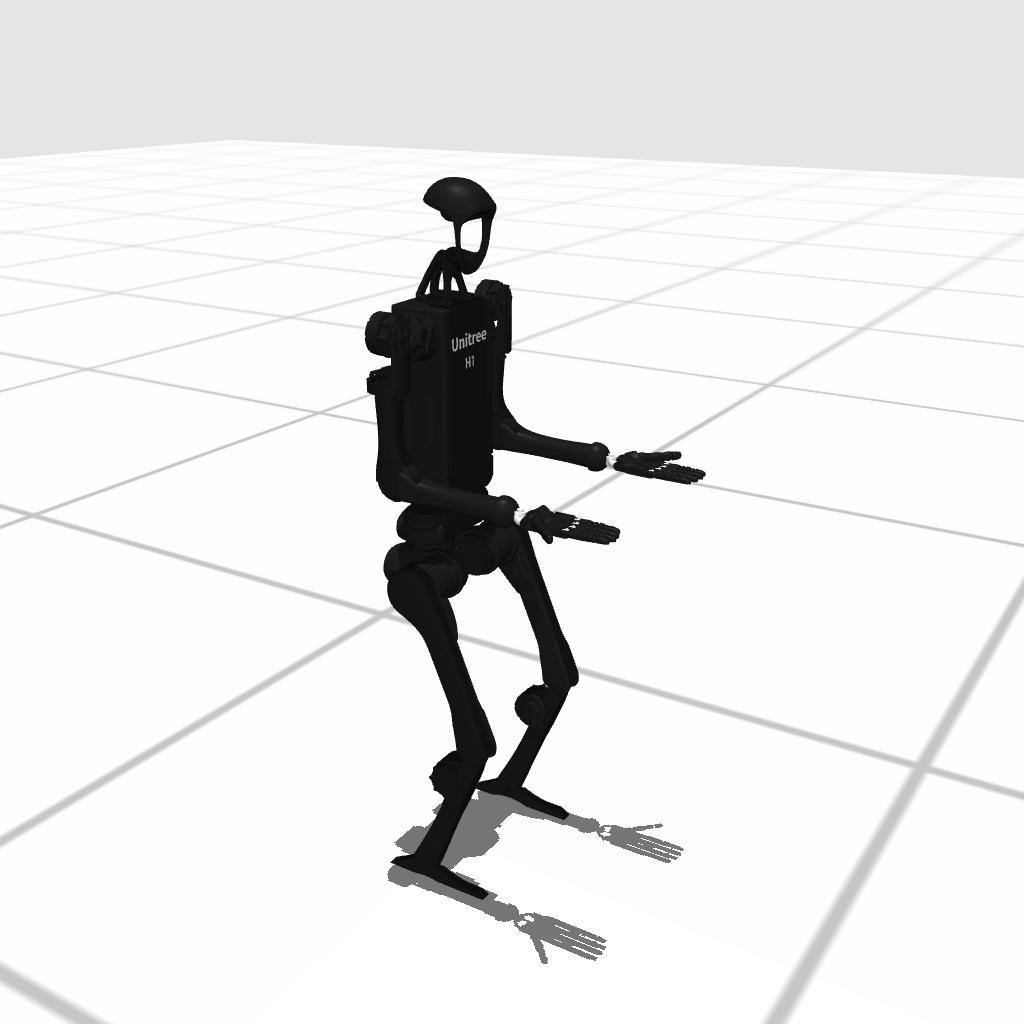}
        \includegraphics[width=0.49\textwidth]{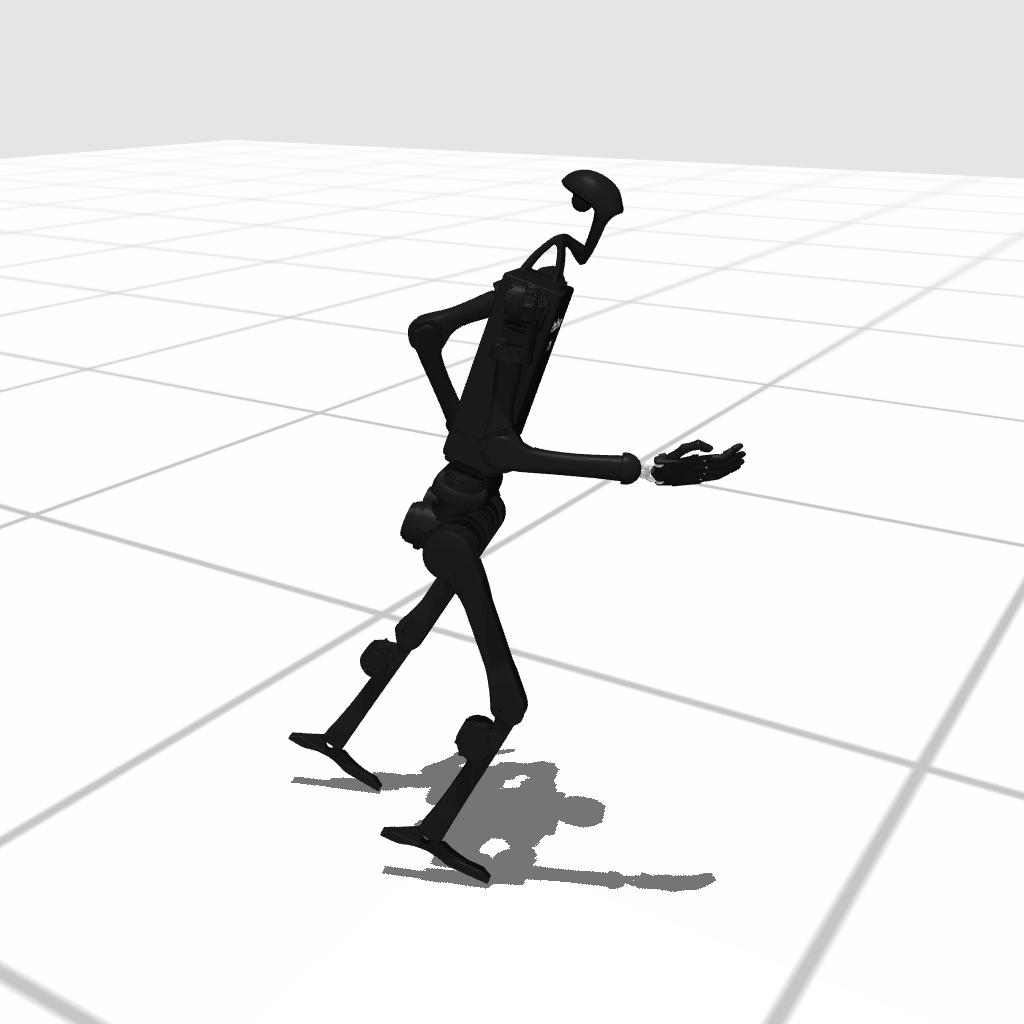}
        \vspace{-1.7em}
        \caption{\texttt{run}}
    \end{subfigure}
    \vspace{0.6em}
    \\
    \begin{subfigure}[ht]{0.325\textwidth}
        \centering
        \includegraphics[width=0.49\textwidth]{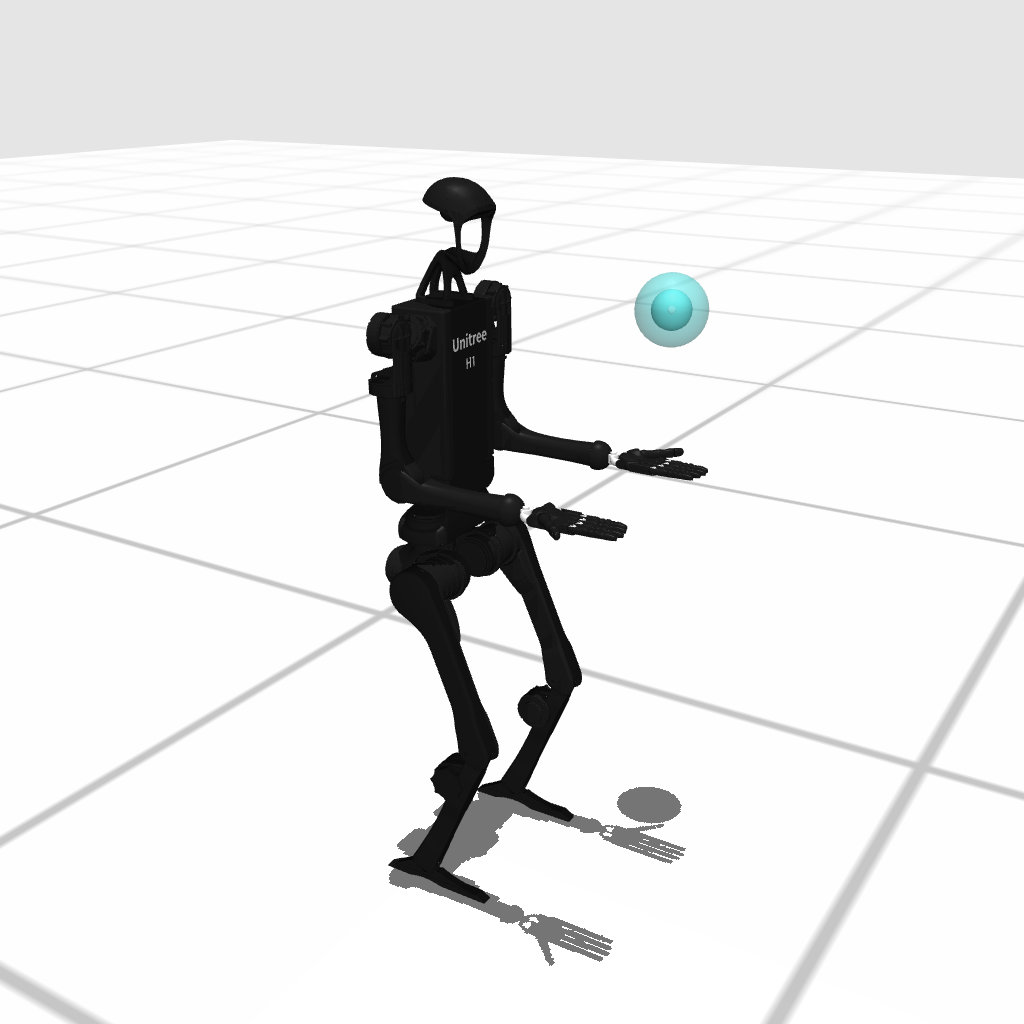}
        \includegraphics[width=0.49\textwidth]{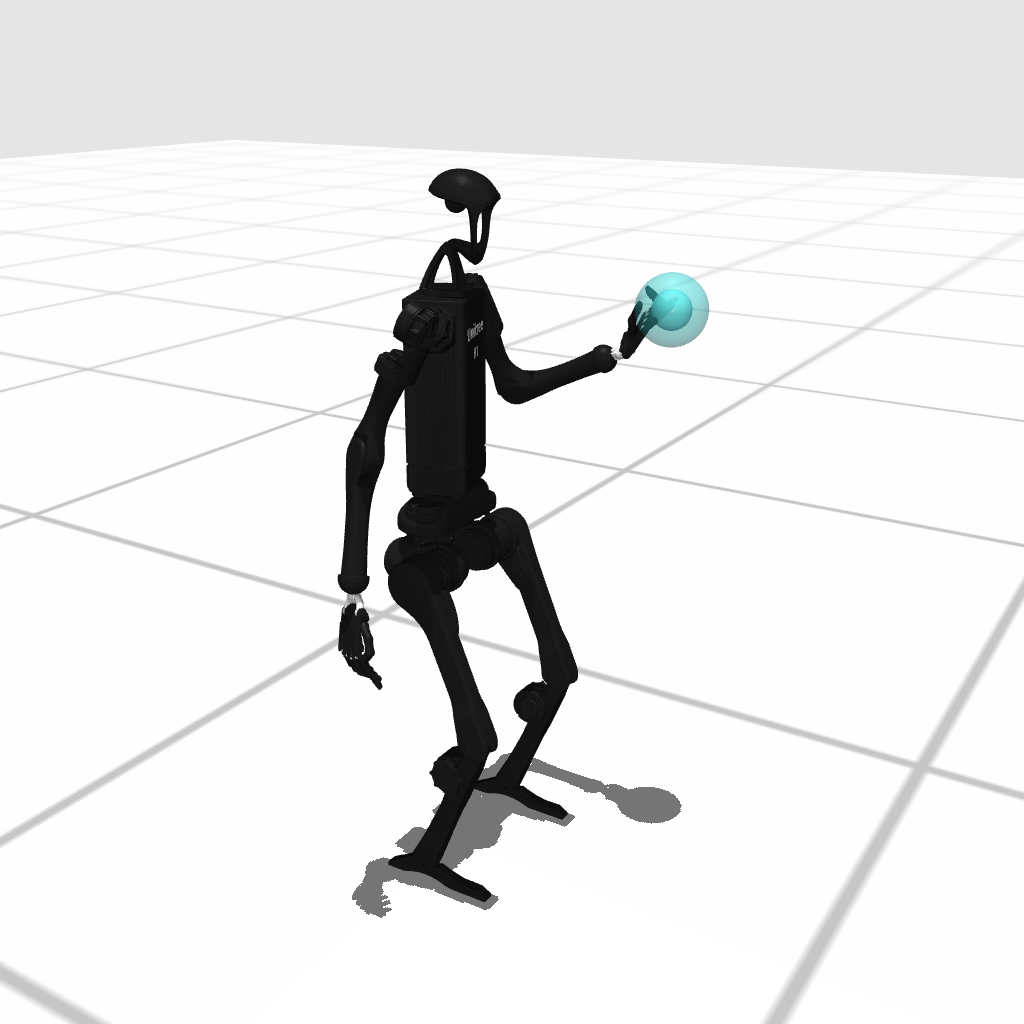}
        \vspace{-1.7em}
        \caption{\texttt{reach}}
    \end{subfigure}
    \hfill
    \begin{subfigure}[ht]{0.325\textwidth}
        \centering
        \includegraphics[width=0.49\textwidth]{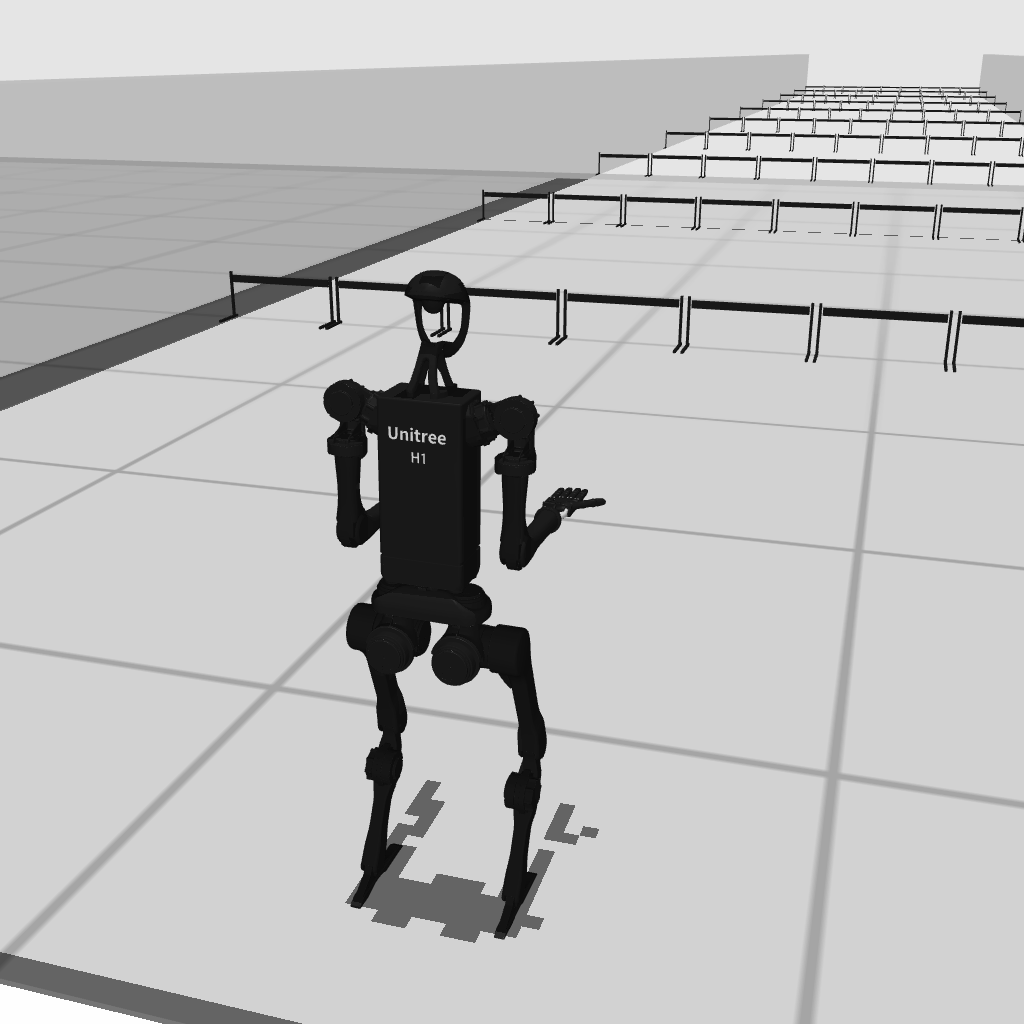}
        \includegraphics[width=0.49\textwidth]{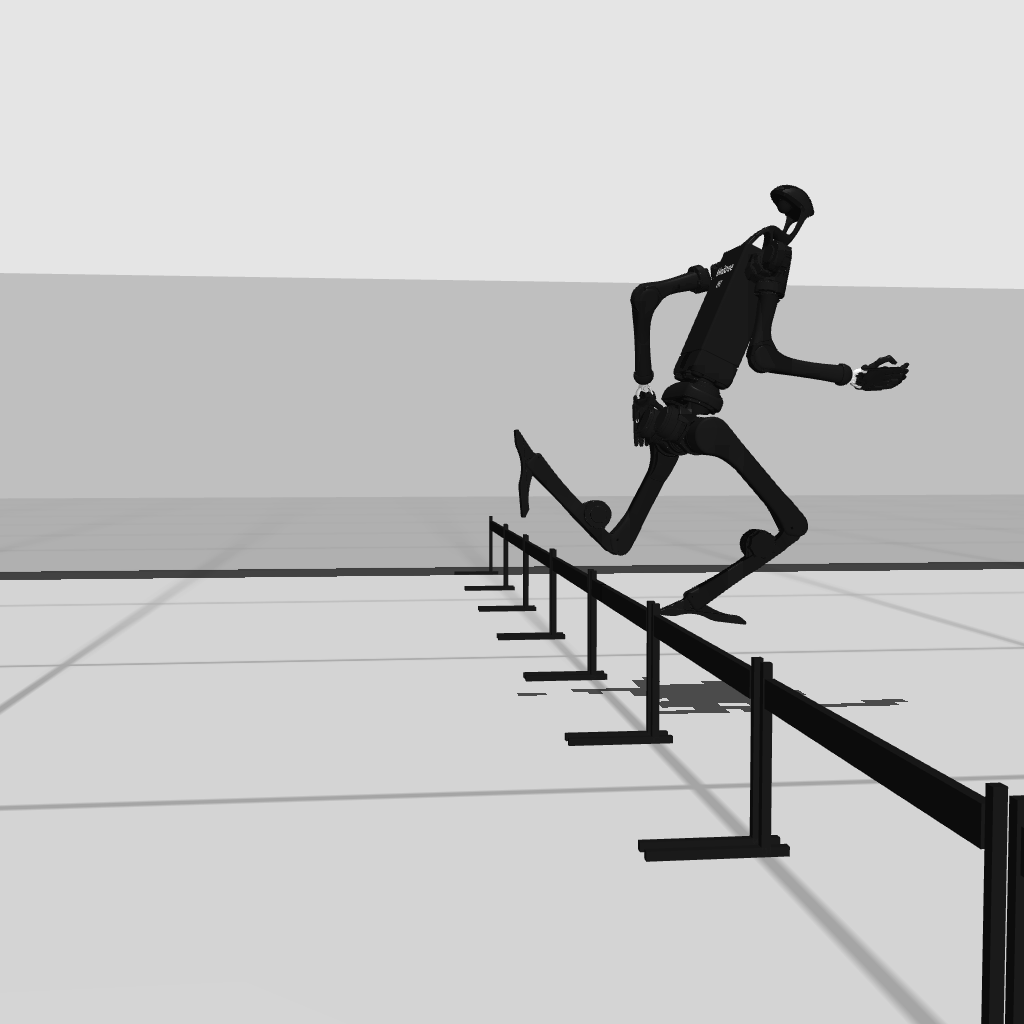}
        \vspace{-1.7em}
        \caption{\texttt{hurdle}}
    \end{subfigure}
    \hfill
    \begin{subfigure}[ht]{0.325\textwidth}
        \centering
        \includegraphics[width=0.49\textwidth]{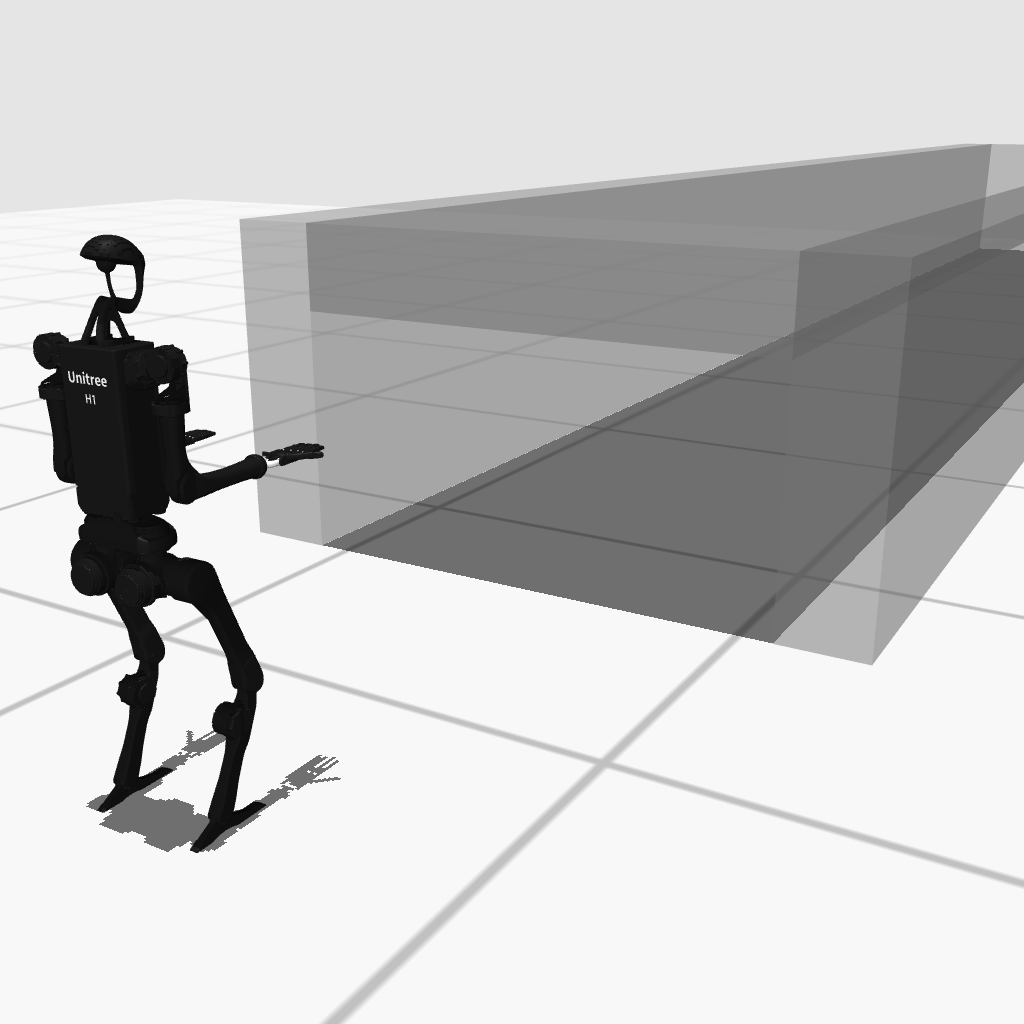}
        \includegraphics[width=0.49\textwidth]{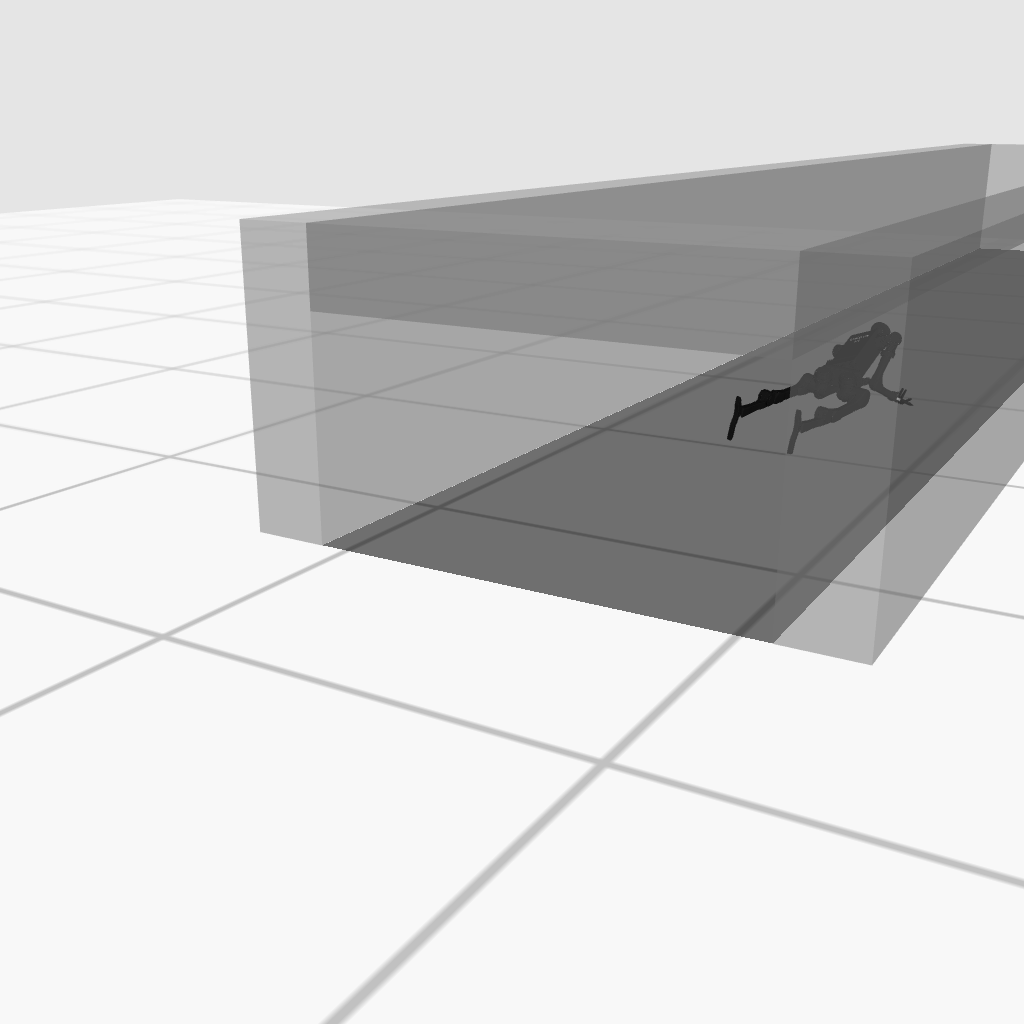}
        \vspace{-1.7em}
        \caption{\texttt{crawl}}
    \end{subfigure}
    \vspace{0.6em}
    \\
    \begin{subfigure}[ht]{0.325\textwidth}
        \centering
        \includegraphics[width=0.49\textwidth]{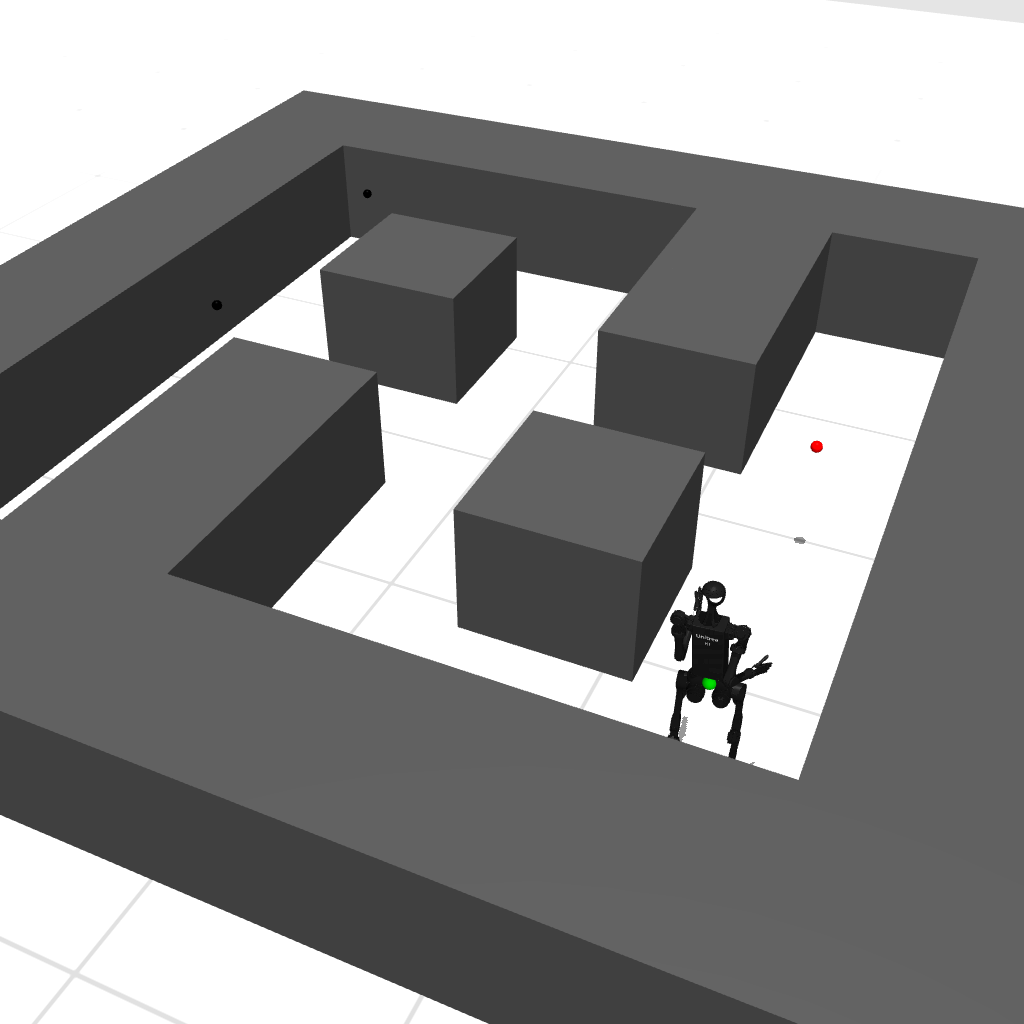}
        \includegraphics[width=0.49\textwidth]{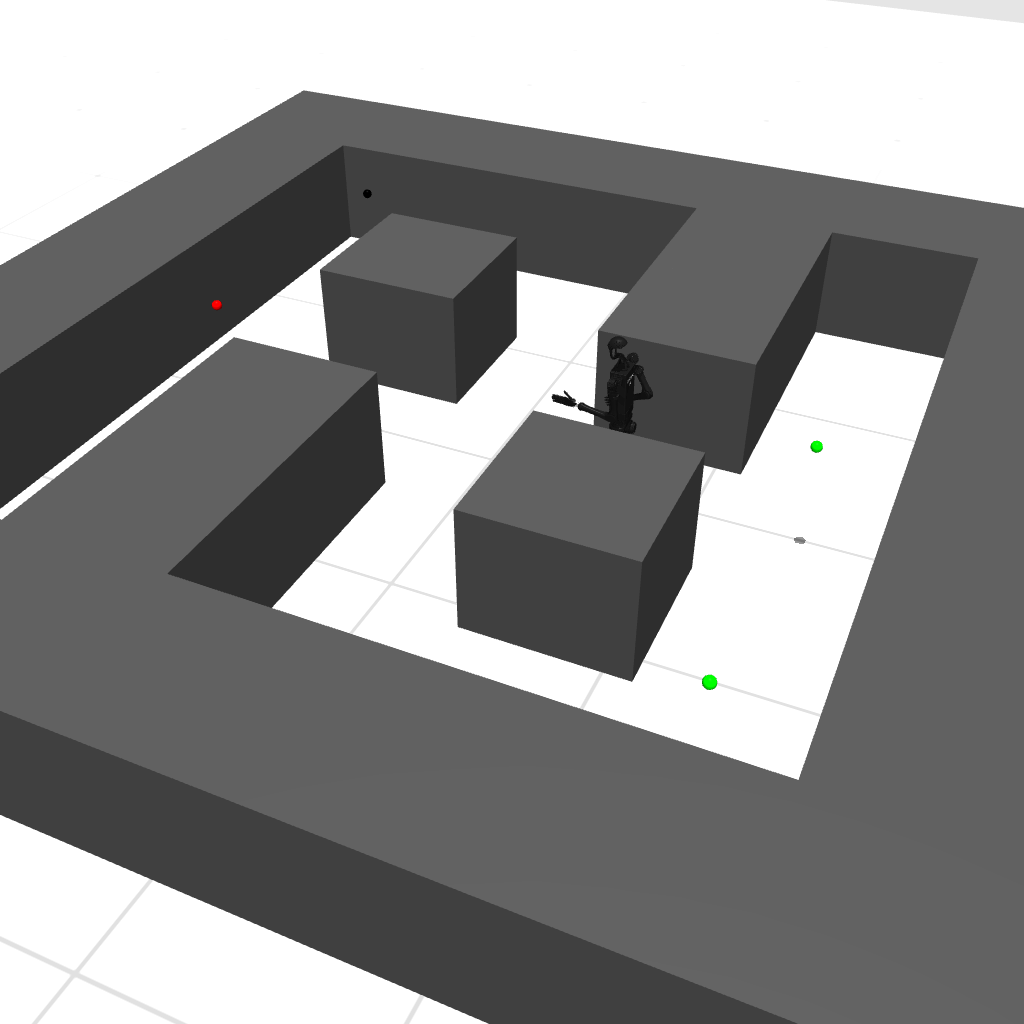}
        \vspace{-1.7em}
        \caption{\texttt{maze}}
    \end{subfigure}
    \hfill
    \begin{subfigure}[ht]{0.325\textwidth}
        \centering
        \includegraphics[width=0.49\textwidth]{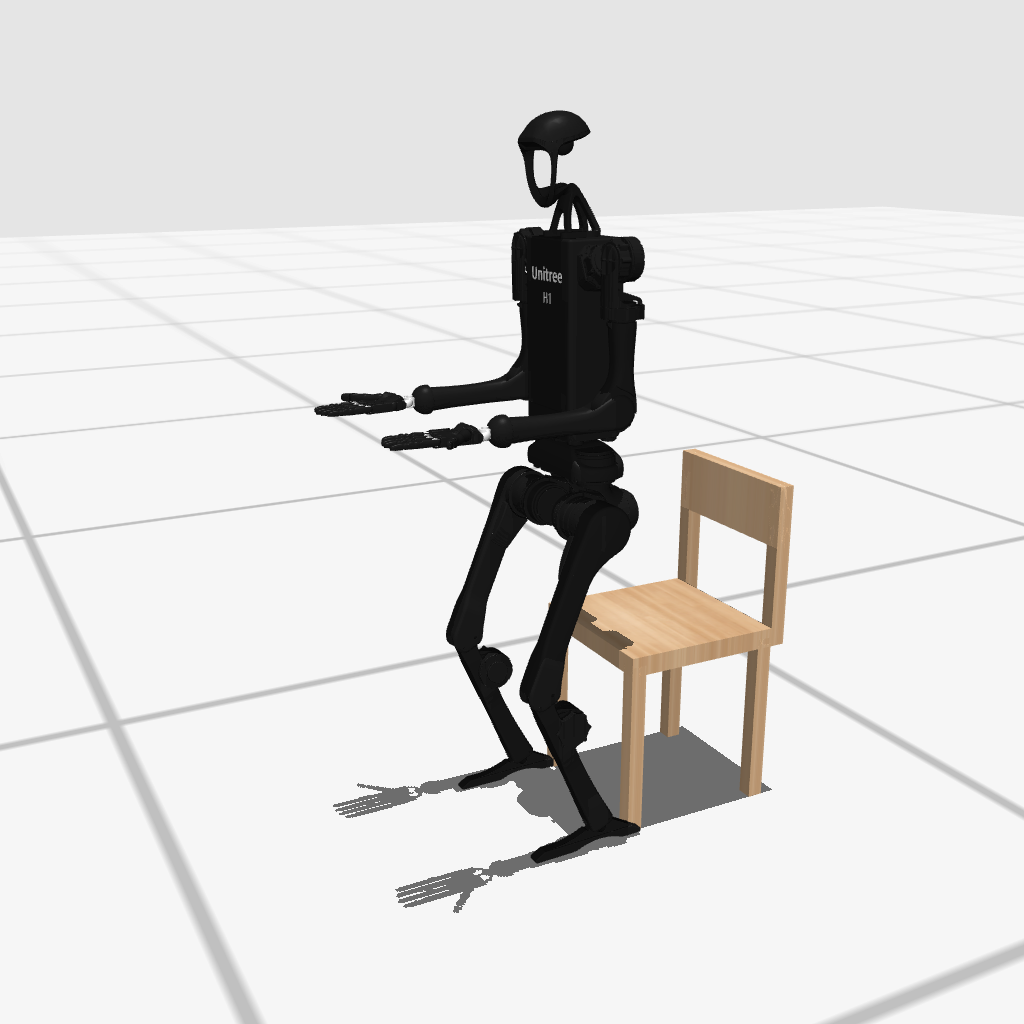}
        \includegraphics[width=0.49\textwidth]{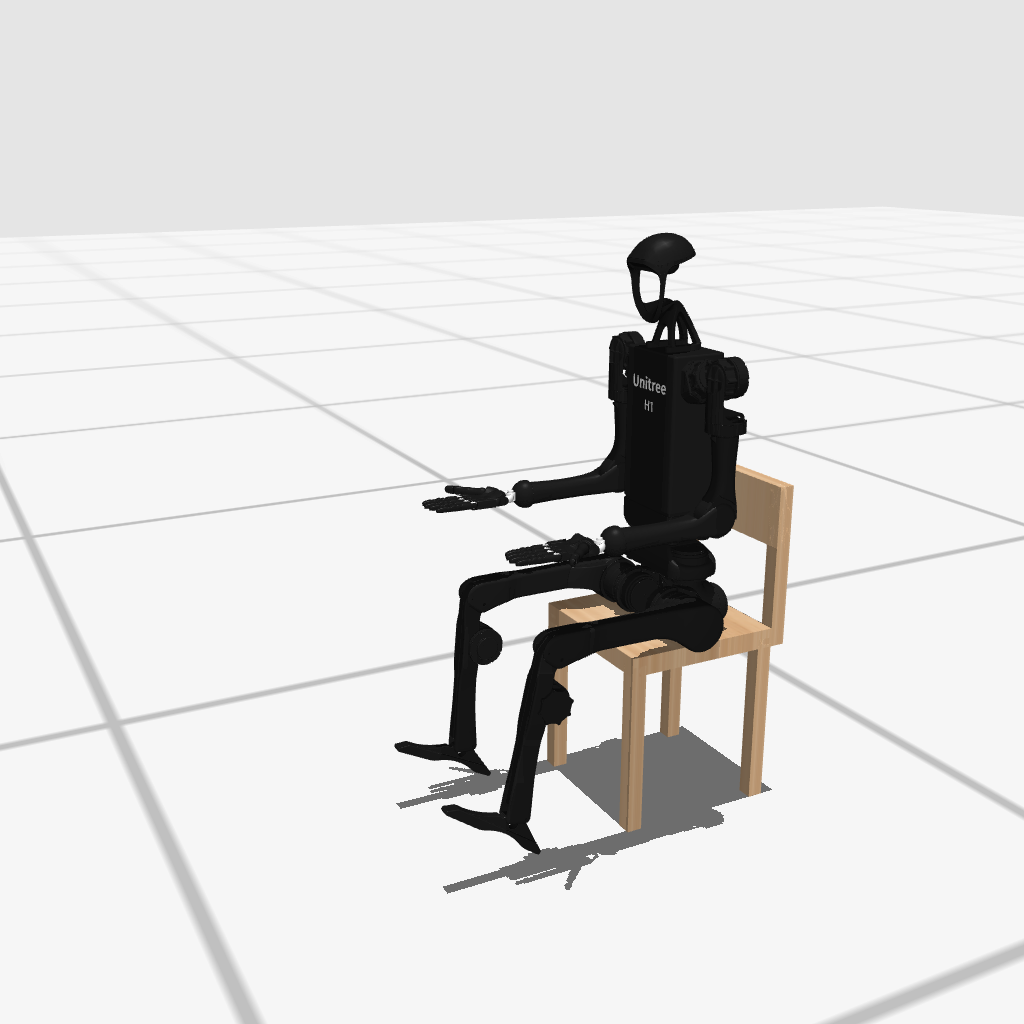}
        \vspace{-1.7em}
        \caption{\texttt{sit}}
    \end{subfigure}
    \hfill
    \begin{subfigure}[ht]{0.325\textwidth}
        \centering
        \includegraphics[width=0.49\textwidth]{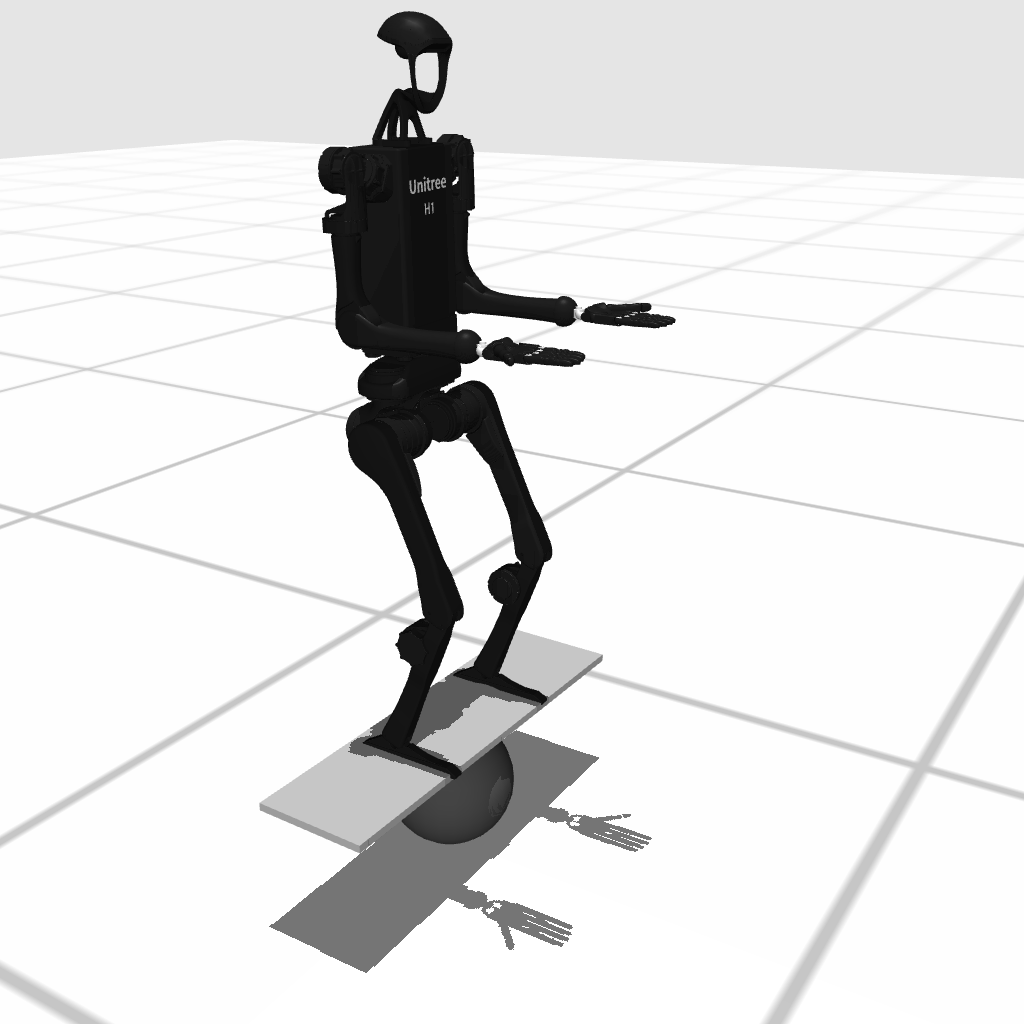}
        \includegraphics[width=0.49\textwidth]{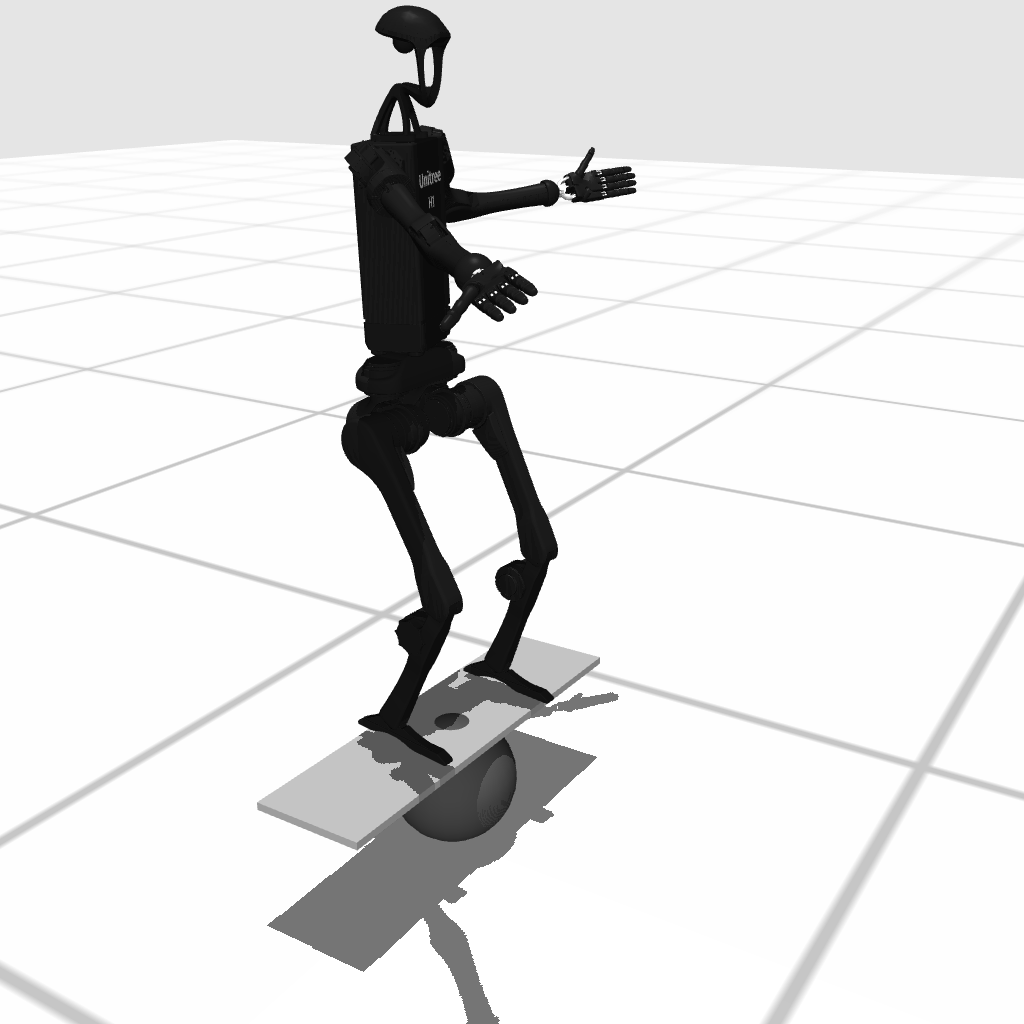}
        \vspace{-1.5em}
        \caption{\texttt{balance}}
    \end{subfigure}
    \vspace{0.6em}
    \\
    \begin{subfigure}[ht]{0.325\textwidth}
        \centering
        \includegraphics[width=0.49\textwidth]{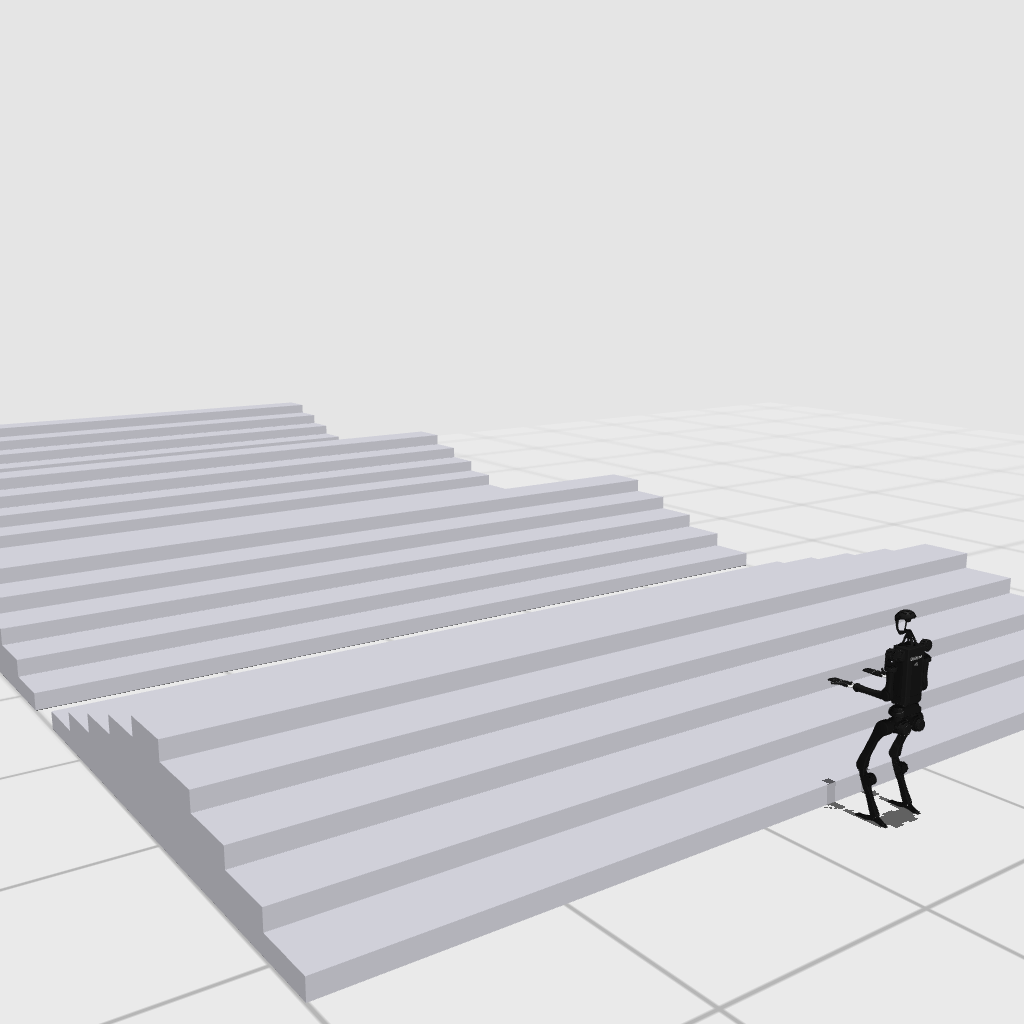}
        \includegraphics[width=0.49\textwidth]{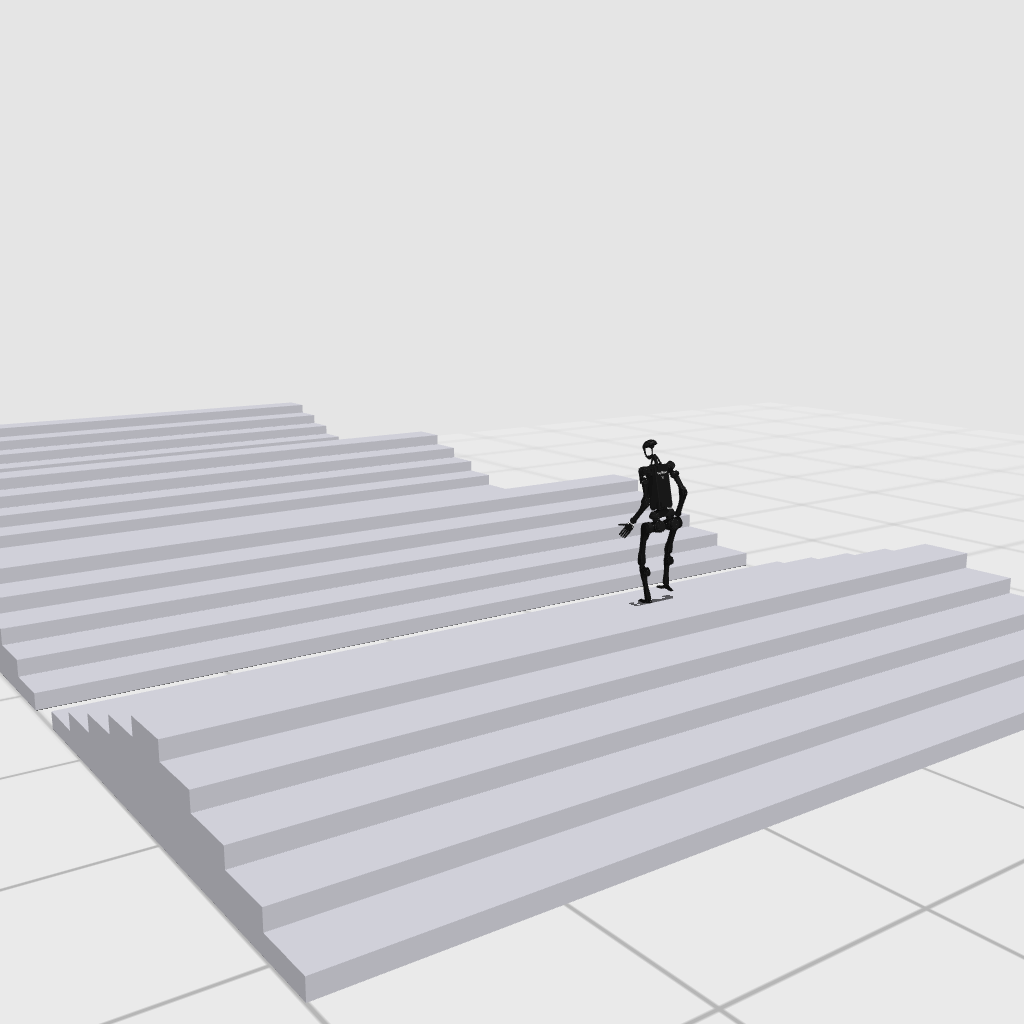}
        \vspace{-1.5em}
        \caption{\texttt{stairs}}
    \end{subfigure}
    \hfill
    \begin{subfigure}[ht]{0.325\textwidth}
        \centering
        \includegraphics[width=0.49\textwidth]{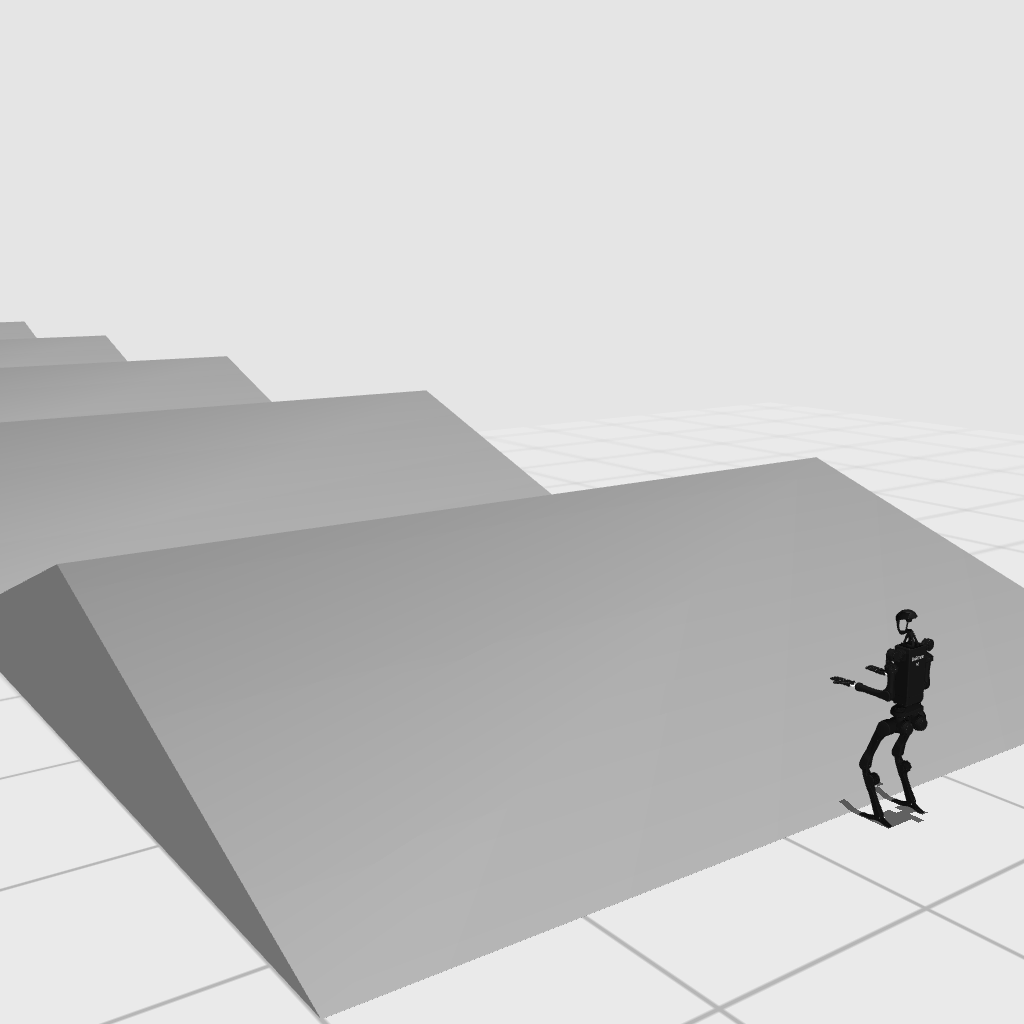}
        \includegraphics[width=0.49\textwidth]{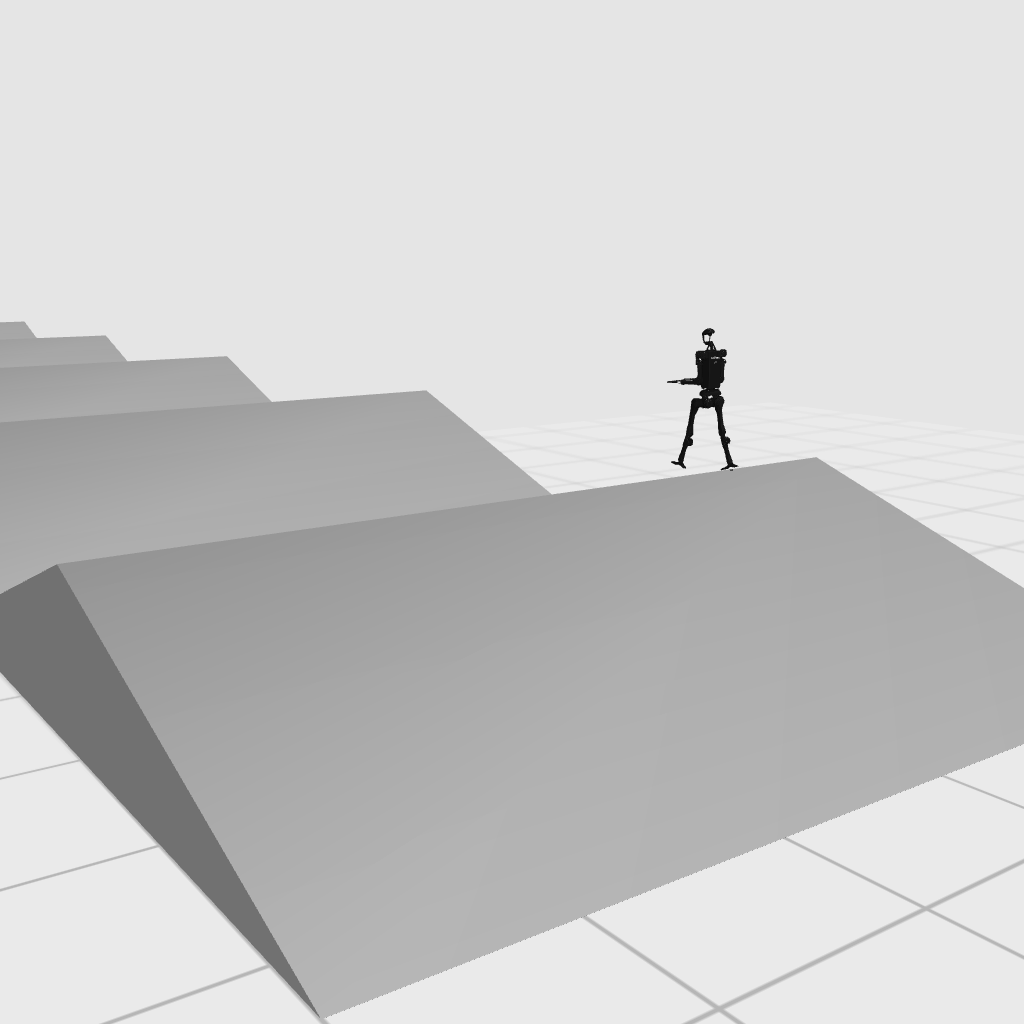}
        \vspace{-1.5em}
        \caption{\texttt{slides}}
    \end{subfigure}
    \hfill
    \begin{subfigure}[ht]{0.325\textwidth}
        \centering
        \includegraphics[width=0.49\textwidth]{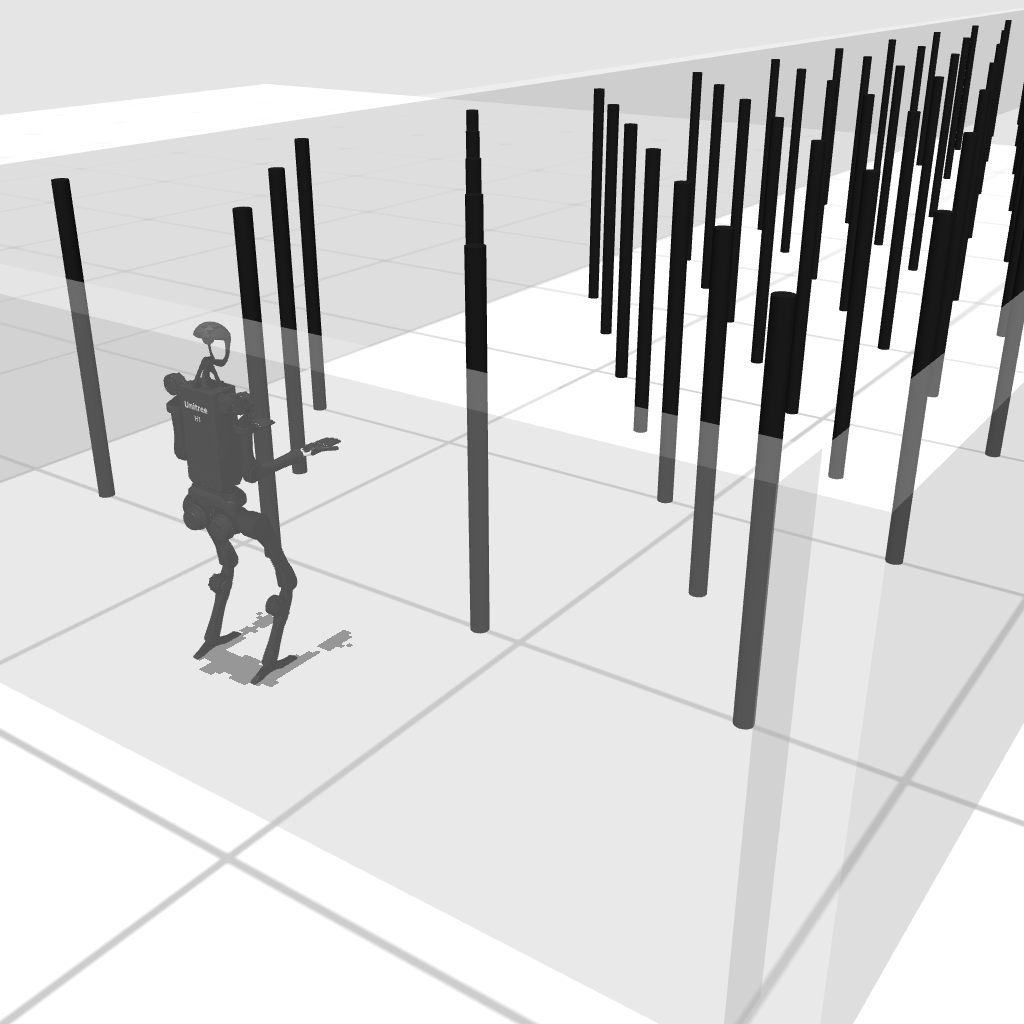}
        \includegraphics[width=0.49\textwidth]{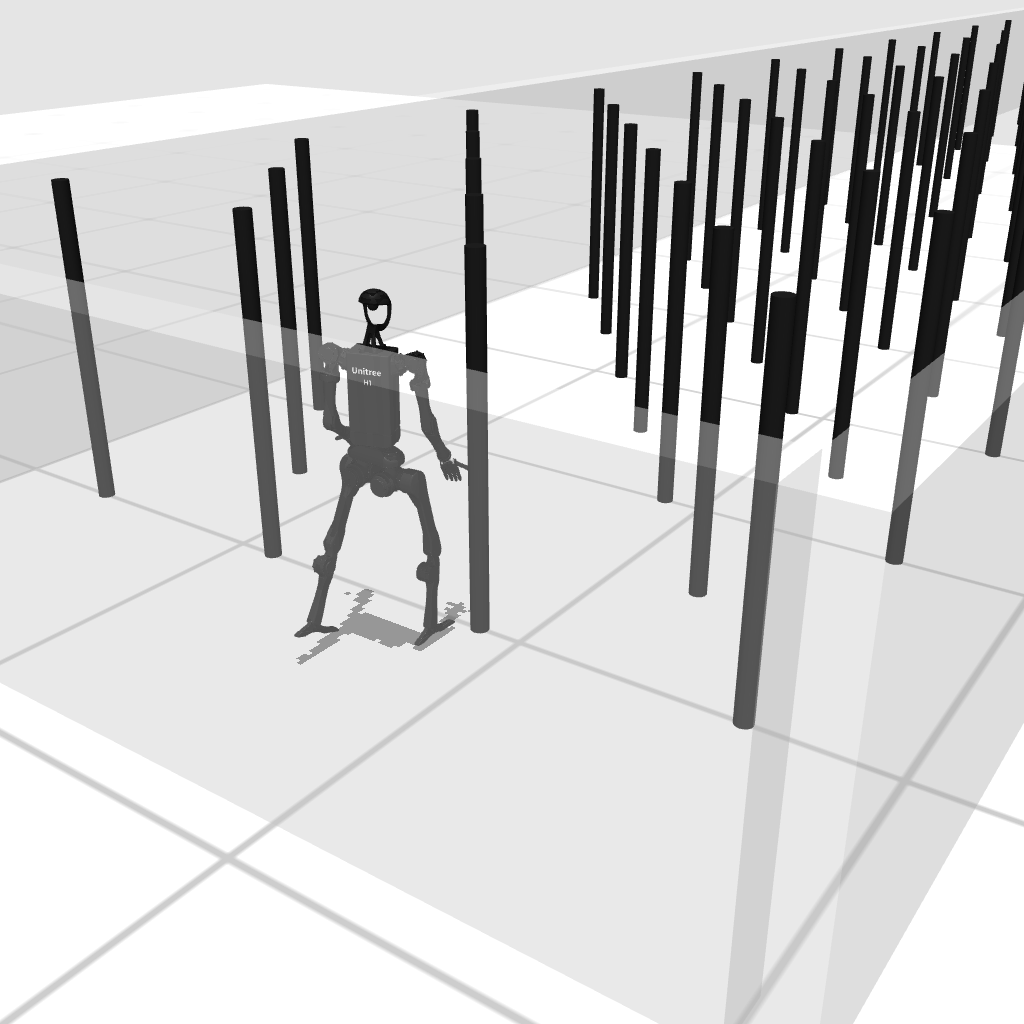}
        \vspace{-1.5em}
        \caption{\texttt{pole}}
    \end{subfigure}
    \caption{
        \textbf{HumanoidBench locomotion task suite.} We devise $12$ benchmarking locomotion tasks that cover a wide variety of interactions and difficulties. This figure illustrates an initial state for each task (left) and examples of the robot performing such tasks (right).
    }
    \label{fig:locomotion_task_suite}
\end{figure*}

\section{Simulated Humanoid Robot Environment}
\label{sec:environment}

In this section, we describe our simulated environment and discuss relevant design choices for the simulated humanoid robot. As illustrated in \Cref{fig:humanoid_env}, we use the Unitree H1 humanoid robot\cref{footnote:h1} with two dexterous Shadow Hands\cref{footnote:shadowhand} as the primary robotic agent of our benchmark. We simulate this humanoid robot using MuJoCo~\citep{todorov2012mujoco} adapting the Unitree H1 model provided by Unitree\footnote{\label{footnote:h1_github}\url{https://github.com/unitreerobotics/unitree_ros}} and the dexterous Shadow Hand models available through MuJoCo Menagerie.\footnote{\label{footnote:menagerie_github}\url{https://github.com/google-deepmind/mujoco_menagerie}}

\textbf{Humanoid Body.}\quad
We implement Unitree H1\cref{footnote:h1}, Unitree G1\cref{footnote:g1}, and Agility Robotics Digit\cref{footnote:digit}, which are well-known humanoid robots with their model files freely available~\citep{menagerie2022github, adu2023exploring}. Unitree H1 is primarily used in our benchmark as it is a full-size humanoid compared to the smaller Unitree G1, and as we observed faster learning compared to the Agility Robotics Digit, which we ascribe to a simpler mechanical design compared to Digit, which features passive joints actuated through a four-bar linkage. 

\textbf{Dexterous Hands.}\quad 
We use two dexterous Shadow Hands\cref{footnote:shadowhand}, which also have model files freely available\cref{footnote:menagerie_github}, and have shown impressive manipulation capabilities both in simulation \cite{zakka2023robopianist} and in the real world \cite{akkaya2019solving}. To make the simulated robot have more human-like morphology, we remove the cumbersome forearms of the dexterous Shadow Hands in HumanoidBench. While this is not currently a realistic model, we anticipate the trend in the industry towards developing slimmer, human-like hands (e.g., Tesla Optimus, Figure 01) so that our design choice aligns better with next-generation humanoid robots.
In addition, we also provide models for the Robotiq 2F-85 parallel-jaw gripper and the $13$-DoF Unitree hands available in the Unitree collection\cref{footnote:h1_github} (see Appendix, \Cref{sec:additional_tasks} for more details).

The observation and action spaces, and degrees of freedom of the robot system with or without the dexterous hands are summarized in \Cref{tab:robot_spec}.

\begin{table}[t]
    \centering
    \begin{tabular}{ccc}
        \toprule
                          & Without hand   & With $2$ hands    \\
        \midrule
        Observation space & $51$              & $151$             \\
        Action space      & $19$              & $61$              \\
        DoF (body)        & $25$              & $25$              \\
        DoF (two hands)   & $0$               & $50$              \\
        \bottomrule
    \end{tabular}
    \caption{\textbf{Humanoid robot specifications with and without hands.} Both the humanoid body (including its floating base) and one Shadow Hand present action spaces ($19$ and $21$, respectively) smaller than their DoFs ($25$), making them underactuated systems. In this table, the observation spaces solely comprise generalized positions and velocities of the robots and do not take into account any environment observations. We use quaternions for the robot floating base orientation, which adds an additional position coordinate compared to the velocity components, which match the DoFs. In the appendix, \Cref{tab:add_robot_spec} shows an exhaustive overview of all the robot configurations available in HumanoidBench.}
    \label{tab:robot_spec}
\end{table}

\textbf{Observations.}\quad
Our simulated environment supports the following observations:  
\begin{itemize}
    \item Proprioceptive robot state (i.e., joint angles and velocities) and task-relevant environment observations (i.e., object poses and velocities). 
    \item Egocentric visual observations from two cameras placed on the robot head (see \Cref{fig:humanoid_env}).
    \item Whole-body tactile sensing using the MuJoCo tactile grid sensor (see \Cref{fig:humanoid_env}). We design tactile sensing at the hands with high resolution and in other body parts with low resolution, similar to humans, with a total of $448$ taxels spread over the entire body, each providing three-dimensional contact force readings. Similar distributed force readings have been captured on real-world systems both on humanoid bodies \cite{mittendorfer2011humanoid} and end-effectors \citep{sferrazza2022sim}. The implementation of such spatially distributed contact sensing required non-trivial mesh adaptations and refinements, which we detail in Appendix, \Cref{sec:environment_details:tactile}.
\end{itemize}

Although other sensory inputs are available from the environment, to investigate challenges in whole-body control of humanoid robots, we first focus on the state-based environment setup, where proprioceptive robot states and object states are used as the agent's input in HumanoidBench. In our state-based environment, we maintain the robot observations the same across tasks to minimize domain knowledge, in contrast to tailoring it to the specific tasks~\cite{tassa2020dm_control}. We leave extending our environment to  benchmarking multimodal perception capabilities \cite{yuan2023robot, sferrazza2023power} of humanoid robots as future work.

\textbf{Actions.}\quad
In HumanoidBench, the humanoid robot is controlled via position control (i.e., specifying the target joint positions). Torque-based control is also supported but we found that position control is generally more stable and allows for lower control frequency than torque control. For both position and torque control, the action space is $61$-dimensional including the two hands, and controlled at \SI{50}{\Hz}.

\begin{figure*}
    \centering
    \includegraphics[width=0.45\textwidth]{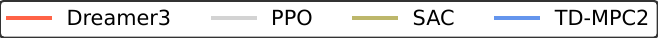}
    \vspace{0.5em}
    \\
    \begin{subfigure}[t]{0.24\textwidth}
        \includegraphics[width=\textwidth]{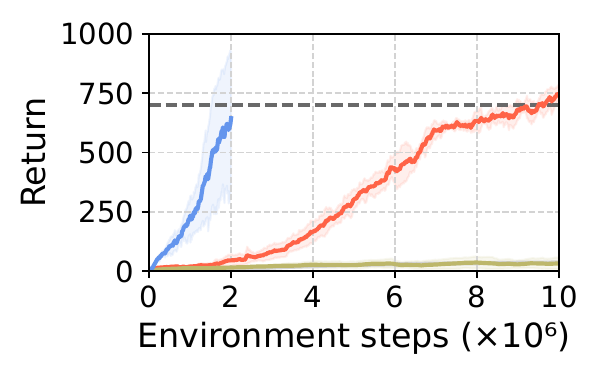}
        \vspace{-1.5em}
        \caption{\texttt{walk}}
    \end{subfigure}
    \hfill
    \begin{subfigure}[t]{0.24\textwidth}
        \includegraphics[width=\textwidth]{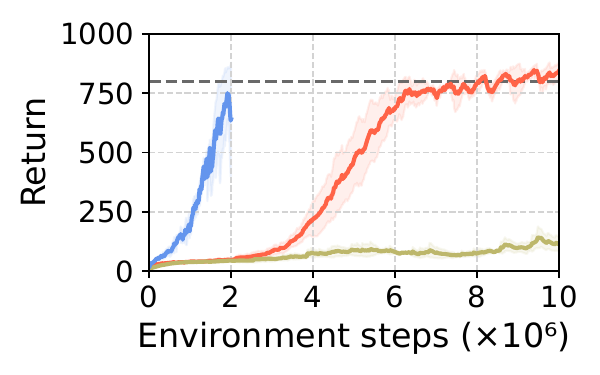}
        \vspace{-1.5em}
        \caption{\texttt{stand}}
    \end{subfigure}
    \hfill
    \begin{subfigure}[t]{0.24\textwidth}
        \includegraphics[width=\textwidth]{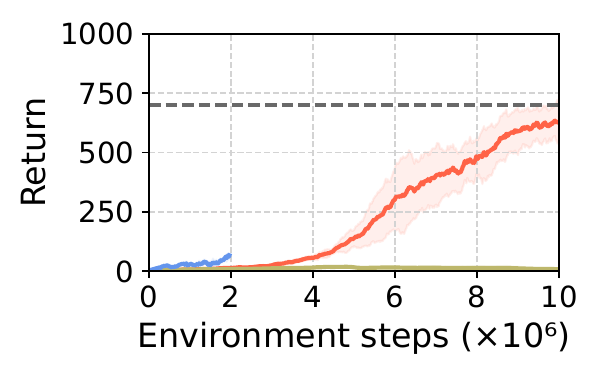}
        \vspace{-1.5em}
        \caption{\texttt{run}}
    \end{subfigure}
    \hfill
    \begin{subfigure}[t]{0.24\textwidth}
        \includegraphics[width=\textwidth]{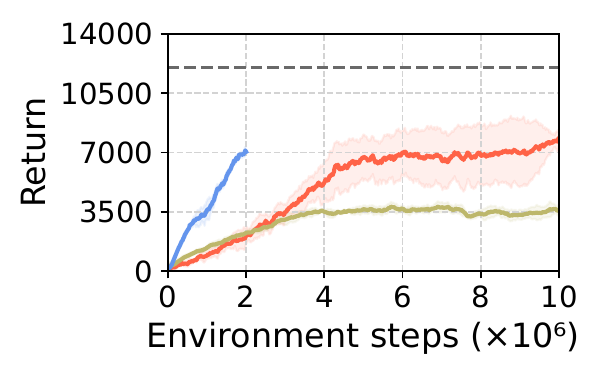}
        \vspace{-1.5em}
        \caption{\texttt{reach}}
    \end{subfigure}
    \\
    \begin{subfigure}[t]{0.24\textwidth}
        \includegraphics[width=\textwidth]{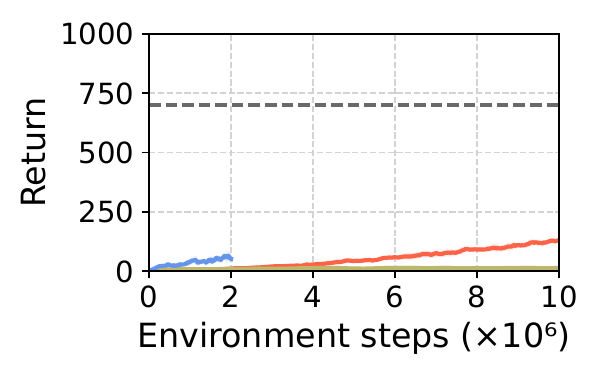}
        \vspace{-1.5em}
        \caption{\texttt{hurdle}}
    \end{subfigure}
    \hfill
    \begin{subfigure}[t]{0.24\textwidth}
        \includegraphics[width=\textwidth]{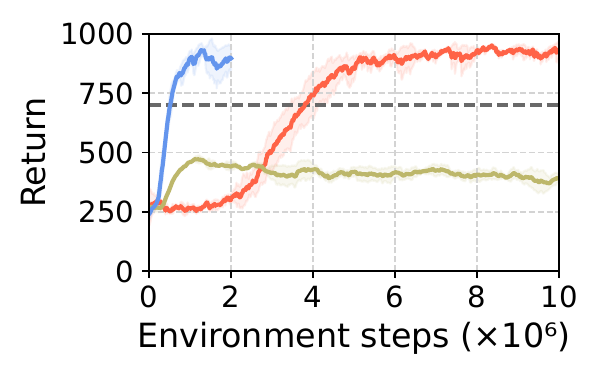}
        \vspace{-1.5em}
        \caption{\texttt{crawl}}
    \end{subfigure}
    \hfill
    \begin{subfigure}[t]{0.24\textwidth}
        \includegraphics[width=\textwidth]{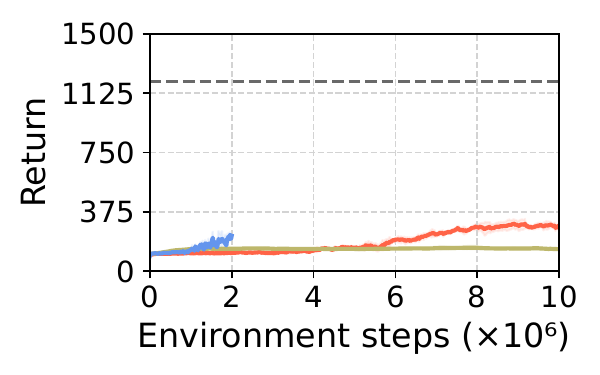}
        \vspace{-1.5em}
        \caption{\texttt{maze}}
    \end{subfigure}
    \hfill
    \begin{subfigure}[t]{0.24\textwidth}
        \includegraphics[width=\textwidth]{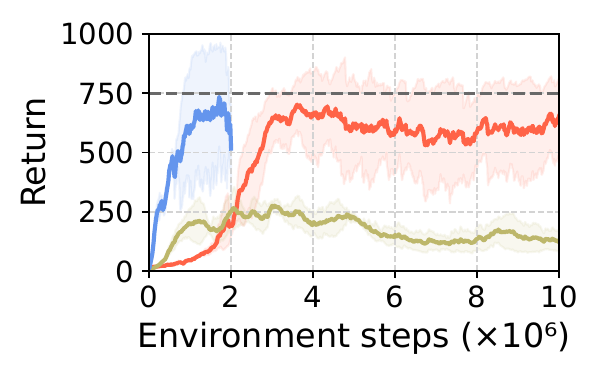}
        \vspace{-1.5em}
        \caption{\texttt{sit\_simple}}
    \end{subfigure}
    \\    
    \begin{subfigure}[t]{0.24\textwidth}
        \includegraphics[width=\textwidth]{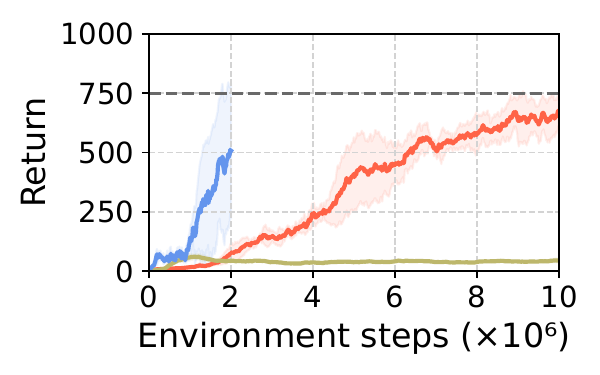}
        \vspace{-1.5em}
        \caption{\texttt{sit\_hard}}
    \end{subfigure}
    \hfill
    \begin{subfigure}[t]{0.24\textwidth}
        \includegraphics[width=\textwidth]{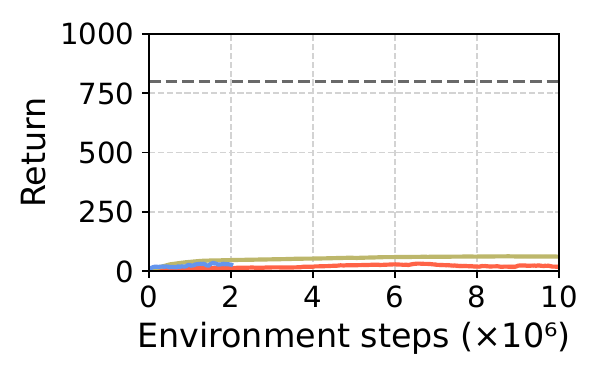}
        \vspace{-1.5em}
        \caption{\texttt{balance\_simple}}
    \end{subfigure}
    \hfill
    \begin{subfigure}[t]{0.24\textwidth}
        \includegraphics[width=\textwidth]{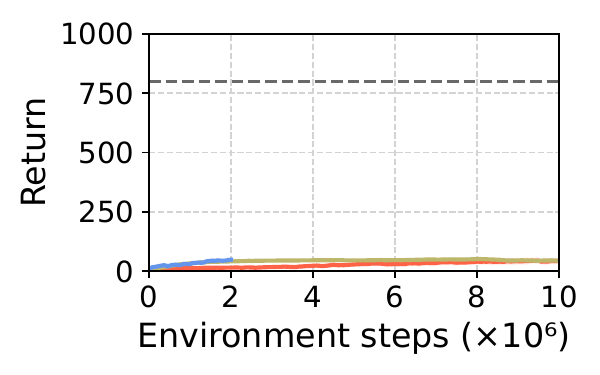}
        \vspace{-1.5em}
        \caption{\texttt{balance\_hard}}
    \end{subfigure}
    \hfill
    \begin{subfigure}[t]{0.24\textwidth}
        \includegraphics[width=\textwidth]{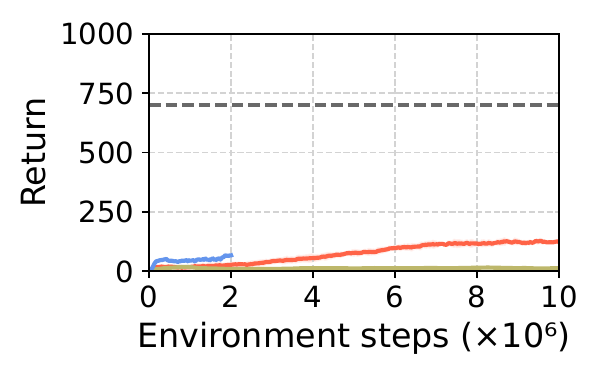}
        \vspace{-1.5em}
        \caption{\texttt{stair}}
    \end{subfigure}
    \\
    \begin{subfigure}[t]{0.24\textwidth}
        \includegraphics[width=\textwidth]{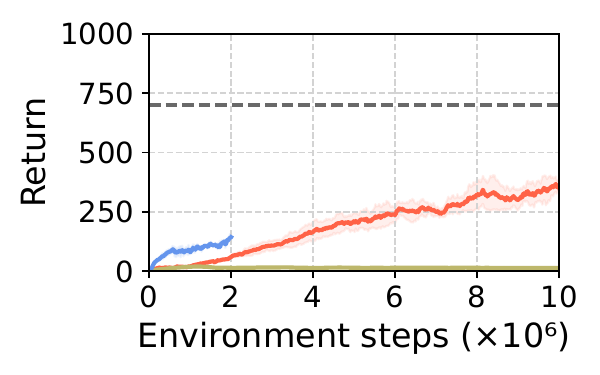}
        \vspace{-1.5em}
        \caption{\texttt{slide}}
    \end{subfigure}
    \hspace{0.5pt}
    \begin{subfigure}[t]{0.24\textwidth}
        \includegraphics[width=\textwidth]{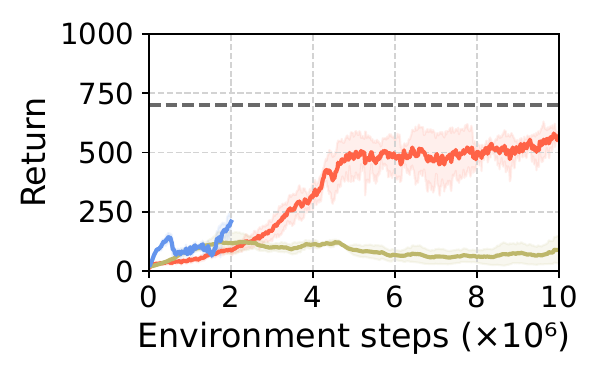}
        \vspace{-1.5em}
        \caption{\texttt{pole}}
    \end{subfigure}
    \caption{
    \textbf{Learning curves of RL algorithms (locomotion).} The curves are averaged over three random seeds and the shaded regions represent the standard deviation. Returns are computed by summing the rewards at all timesteps of an episode. The dashed lines qualitatively indicate task success. 
    We run PPO on the \texttt{walk} task but it is not visible in the plot since it only achieves very low returns.
    }
    \label{fig:benchmark_returns_1}
\end{figure*}

\begin{figure*}
    \centering
    \includegraphics[width=0.45\textwidth]{fig/benchmark/legend_border.pdf}
    \vspace{0.5em}
    \\
    \begin{subfigure}[t]{0.24\textwidth}
        \includegraphics[width=\textwidth]{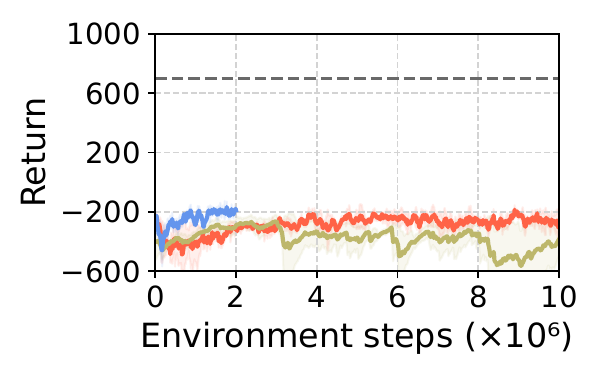}
        \vspace{-1.5em}
        \caption{\texttt{push}}
    \end{subfigure}
    \hfill
    \begin{subfigure}[t]{0.24\textwidth}
        \includegraphics[width=\textwidth]{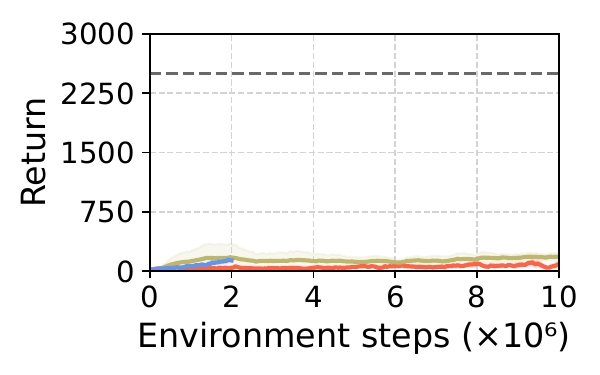}
        \vspace{-1.5em}
        \caption{\texttt{cabinet}}
    \end{subfigure}
    \hfill
    \begin{subfigure}[t]{0.24\textwidth}
        \includegraphics[width=\textwidth]{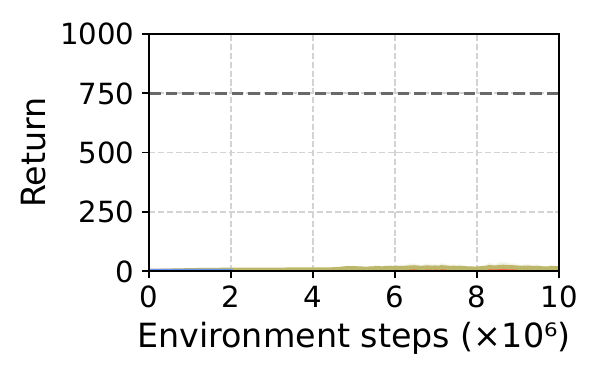}
        \vspace{-1.5em}
        \caption{\texttt{highbar}}
    \end{subfigure}
    \hfill
    \begin{subfigure}[t]{0.24\textwidth}
        \includegraphics[width=\textwidth]{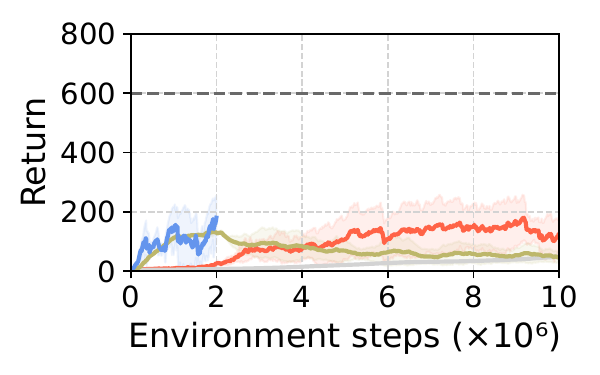}
        \vspace{-1.5em}
        \caption{\texttt{door}}
    \end{subfigure}
    \\
    \begin{subfigure}[t]{0.24\textwidth}
        \includegraphics[width=\textwidth]{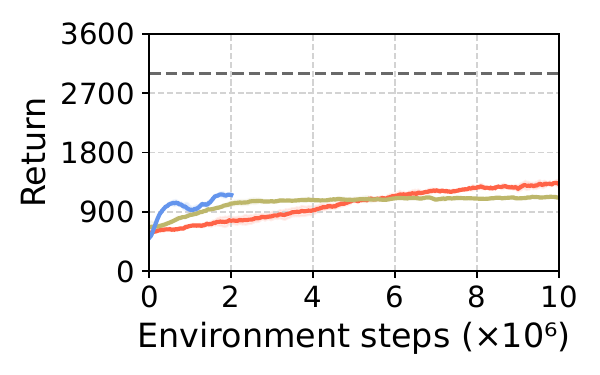}
        \vspace{-1.5em}
        \caption{\texttt{truck}}
    \end{subfigure}
    \hfill
    \begin{subfigure}[t]{0.24\textwidth}
        \includegraphics[width=\textwidth]{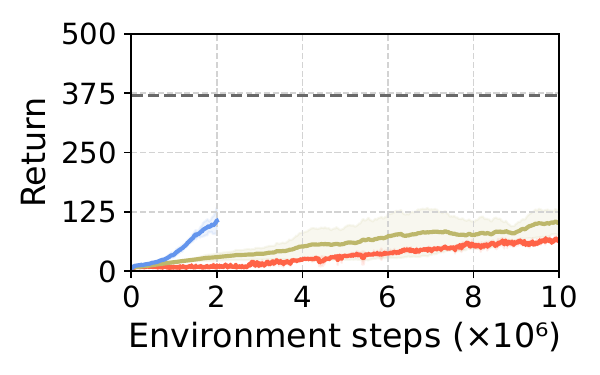}
        \vspace{-1.5em}
        \caption{\texttt{cube}}
    \end{subfigure}
    \hfill
    \begin{subfigure}[t]{0.24\textwidth}
        \includegraphics[width=\textwidth]{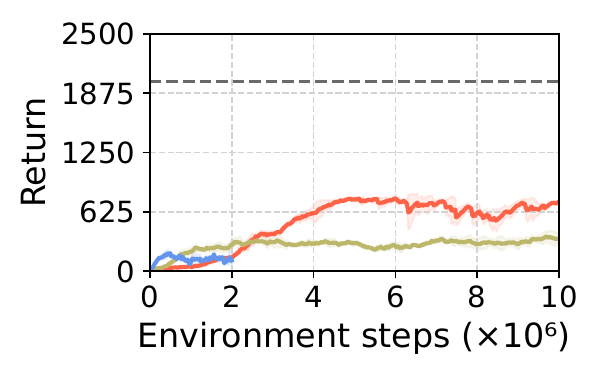}
        \vspace{-1.5em}
        \caption{\texttt{bookshelf\_simple}}
    \end{subfigure}
    \hfill
    \begin{subfigure}[t]{0.24\textwidth}
        \includegraphics[width=\textwidth]{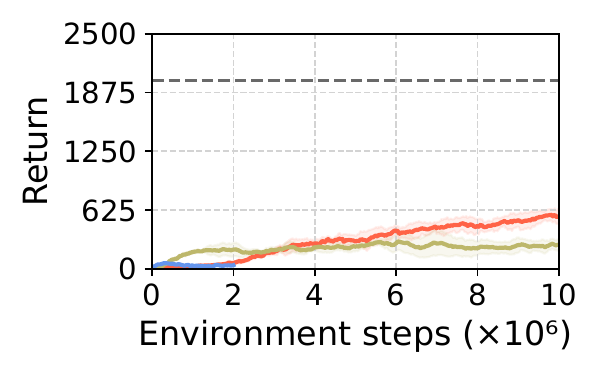}
        \vspace{-1.5em}
        \caption{\texttt{bookshelf\_hard}}
    \end{subfigure}
    \\    
    \begin{subfigure}[t]{0.24\textwidth}
        \includegraphics[width=\textwidth]{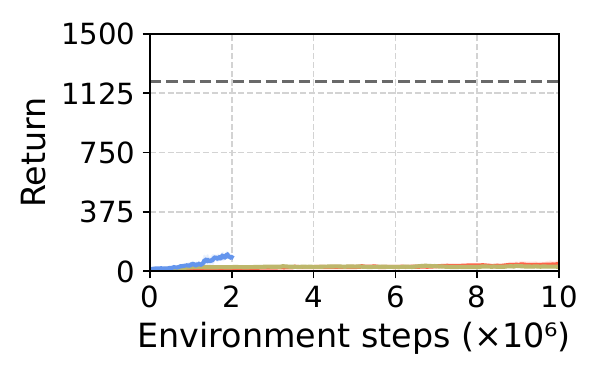}
        \vspace{-1.5em}
        \caption{\texttt{basketball}}
    \end{subfigure}
    \hfill
    \begin{subfigure}[t]{0.24\textwidth}
        \includegraphics[width=\textwidth]{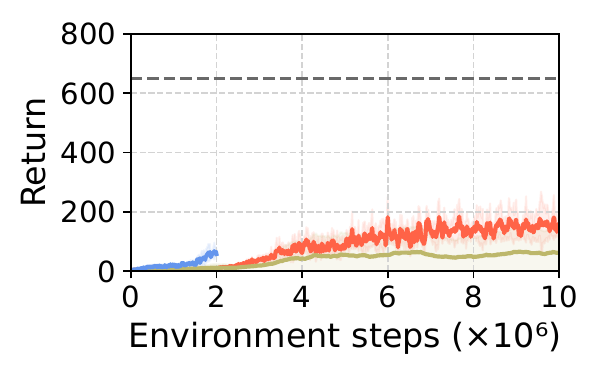}
        \vspace{-1.5em}
        \caption{\texttt{window}}
    \end{subfigure}
    \hfill
    \begin{subfigure}[t]{0.24\textwidth}
        \includegraphics[width=\textwidth]{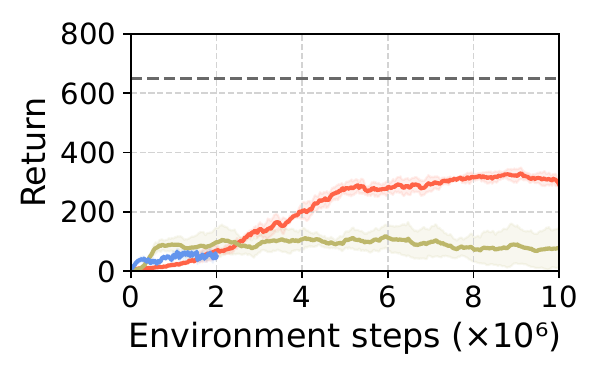}
        \vspace{-1.5em}
        \caption{\texttt{spoon}}
    \end{subfigure}
    \hfill
    \begin{subfigure}[t]{0.24\textwidth}
        \includegraphics[width=\textwidth]{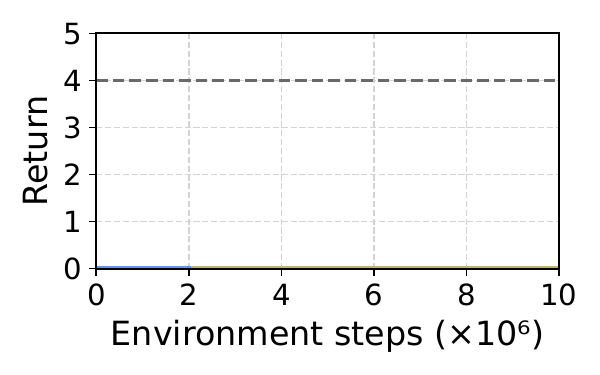}
        \vspace{-1.5em}
        \caption{\texttt{kitchen}}
    \end{subfigure}
    \\
    \begin{subfigure}[t]{0.24\textwidth}
        \includegraphics[width=\textwidth]{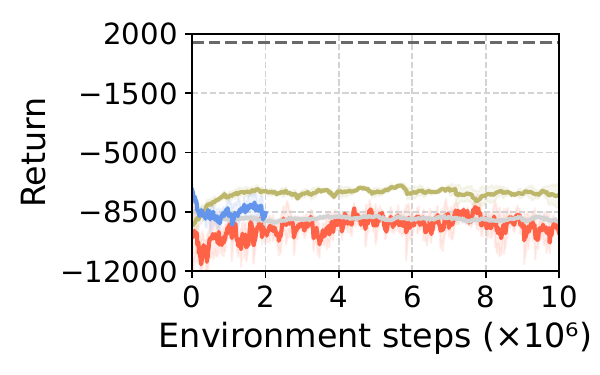}
        \vspace{-1.5em}
        \caption{\texttt{package}}
    \end{subfigure}
    \hspace{0.5pt}
    \begin{subfigure}[t]{0.24\textwidth}
        \includegraphics[width=\textwidth]{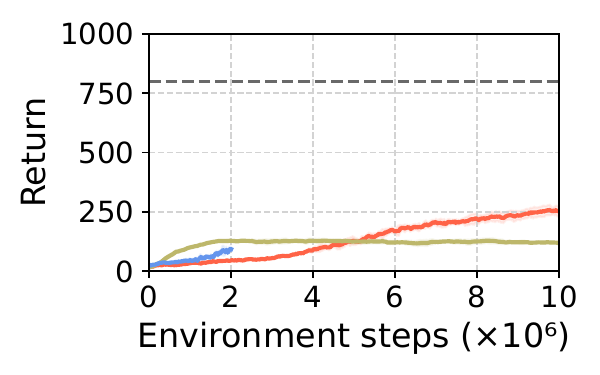}
        \vspace{-1.5em}
        \caption{\texttt{powerlift}}
    \end{subfigure}
    \hspace{0.5pt}
    \begin{subfigure}[t]{0.24\textwidth}
        \includegraphics[width=\textwidth]{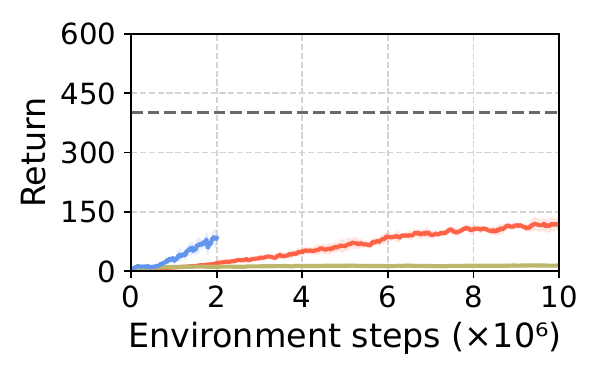}
        \vspace{-1.5em}
        \caption{\texttt{room}}
    \end{subfigure}
    \\
    \begin{subfigure}[t]{0.24\textwidth}
        \includegraphics[width=\textwidth]{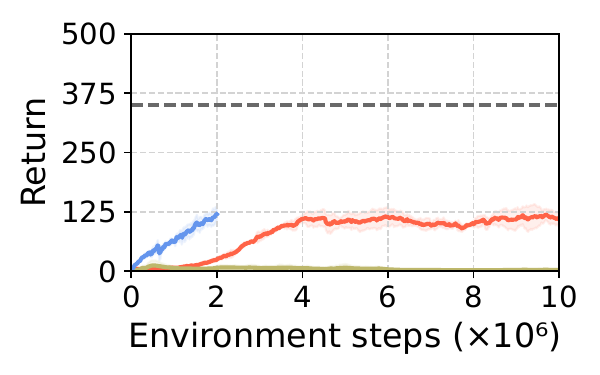}
        \vspace{-1.5em}
        \caption{\texttt{insert\_small}}
    \end{subfigure}
    \hspace{0.5pt}
    \begin{subfigure}[t]{0.24\textwidth}
        \includegraphics[width=\textwidth]{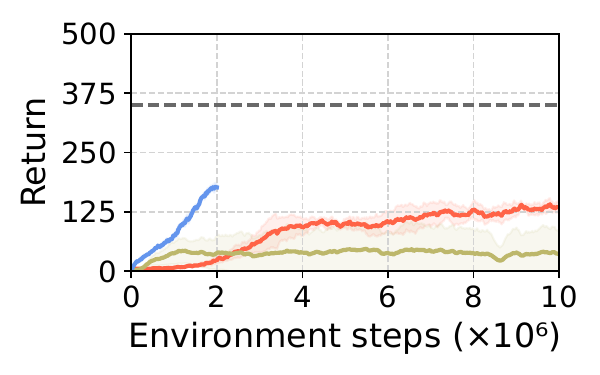}
        \vspace{-1.5em}
        \caption{\texttt{insert\_normal}}
    \end{subfigure}
    \caption{
    \textbf{Learning curves of RL algorithms (manipulation).} The curves are averaged over three random seeds and the shaded regions represent the standard deviation. The dashed lines qualitatively indicate task success. Note that \texttt{kitchen} is the only environment with a purely discrete, sparse reward, with a maximum of $4$.
    }
    \label{fig:benchmark_returns_2}
\end{figure*}

\section{HumanoidBench}
\label{sec:humanoidbench}

Humanoid robots promise to solve human-like tasks in human-tailored environments, possibly using human tools. However, their form factor and hardware challenges make real-world research challenging, making simulation a crucial tool to advance algorithmic research in the field.

To this end, we present HumanoidBench, a humanoid benchmark for robot learning and control, which features a high-dimensional action space (up to $61$ different actuators) and enables research in complex whole-body coordination. 

We benchmark $27$ tasks, consisting of $12$ locomotion tasks and $15$ distinct manipulation tasks, as illustrated in \Cref{fig:locomotion_task_suite} and \Cref{fig:manipulation_task_suite}. A set of locomotion tasks aim to provide interesting but simpler humanoid control scenarios, bypassing intricate dexterous hand control. On the other hand, whole-body manipulation tasks render a comprehensive evaluation of the state-of-the-art algorithms on challenging tasks with unique challenges that require coordination across the entire robot body, ranging from toy examples (e.g., pushing a box on a table) to practical applications (e.g., truck unloading, shelf rearrangement).

In this section, we briefly describe the tasks in our benchmark task suite. Further details about each of the tasks, including task initialization, reward functions, as well as different variations of the tasks, are provided in Appendix, \Cref{sec:environment_details:task_specification}.

\subsection{Locomotion Tasks}
\label{sec:benchmark:locomotion}

\begin{itemize}
    \item \texttt{walk}: Keep forward velocity (in the global $x$-direction) close to \SI{1}{\meter/\second} without falling to the ground.
    \item \texttt{stand}: Maintain a standing pose throughout the provided amount of time.
    \item \texttt{run}: Run forward at a speed of \SI{5}{\meter/\second}.
    \item \texttt{reach}: Reach a randomly initialized 3D point with the left hand.
    \item \texttt{hurdle}: Keep forward velocity close to \SI{5}{\meter/\second} while successfully overcoming hurdles.
    \item \texttt{crawl}: Keep forward velocity close to \SI{1}{\meter/\second} while passing inside a tunnel.
    \item \texttt{maze}: Reach the goal position in a maze by taking multiple turns at the intersections.
    \item \texttt{sit}: Sit onto a chair situated closely behind the robot.
    \item \texttt{balance}: Stay balanced on the unstable board.
    \item \texttt{stair}: Traverse an iterating sequence of upward and downward stairs at \SI{1}{\meter/\second}.
    \item \texttt{slide}: Walk over an iterating sequence of upward and downward slides at \SI{1}{\meter/\second}.
    \item \texttt{pole}: Travel in forward direction over a dense forest of high thin poles, without colliding with them.
\end{itemize}

\subsection{Whole-Body Manipulation Tasks}
\label{sec:benchmark:manipulation}

\begin{itemize}
    \item \texttt{push}: Move a box to a randomly initialized 3D point on a table.
    \item \texttt{cabinet}: Open four different types of cabinet doors (e.g., hinge doors, sliding door, drawer).
    \item \texttt{highbar}: Athletically swing while staying attached to a horizontal high bar until reaching a vertical upside-down position.
    \item \texttt{door}: Pull a door and traverse it while keeping the door open.
    \item \texttt{truck}: Unload packages from a truck by moving them onto a platform.
    \item \texttt{cube}: Manipulate two cubes in-hand until they both reach a randomly initialized target orientation.
    \item \texttt{bookshelf}: Pick and place several items across shelves in a given order.
    \item \texttt{basketball}: Catch a ball coming from random directions and throw it into the basket.
    \item \texttt{window}: Grab a window wiping tool and keep its tip parallel to a window by following a prescribed vertical velocity.
    \item \texttt{spoon}: Grab a spoon and use it to follow a circular pattern inside a pot.
    \item \texttt{kitchen}~\citep{gupta2019relay}: Execute a sequence of actions in a kitchen environment, namely, open a microwave door, move a kettle, and turning burner and light switches.
    \item \texttt{package}: Move a box to a randomly initialized target position.
    \item \texttt{powerlift}: Lift a barbell shaped object of a designated mass.
    \item \texttt{room}: Organize a \SI{5}{\meter} by \SI{5}{\meter} space populated with randomly scattered object to minimize the variance of scattered objects' locations in $x$, $y$-axis directions.
    \item \texttt{insert}: Insert the ends of a rectangular peg into two tight target blocks.
\end{itemize}

\section{Benchmarking Results}
\label{sec:benchmark}

To identify the challenges in learning with humanoid robots, we benchmark reinforcement learning (RL) algorithms on HumanoidBench, which promises for robots to learn from their own experience. Remarkably, this class of algorithms requires limited domain expertise and does not necessarily rely on expert demonstrations, which are not only expensive but also challenging to collect for humanoid robots.\footnote{While we do not benchmark classical model-based control approaches \citep{feng2014optimization, kuindersma2016optimization} in this work, our environments support actions obtained by using any type of controllers.}

\subsection{Baselines}
\label{sec:benchmark:baselines}

We evaluate all tasks in our benchmark with four RL methods (DreamerV3, TD-MPC2, SAC, PPO). Please refer to Appendix, \Cref{sec:training_details} for implementation details.
\begin{itemize}
    \item \textbf{DreamerV3}~\citep{hafner2023mastering}: the state-of-the-art model-based RL algorithm, learning from imaginary model rollouts.
    \item \textbf{TD-MPC2}~\citep{hansen2024tdmpc2}: the state-of-the-art model-based RL algorithm with online planning.
    \item \textbf{SAC} (Soft Actor-Critic~\citep{haarnoja2018sac}): the state-of-the-art off-policy model-free RL algorithm.
    \item \textbf{PPO} (Proximal Policy Optimization~\citep{schulman2017ppo}): the state-of-the-art on-policy model-free RL algorithm.
\end{itemize}

\subsection{Results}
\label{sec:benchmark:results}

We report benchmarking results in \Cref{fig:benchmark_returns_1} and \Cref{fig:benchmark_returns_2}, where we ran each of the algorithms for approximately $48$ hours, resulting in the visible differences in environment steps (e.g., $2$M steps for TD-MPC2, $10$M steps for DreamerV3). We only run PPO on a subset of tasks (\texttt{walk}, \texttt{kitchen}, \texttt{door}, \texttt{package}), given its inferior performance without massive parallelization. Each of the environments is evaluated with a combination of dense rewards and sparse subtask completion rewards, and for each of these we provide qualitative measures of task success (see dashed lines in \Cref{fig:benchmark_returns_1} and \Cref{fig:benchmark_returns_2}). A detailed description of the reward functions used for each environment is available in Appendix, \Cref{sec:environment_details:task_specification}.

All the baseline algorithms perform below the success threshold on most tasks, particularly struggling on tasks that require long-horizon planning and intricate whole-body coordination in a high-dimensional action space. Surprisingly, these state-of-the-art RL algorithms require a large number of steps to learn even simple locomotion tasks, such as \texttt{walk}, which has been extensively studied with a simplified humanoid agent in the DeepMind Control Suite~\citep{tassa2020dm_control}. 

This poor performance is mainly attributed to \textit{the high-dimensionality of the state and action spaces} of our humanoid robot agent with dexterous hands. Although the hands of the humanoid robot are barely used for most locomotion tasks, the RL algorithms fail to ignore this information, which makes policy learning challenging. In addition, these high-dimensional state and action spaces result in a much larger exploration space, which makes exploration slow or infeasible with simple maximum entropy approaches. This implies the need for incorporating behavioral priors or commonsense knowledge about the world that can ease the exploration problem, when it comes to learning on more complex agents, like humanoid robots. We investigate this further in \Cref{sec:benchmark:ablation_hands}.

This problem becomes even more severe in manipulation tasks, resulting in particularly low rewards in all such tasks. Before learning any manipulation skills, an agent must learn locomotion skills to balance and move towards an object or the world to interact. All the policies barely learn to stabilize using the dense reward, but struggle to learn any complex manipulation skills.

\begin{figure}[t]
    \centering
    \includegraphics[width=\linewidth]{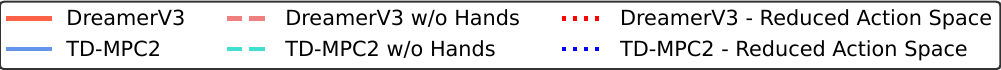}
    \begin{subfigure}[t]{0.48\linewidth}
        \includegraphics[width=\textwidth]{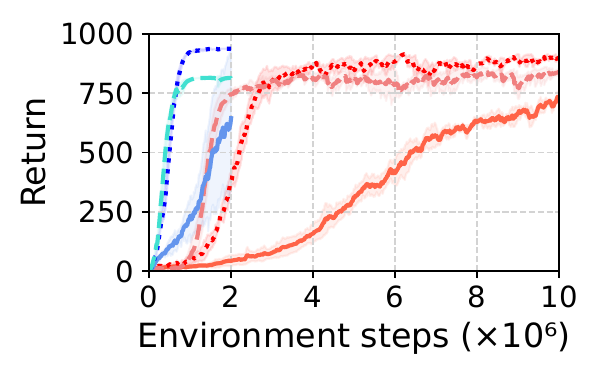}
        \vspace{-1.5em}
        \caption{\texttt{walk}}
    \end{subfigure}
    \hfill
    \begin{subfigure}[t]{0.48\linewidth}
        \includegraphics[width=\textwidth]{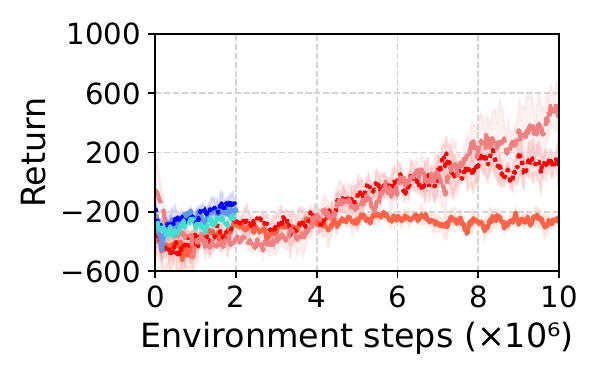}
        \vspace{-1.5em}
        \caption{\texttt{push}}
    \end{subfigure}
 
    \caption{\textbf{Performance with and without dexterous hands.} The curves are averaged over three random seeds and the shaded regions represent the standard deviation.}
    \label{fig:ablation_hands}
\end{figure}

\subsection{With Hands vs. Alternative Configurations}
\label{sec:benchmark:ablation_hands}

\textbf{With Hands vs. Without Hands.}\quad
As discussed in \Cref{sec:benchmark:results}, controlling humanoid robots with dexterous hands is challenging due to their high degrees of freedom and complex dynamics. Thus, we investigate the difficulty of RL training with a large action space (i.e., additional $42$ dimensions with two dexterous Shadow Hands) on \texttt{walk} that does not necessarily require to control dexterous hands. The results in \Cref{fig:ablation_hands} show that the presence of hands, with their additional joints and actuators, leads to a large decrease in performance compared to training the same task without the dexterous hands (see differences in observation and action space in \Cref{tab:robot_spec}). 

\textbf{Reduced Action Space.}\quad
To verify whether such difficulties stem from the dimensionality of the action space, we benchmark our full robot model, but fix the actuation of the hands ($42$D), which we set to zero. In this way, the action dimensionality is reduced from $61$D in the original model to $19$D. Note that the observations and masses induced by the presence of the hands are retained (i.e., observation space remains $151$D). \Cref{fig:ablation_hands} shows that the RL algorithms learn significantly faster in the reduced action space setup than the ones trained with the full action space. This confirms that most of the performance drop is indeed due to the increased action dimensionality. 

We observe similar trends in the more complex manipulation task, \texttt{push}, which presents substantially different dynamics in the task approach (e.g., pushing with and without hands).

\subsection{Flat vs. Hierarchical Reinforcement Learning}
\label{sec:benchmark:ablation_hierarchy}

\begin{figure*}[t]
    \centering
    \begin{subfigure}[t]{0.37\linewidth}
        \includegraphics[width=\textwidth]{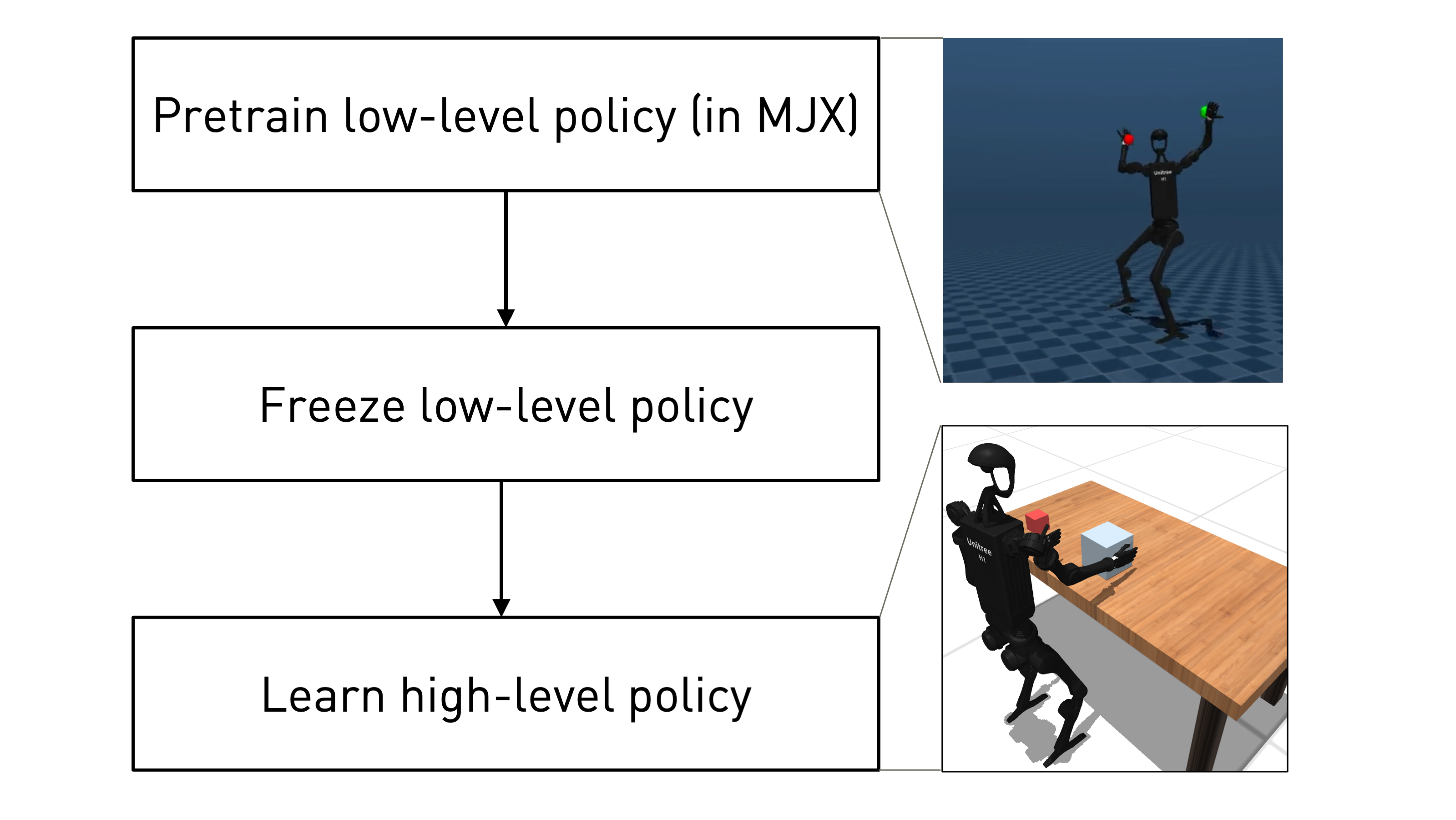}
        \vspace{-1.5em}
        \caption{Hierarchical learning pipeline}
    \end{subfigure}
    \begin{subfigure}[t]{0.34\linewidth}
        \includegraphics[width=\textwidth]{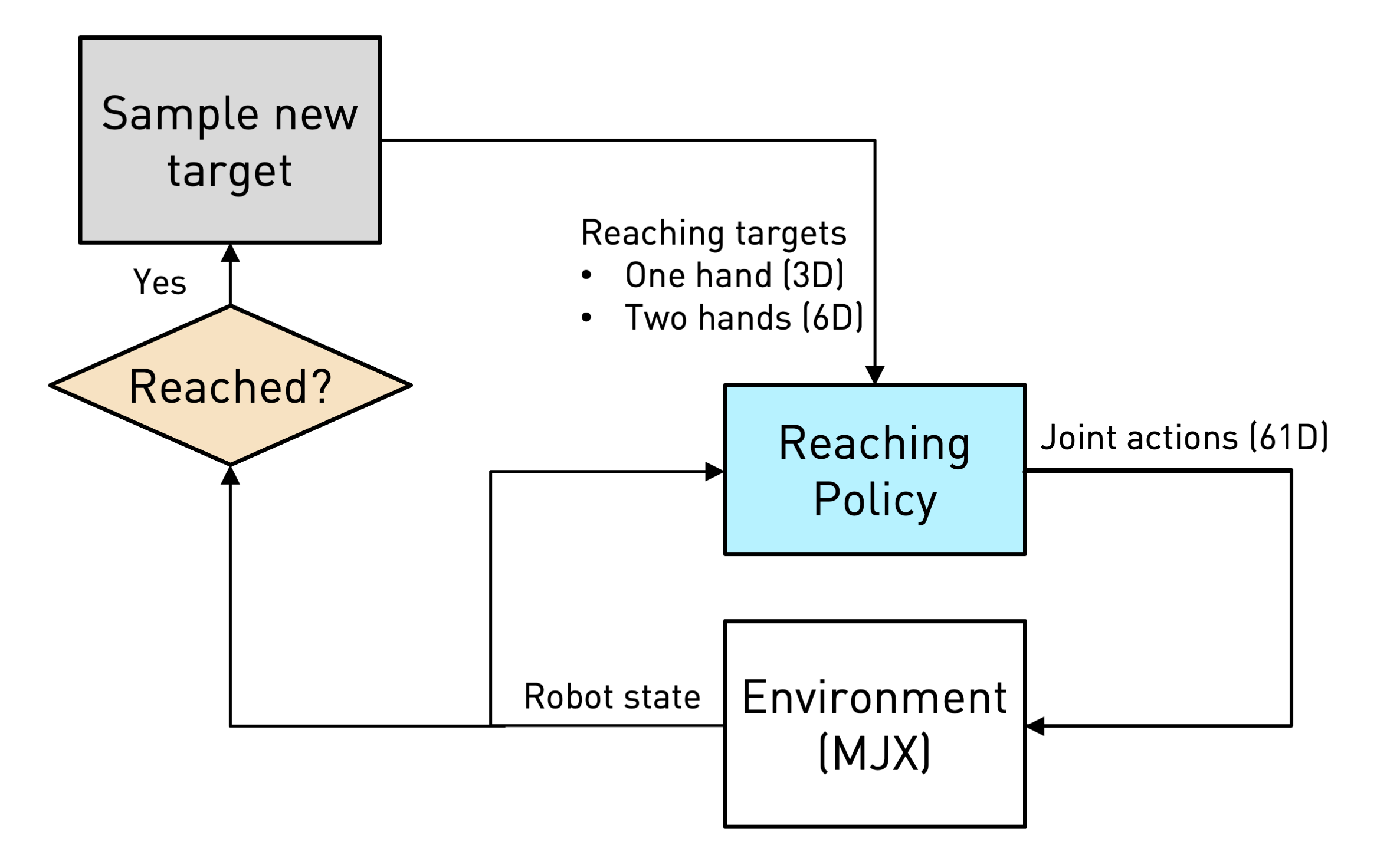}
        \vspace{-1.5em}
        \caption{Low-level policy pretraining}
    \end{subfigure}
    \begin{subfigure}[t]{0.27\linewidth}
        \includegraphics[width=\textwidth]{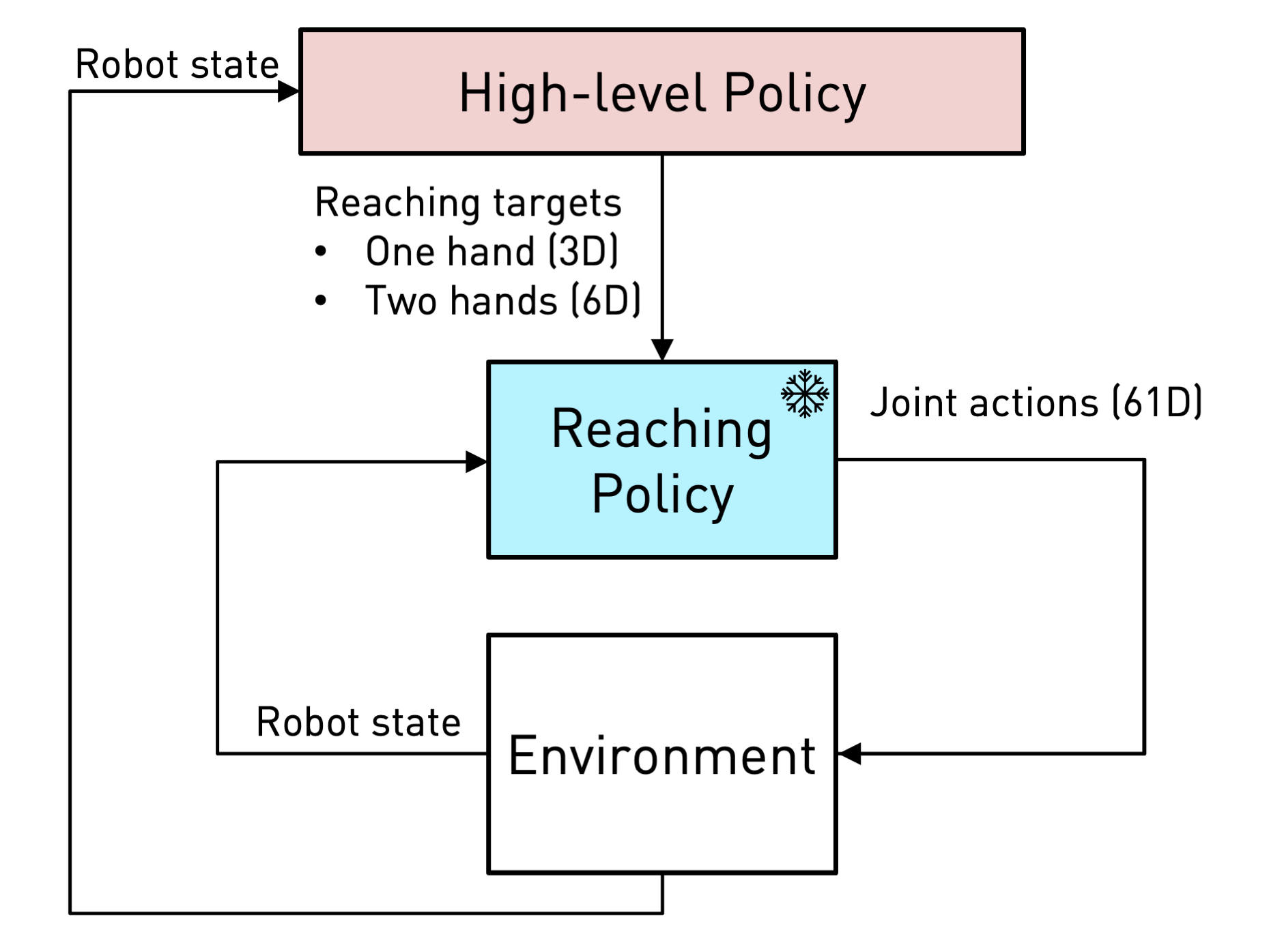}
        \vspace{-1.5em}
        \caption{High-level policy training}
    \end{subfigure}
    \caption{\textbf{Our hierarchical RL pipeline (a).} (b)~A robust low-level reaching policy is pretrained using PPO in a MuJoCo MJX-based reaching environment, as shown in the top snapshot in (a). (c)~The high-level policy then leverages the pretrained reaching policy to move to a desired position and learns to solve a downstream task, shown in the bottom snapshot in (a). Note that the reaching policy weights are frozen during the high-level policy training.}
    \label{fig:hierarchical}
\end{figure*}

As shown in the previous subsection, flat, end-to-end RL approaches fail to learn most of the tasks in HumanoidBench. Many of such tasks require long-horizon planning and necessitate acquiring a diverse set of skills (e.g., balancing, walking, reaching). These tasks can be addressed by hierarchical RL, which introduces additional structure into the learning problem. In HRL, one or multiple low-level skill policies are provided to a high-level planning policy that outputs setpoints for the lower-level policies. In practice, such setpoints comprise the action space of the high-level policy. This framework is very general, and there are no constraints on how to obtain both low-level and high-level policies. However, here we focus on training both of these through reinforcement learning~\citep{sutton1999between}.

\textbf{Hierarchical RL Implementation.}\quad 
We implement a hierarchical RL approach on two manipulation tasks, namely, the \texttt{push} and \texttt{package} tasks. As a low-level skill, \texttt{push} uses a \textit{one-hand reaching policy}, which allows the robot to reach a 3D point in space with its left hand, while \texttt{package} uses a \textit{two-hand reaching policy}, where both hands are commanded to reach different 3D targets. \Cref{fig:hierarchical} illustrates the overview of our hierarchical RL implementation. 

\textbf{Low-level Reaching Policy Pretraining.}\quad 
We treat the low-level reaching policy as a pretrained frozen block that can be reused across tasks. Since this policy does not improve during training of the high-level policy, it needs to be very \textit{robust} to cope with the continually shifting reaching targets that the high-level policy sets during exploration. However, the results in the previous section show that even a one-hand reaching task is hard to learn.

On the other hand, while our experiments above confirm that PPO exhibits poor sample efficiency compared to the off-policy algorithms, it is worth noting that PPO has achieved significant success in robotic locomotion by exploiting large-scale parallelization of environments on GPUs~\cite{makoviychuk2021isaac}. To achieve robust reaching policies, we exploit hardware acceleration by pretraining the low-level reaching policies in the recently released MuJoCo MJX\footnote{\url{https://mujoco.readthedocs.io/en/stable/mjx.html}}, which enables training PPO on thousands of parallel environments. 

For low-level reaching policy training, we employ a simplified H1 model that only considers collisions between feet and ground in the MuJoCo MJX environments, as in our experience the advantages stemming from parallelization are largely reduced when considering all numerous humanoid geometries (hindering training of the more complex benchmark tasks via MJX). We also remove the hands from the model to further increase training efficiency. The simplified reaching task environments for pretraining reset the target once reached. To achieve robust low-level reaching policies, we apply force perturbations at each of the links during training. We train the one-hand reaching policy for $2$ billion steps ($36$ hours) and the two-hand reaching policy for $4$ billion steps ($60$ hours) on $32{,}768$ parallel environments. The pretrained reaching policies successfully transfer to the original (non-simplified, simulated in classical MuJoCo) humanoid environments.

\textbf{High-level Policy Training.}\quad 
Then, we use the pretrained reaching policies (frozen) as low-level policies and only train a high-level policy using either DreamerV3 and TD-MPC2 on the \texttt{push} and \texttt{package} tasks. To facilitate exploration, we restrict the range of reaching targets to the robot workspace.

\begin{figure}[t]
    \centering
    \includegraphics[width=0.8\linewidth]{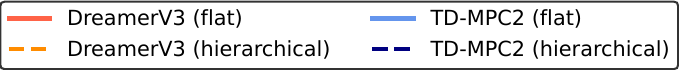}
    \begin{subfigure}[t]{0.48\linewidth}
        \includegraphics[width=\textwidth]{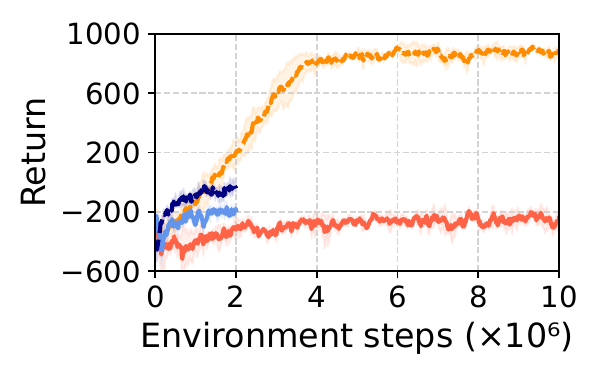}
        \vspace{-1.7em}
        \caption{\texttt{push}}
    \end{subfigure}
    \hfill
    \begin{subfigure}[t]{0.48\linewidth}
        \includegraphics[width=\textwidth]{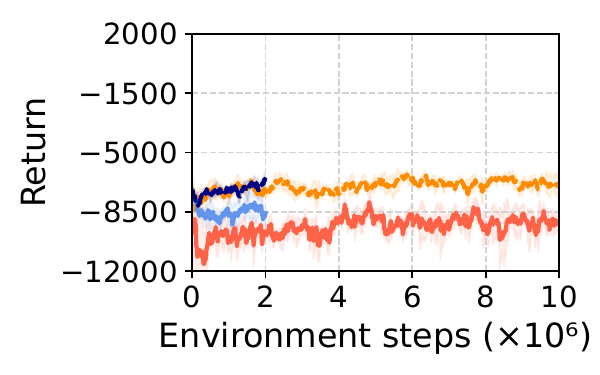}
        \vspace{-1.7em}
        \caption{\texttt{package}}
    \end{subfigure}
 
    \caption{\textbf{Comparison between flat policies and hierarchical policies.} The curves are averaged over three random seeds and the shaded regions represent the standard deviation.}
    \label{fig:ablation_hierarchy}
\end{figure}

\textbf{Hierarchical RL Results.}\quad 
In \Cref{fig:ablation_hierarchy}, our hierarchical architecture significantly outperforms the flat, end-to-end baselines on the \texttt{push} task, achieving very high success rates with DreamerV3. 
While the low-level policy has undergone additional pretraining, this can be in principle reused across tasks. 
On the other hand, we note a less pronounced performance improvement in the more challenging \texttt{package} task. While getting closer to picking up the package with our hierarchical approach, the policy struggles in lifting it (having never experienced it during training).

These results confirm that the tasks in our benchmark present challenges that can be addressed with a more structured approach to the learning problem, and we hope this stimulates further directions for future research.

\subsection{Common Failures}
\label{sec:results:failures}

\begin{figure}[h]
    \centering
    \begin{subfigure}[t]{0.32\linewidth}
        \includegraphics[width=\textwidth]{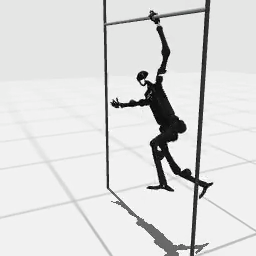}
        \caption{\texttt{highbar}}
    \end{subfigure}
    \begin{subfigure}[t]{0.32\linewidth}
        \includegraphics[width=\textwidth]{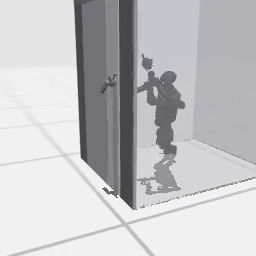}
        \caption{\texttt{door}}
    \end{subfigure}
    \begin{subfigure}[t]{0.32\linewidth}
        \includegraphics[width=\textwidth]{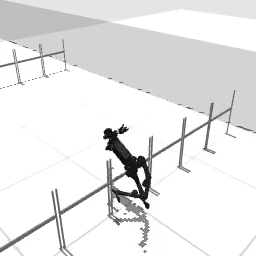}
        \caption{\texttt{hurdle}}
    \end{subfigure}    
    \caption{\textbf{Failure Scenarios.} This figure presents a selection of common failures that occur while training our benchmark tasks.}
    \label{fig:failures}
\end{figure}

In this subsection, we remark on notable challenges and common failures for some representative tasks in our benchmark, which denote the challenge in learning with high-dimensional action spaces and limited planning horizon of the state-of-the-art RL algorithms.

\textbf{\textit{Common Failure on \texttt{highbar}.}}
In the \texttt{highbar} task, the Unitree H1 robot conservatively learns to maintain contact with the bar to avoid episode termination, but experiences difficulties in performing the whole-body rotation trajectory. This is indicative of short horizon planning and a recurrent challenge in many of the long-horizon benchmark tasks, despite the availability of dense rewards.

\textbf{\textit{Common Failure on \texttt{door}.}}
In the \texttt{door} task, the robot is well-guided to turn the door hatch to unlock the door, but it finds it challenging to learn the precise motion required to pull the door towards its opening position. This is mainly because pulling the door requires not only pulling its arm but also moving the whole body backwards. The coordination between multiple body parts and seamless interaction between manipulation and locomotion skills are common challenges in training humanoid robots.

\textbf{\textit{Common Failure on \texttt{hurdle}.}}
In the \texttt{hurdle} environment, the robot learns to run forward with the expected velocity but does not recognize the need to surpass the hurdle by jumping, which is a hard exploration problem. Previous work has shown that in OpenAI gym Walker2d, the forward-moving reward is sufficient to learn this behavior~\citep{lee2019composing}. On the other hand, the humanoid robot finds conservative poses to collide with the hurdle such that it can stabilize without terminating the episode after hitting the obstacle, without further exploring high-reward jumping behaviors.


\section{Conclusion} 
\label{sec:conclusion}

We presented HumanoidBench, a high-dimensional humanoid robot control benchmark.
Ours is the first example of a comprehensive humanoid environment with a diversity of locomotion and manipulation tasks, ranging from toy examples to practical humanoid applications. We set a high bar with our complex tasks, in the hope to stimulate the community to accelerate the development of whole-body algorithms for such robotic platforms.

\textbf{Future work.}\quad HumanoidBench already includes multimodal high-dimensional observations in the form of egocentric vision and whole-body tactile sensing. While our experiments only benchmarked the performance of state-based environments, studying the interplay between different modalities is a compelling direction for future work. 

Extensions of the humanoid environment will also eventually include more realistic objects and environments with real-world diversity and higher-quality rendering. As for dexterous manipulation tasks, we envision screwing and furniture assembly tasks being part of our framework, given that they are particularly tailored for bimanual manipulation.

Here we have focused on reinforcement learning algorithms because collecting physical demonstrations with humanoid robots is particularly challenging. However, we believe that other means could be employed to bootstrap learning (e.g., learning from human videos).

Finally, while this was not the focus of our work, the impressive results obtained via domain randomization in the newly developed MuJoCo MJX show promise to study sim-to-real transfer in more depth, following the large success of the field in quadrupedal locomotion \cite{hwangbo2019learning}.

\section*{Acknowledgments}

This work was supported in part by the SNSF Postdoc Mobility Fellowship 211086, ONR MURI N00014-22-1-2773, BAIR Industrial Consortium, Komatsu, InnoHK Centre for Logistics Robotics, an ONR DURIP grant, the Institute of Information \& Communications Technology Planning \& Evaluation (IITP) grant and the National Research Foundation of Korea (NRF) grant funded by the Korean government (MSIT) (RS-2020-II201361, Artificial Intelligence Graduate School Program (Yonsei University) and RS-2024-00333634). We also thank Google TPU Research Cloud (TRC) for granting us access to TPUs for research.

\bibliographystyle{plainnat}
\bibliography{conferences, references}

\begin{thebibliography}{71}
\providecommand{\natexlab}[1]{#1}
\providecommand{\url}[1]{\texttt{#1}}
\expandafter\ifx\csname urlstyle\endcsname\relax
  \providecommand{\doi}[1]{doi: #1}\else
  \providecommand{\doi}{doi: \begingroup \urlstyle{rm}\Url}\fi

\bibitem[Adu-Bredu et~al.(2023)Adu-Bredu, Gibson, and Grizzle]{adu2023exploring}
Alphonsus Adu-Bredu, Grant Gibson, and Jessy Grizzle.
\newblock Exploring kinodynamic fabrics for reactive whole-body control of underactuated humanoid robots.
\newblock In \emph{IEEE/RSJ International Conference on Intelligent Robots and Systems}, pages 10397--10404. IEEE, 2023.

\bibitem[Akkaya et~al.(2019)Akkaya, Andrychowicz, Chociej, Litwin, McGrew, Petron, Paino, Plappert, Powell, Ribas, et~al.]{akkaya2019solving}
Ilge Akkaya, Marcin Andrychowicz, Maciek Chociej, Mateusz Litwin, Bob McGrew, Arthur Petron, Alex Paino, Matthias Plappert, Glenn Powell, Raphael Ribas, et~al.
\newblock Solving rubik's cube with a robot hand.
\newblock \emph{arXiv preprint arXiv:1910.07113}, 2019.

\bibitem[Al-Hafez et~al.(2023)Al-Hafez, Zhao, Peters, and Tateo]{alhafez2023locomujoco}
Firas Al-Hafez, Guoping Zhao, Jan Peters, and Davide Tateo.
\newblock Locomujoco: A comprehensive imitation learning benchmark for locomotion.
\newblock \emph{6th Robot Learning Workshop at NeurIPS}, 2023.

\bibitem[Bacon et~al.(2017)Bacon, Harb, and Precup]{bacon2017option-critic}
Pierre-Luc Bacon, Jean Harb, and Doina Precup.
\newblock The option-critic architecture.
\newblock In \emph{Association for the Advancement of Artificial Intelligence}, pages 1726--1734, 2017.

\bibitem[{Bellemare} et~al.(2013){Bellemare}, {Naddaf}, {Veness}, and {Bowling}]{bellemare13arcade}
M.~G. {Bellemare}, Y.~{Naddaf}, J.~{Veness}, and M.~{Bowling}.
\newblock The arcade learning environment: An evaluation platform for general agents.
\newblock \emph{Journal of Artificial Intelligence Research}, 47:\penalty0 253--279, jun 2013.

\bibitem[Berg et~al.(2023)Berg, Caggiano, and Kumar]{berg2023sar}
Cameron~H Berg, Vittorio Caggiano, and Vikash Kumar.
\newblock {SAR: Generalization of Physiological Dexterity via Synergistic Action Representation}.
\newblock In \emph{Robotics: Science and Systems}, 2023.

\bibitem[Brockman et~al.(2016)Brockman, Cheung, Pettersson, Schneider, Schulman, Tang, and Zaremba]{brockman2016openai}
Greg Brockman, Vicki Cheung, Ludwig Pettersson, Jonas Schneider, John Schulman, Jie Tang, and Wojciech Zaremba.
\newblock Openai gym.
\newblock \emph{arXiv preprint arXiv:1606.01540}, 2016.

\bibitem[Caggiano et~al.(2022)Caggiano, Wang, Durandau, Sartori, and Kumar]{caggiano2022myosuite}
Vittorio Caggiano, Huawei Wang, Guillaume Durandau, Massimo Sartori, and Vikash Kumar.
\newblock Myosuite: A contact-rich simulation suite for musculoskeletal motor control.
\newblock In \emph{Learning for Dynamics and Control}, pages 492--507. PMLR, 2022.

\bibitem[Caggiano et~al.(2023)Caggiano, Dasari, and Kumar]{caggiano2023myodex}
Vittorio Caggiano, Sudeep Dasari, and Vikash Kumar.
\newblock Myodex: a generalizable prior for dexterous manipulation.
\newblock In \emph{International Conference on Machine Learning}, pages 3327--3346. PMLR, 2023.

\bibitem[Chen et~al.(2023)Chen, Geng, Zhong, Ji, Jiang, Lu, Dong, and Yang]{chen2023bi}
Yuanpei Chen, Yiran Geng, Fangwei Zhong, Jiaming Ji, Jiechuang Jiang, Zongqing Lu, Hao Dong, and Yaodong Yang.
\newblock Bi-dexhands: Towards human-level bimanual dexterous manipulation.
\newblock \emph{IEEE Transactions on Pattern Analysis and Machine Intelligence}, 2023.

\bibitem[Cheng et~al.(2024)Cheng, Ji, Chen, Yang, Yang, and Wang]{cheng2024expressive}
Xuxin Cheng, Yandong Ji, Junming Chen, Ruihan Yang, Ge~Yang, and Xiaolong Wang.
\newblock Expressive whole-body control for humanoid robots.
\newblock \emph{arXiv preprint arXiv:2402.16796}, 2024.

\bibitem[Chi et~al.(2023)Chi, Feng, Du, Xu, Cousineau, Burchfiel, and Song]{chi2023diffusion}
Cheng Chi, Siyuan Feng, Yilun Du, Zhenjia Xu, Eric Cousineau, Benjamin Burchfiel, and Shuran Song.
\newblock Diffusion policy: Visuomotor policy learning via action diffusion.
\newblock In \emph{Robotics: Science and Systems}, 2023.

\bibitem[Duan et~al.(2017)Duan, Andrychowicz, Stadie, Ho, Schneider, Sutskever, Abbeel, and Zaremba]{duan2017one-shot}
Yan Duan, Marcin Andrychowicz, Bradly Stadie, Jonathan Ho, Jonas Schneider, Ilya Sutskever, Pieter Abbeel, and Wojciech Zaremba.
\newblock One-shot imitation learning.
\newblock In \emph{Advances in Neural Information Processing Systems}, pages 1087--1098, 2017.

\bibitem[Feng et~al.(2014)Feng, Whitman, Xinjilefu, and Atkeson]{feng2014optimization}
Siyuan Feng, Eric Whitman, X~Xinjilefu, and Christopher~G Atkeson.
\newblock Optimization based full body control for the atlas robot.
\newblock In \emph{2014 IEEE-RAS International Conference on Humanoid Robots}, pages 120--127. IEEE, 2014.

\bibitem[Fu et~al.(2024)Fu, Zhao, and Finn]{fu2024mobile}
Zipeng Fu, Tony~Z Zhao, and Chelsea Finn.
\newblock Mobile aloha: Learning bimanual mobile manipulation with low-cost whole-body teleoperation.
\newblock \emph{arXiv preprint arXiv:2401.02117}, 2024.

\bibitem[Ghosh(2023)]{jaxrl_minimal}
Dibya Ghosh.
\newblock dibyaghosh/jaxrl\_m, 2023.
\newblock URL \url{https://github.com/dibyaghosh/jaxrl_m}.

\bibitem[Gupta et~al.(2019)Gupta, Kumar, Lynch, Levine, and Hausman]{gupta2019relay}
Abhishek Gupta, Vikash Kumar, Corey Lynch, Sergey Levine, and Karol Hausman.
\newblock Relay policy learning: Solving long-horizon tasks via imitation and reinforcement learning.
\newblock \emph{Conference on Robot Learning}, 2019.

\bibitem[Haarnoja et~al.(2018)Haarnoja, Zhou, Abbeel, and Levine]{haarnoja2018sac}
Tuomas Haarnoja, Aurick Zhou, Pieter Abbeel, and Sergey Levine.
\newblock Soft actor-critic: Off-policy maximum entropy deep reinforcement learning with a stochastic actor.
\newblock In \emph{International Conference on Machine Learning}, pages 1856--1865, 2018.

\bibitem[Haarnoja et~al.(2023)Haarnoja, Moran, Lever, Huang, Tirumala, Wulfmeier, Humplik, Tunyasuvunakool, Siegel, Hafner, Bloesch, Hartikainen, Byravan, Hasenclever, Tassa, Sadeghi, Batchelor, Casarini, Saliceti, Game, Sreendra, Patel, Gwira, Huber, Hurley, Nori, Hadsell, and Heess]{haarnoja2023learning}
Tuomas Haarnoja, Ben Moran, Guy Lever, Sandy~H Huang, Dhruva Tirumala, Markus Wulfmeier, Jan Humplik, Saran Tunyasuvunakool, Noah~Y Siegel, Roland Hafner, Michael Bloesch, Kristian Hartikainen, Arunkumar Byravan, Leonard Hasenclever, Yuval Tassa, Fereshteh Sadeghi, Nathan Batchelor, Federico Casarini, Stefano Saliceti, Charles Game, Neil Sreendra, Kushal Patel, Marlon Gwira, Andrea Huber, Nicole Hurley, Francesco Nori, Raia Hadsell, and Nicolas Heess.
\newblock Learning agile soccer skills for a bipedal robot with deep reinforcement learning.
\newblock \emph{arXiv preprint arXiv:2304.13653}, 2023.

\bibitem[Hafner et~al.(2023)Hafner, Pasukonis, Ba, and Lillicrap]{hafner2023mastering}
Danijar Hafner, Jurgis Pasukonis, Jimmy Ba, and Timothy Lillicrap.
\newblock Mastering diverse domains through world models.
\newblock \emph{arXiv preprint arXiv:2301.04104}, 2023.

\bibitem[Hansen et~al.(2024)Hansen, Su, and Wang]{hansen2024tdmpc2}
Nicklas Hansen, Hao Su, and Xiaolong Wang.
\newblock Td-mpc2: Scalable, robust world models for continuous control.
\newblock In \emph{International Conference on Learning Representations}, 2024.

\bibitem[Heo et~al.(2023)Heo, Lee, Lee, and Lim]{heo2023furniturebench}
Minho Heo, Youngwoon Lee, Doohyun Lee, and Joseph~J. Lim.
\newblock Furniturebench: Reproducible real-world benchmark for long-horizon complex manipulation.
\newblock In \emph{Robotics: Science and Systems}, 2023.

\bibitem[Hwangbo et~al.(2019)Hwangbo, Lee, Dosovitskiy, Bellicoso, Tsounis, Koltun, and Hutter]{hwangbo2019learning}
Jemin Hwangbo, Joonho Lee, Alexey Dosovitskiy, Dario Bellicoso, Vassilios Tsounis, Vladlen Koltun, and Marco Hutter.
\newblock Learning agile and dynamic motor skills for legged robots.
\newblock \emph{Science Robotics}, 4\penalty0 (26):\penalty0 eaau5872, 2019.

\bibitem[James et~al.(2020)James, Ma, Rovick~Arrojo, and Davison]{james2019rlbench}
Stephen James, Zicong Ma, David Rovick~Arrojo, and Andrew~J. Davison.
\newblock Rlbench: The robot learning benchmark \& learning environment.
\newblock \emph{IEEE Robotics and Automation Letters}, 2020.

\bibitem[Kannan et~al.(2021)Kannan, Hafner, Finn, and Erhan]{kannan2021robodesk}
Harini Kannan, Danijar Hafner, Chelsea Finn, and Dumitru Erhan.
\newblock Robodesk: A multi-task reinforcement learning benchmark.
\newblock \url{https://github.com/google-research/robodesk}, 2021.

\bibitem[Kuindersma et~al.(2016)Kuindersma, Deits, Fallon, Valenzuela, Dai, Permenter, Koolen, Marion, and Tedrake]{kuindersma2016optimization}
Scott Kuindersma, Robin Deits, Maurice Fallon, Andr{\'e}s Valenzuela, Hongkai Dai, Frank Permenter, Twan Koolen, Pat Marion, and Russ Tedrake.
\newblock Optimization-based locomotion planning, estimation, and control design for the atlas humanoid robot.
\newblock \emph{Autonomous robots}, 40:\penalty0 429--455, 2016.

\bibitem[Kumar et~al.(2021)Kumar, Fu, Pathak, and Malik]{kumar2021rma}
Ashish Kumar, Zipeng Fu, Deepak Pathak, and Jitendra Malik.
\newblock Rma: Rapid motor adaptation for legged robots.
\newblock In \emph{Robotics: Science and Systems}, 2021.

\bibitem[Lee et~al.(2019{\natexlab{a}})Lee, Park, Lee, and Lee]{lee2019scalable}
Seunghwan Lee, Moonseok Park, Kyoungmin Lee, and Jehee Lee.
\newblock Scalable muscle-actuated human simulation and control.
\newblock \emph{ACM Transactions on Graphics}, 38\penalty0 (4):\penalty0 1--13, 2019{\natexlab{a}}.

\bibitem[Lee et~al.(2019{\natexlab{b}})Lee, Sun, Somasundaram, Hu, and Lim]{lee2019composing}
Youngwoon Lee, Shao-Hua Sun, Sriram Somasundaram, Edward~S. Hu, and Joseph~J. Lim.
\newblock Composing complex skills by learning transition policies.
\newblock In \emph{International Conference on Learning Representations}, 2019{\natexlab{b}}.
\newblock URL \url{https://openreview.net/forum?id=rygrBhC5tQ}.

\bibitem[Lee et~al.(2020)Lee, Yang, and Lim]{lee2020learning}
Youngwoon Lee, Jingyun Yang, and Joseph~J. Lim.
\newblock Learning to coordinate manipulation skills via skill behavior diversification.
\newblock In \emph{International Conference on Learning Representations}, 2020.

\bibitem[Lee et~al.(2021)Lee, Hu, and Lim]{lee2021ikea}
Youngwoon Lee, Edward~S Hu, and Joseph~J Lim.
\newblock {IKEA} furniture assembly environment for long-horizon complex manipulation tasks.
\newblock In \emph{IEEE International Conference on Robotics and Automation}, 2021.
\newblock URL \url{https://clvrai.com/furniture}.

\bibitem[Li et~al.(2022)Li, Zhang, Wong, Gokmen, Srivastava, Mart{\'\i}n-Mart{\'\i}n, Wang, Levine, Lingelbach, Sun, Anvari, Hwang, Sharma, Aydin, Bansal, Hunter, Kim, Lou, Matthews, Villa-Renteria, Tang, Tang, Xia, Savarese, Gweon, Liu, Wu, and Fei-Fei]{li2022behavior}
Chengshu Li, Ruohan Zhang, Josiah Wong, Cem Gokmen, Sanjana Srivastava, Roberto Mart{\'\i}n-Mart{\'\i}n, Chen Wang, Gabrael Levine, Michael Lingelbach, Jiankai Sun, Mona Anvari, Minjune Hwang, Manasi Sharma, Arman Aydin, Dhruva Bansal, Samuel Hunter, Kyu-Young Kim, Alan Lou, Caleb~R Matthews, Ivan Villa-Renteria, Jerry~Huayang Tang, Claire Tang, Fei Xia, Silvio Savarese, Hyowon Gweon, Karen Liu, Jiajun Wu, and Li~Fei-Fei.
\newblock Behavior-1k: A benchmark for embodied ai with 1,000 everyday activities and realistic simulation.
\newblock In \emph{Conference on Robot Learning}, 2022.

\bibitem[Lin(1993)]{lin1993hierarchical}
L-J Lin.
\newblock Hierarchical learning of robot skills by reinforcement.
\newblock In \emph{IEEE International Conference on Neural Networks}, pages 181--186. IEEE, 1993.

\bibitem[Lin et~al.(2020)Lin, Wang, Olkin, and Held]{corl2020softgym}
Xingyu Lin, Yufei Wang, Jake Olkin, and David Held.
\newblock Softgym: Benchmarking deep reinforcement learning for deformable object manipulation.
\newblock In \emph{Conference on Robot Learning}, 2020.

\bibitem[Lu et~al.(2022)Lu, Kuba, Letcher, Metz, Schroeder~de Witt, and Foerster]{lu2022discovered}
Chris Lu, Jakub Kuba, Alistair Letcher, Luke Metz, Christian Schroeder~de Witt, and Jakob Foerster.
\newblock Discovered policy optimisation.
\newblock \emph{Advances in Neural Information Processing Systems}, 35:\penalty0 16455--16468, 2022.

\bibitem[Makoviychuk et~al.(2021)Makoviychuk, Wawrzyniak, Guo, Lu, Storey, Macklin, Hoeller, Rudin, Allshire, Handa, and State]{makoviychuk2021isaac}
Viktor Makoviychuk, Lukasz Wawrzyniak, Yunrong Guo, Michelle Lu, Kier Storey, Miles Macklin, David Hoeller, Nikita Rudin, Arthur Allshire, Ankur Handa, and Gavriel State.
\newblock Isaac gym: High performance gpu based physics simulation for robot learning.
\newblock In \emph{Neural Information Processing Systems Datasets and Benchmarks Track}, 2021.

\bibitem[Mandlekar et~al.(2021)Mandlekar, Xu, Wong, Nasiriany, Wang, Kulkarni, Fei-Fei, Savarese, Zhu, and Mart\'{i}n-Mart\'{i}n]{mandlekar2021robomimic}
Ajay Mandlekar, Danfei Xu, Josiah Wong, Soroush Nasiriany, Chen Wang, Rohun Kulkarni, Li~Fei-Fei, Silvio Savarese, Yuke Zhu, and Roberto Mart\'{i}n-Mart\'{i}n.
\newblock What matters in learning from offline human demonstrations for robot manipulation.
\newblock In \emph{Conference on Robot Learning}, 2021.

\bibitem[Mattern et~al.(2024)Mattern, Schumacher, L{\'o}pez, Raabe, Ernst, Aubret, and Triesch]{mattern2024mimo}
Dominik Mattern, Pierre Schumacher, Francisco~M L{\'o}pez, Marcel~C Raabe, Markus~R Ernst, Arthur Aubret, and Jochen Triesch.
\newblock Mimo: A multi-modal infant model for studying cognitive development.
\newblock \emph{IEEE Transactions on Cognitive and Developmental Systems}, 2024.

\bibitem[Mees et~al.(2022)Mees, Hermann, Rosete-Beas, and Burgard]{mees2022calvin}
Oier Mees, Lukas Hermann, Erick Rosete-Beas, and Wolfram Burgard.
\newblock Calvin: A benchmark for language-conditioned policy learning for long-horizon robot manipulation tasks.
\newblock \emph{IEEE Robotics and Automation Letters}, 2022.

\bibitem[Merel et~al.(2020)Merel, Tunyasuvunakool, Ahuja, Tassa, Hasenclever, Pham, Erez, Wayne, and Heess]{merel2020catch}
Josh Merel, Saran Tunyasuvunakool, Arun Ahuja, Yuval Tassa, Leonard Hasenclever, Vu~Pham, Tom Erez, Greg Wayne, and Nicolas Heess.
\newblock Catch \& carry: reusable neural controllers for vision-guided whole-body tasks.
\newblock \emph{ACM Transactions on Graphics}, 39\penalty0 (4):\penalty0 39--1, 2020.

\bibitem[Mittendorfer and Cheng(2011)]{mittendorfer2011humanoid}
Philipp Mittendorfer and Gordon Cheng.
\newblock Humanoid multimodal tactile-sensing modules.
\newblock \emph{IEEE Transactions on robotics}, 27\penalty0 (3):\penalty0 401--410, 2011.

\bibitem[Nachum et~al.(2018)Nachum, Gu, Lee, and Levine]{nachum2018data}
Ofir Nachum, Shixiang~Shane Gu, Honglak Lee, and Sergey Levine.
\newblock Data-efficient hierarchical reinforcement learning.
\newblock In \emph{Advances in Neural Information Processing Systems}, pages 3303--3313, 2018.

\bibitem[Narang et~al.(2022)Narang, Storey, Akinola, Macklin, Reist, Wawrzyniak, Guo, Moravanszky, State, Lu, Handa, and Fox]{narang2022factory}
Yashraj Narang, Kier Storey, Iretiayo Akinola, Miles Macklin, Philipp Reist, Lukasz Wawrzyniak, Yunrong Guo, Adam Moravanszky, Gavriel State, Michelle Lu, Ankur Handa, and Dieter Fox.
\newblock Factory: Fast contact for robotic assembly.
\newblock In \emph{Robotics: Science and Systems}, 2022.

\bibitem[OpenAI et~al.(2020)OpenAI, Andrychowicz, Baker, Chociej, Jozefowicz, McGrew, Pachocki, Petron, Plappert, Powell, Ray, Schneider, Sidor, Tobin, Welinder, Weng, and Zaremba]{andrychowicz2020learning}
OpenAI, Marcin Andrychowicz, Bowen Baker, Maciek Chociej, Rafal Jozefowicz, Bob McGrew, Jakub Pachocki, Arthur Petron, Matthias Plappert, Glenn Powell, Alex Ray, Jonas Schneider, Szymon Sidor, Josh Tobin, Peter Welinder, Lilian Weng, and Wojciech Zaremba.
\newblock Learning dexterous in-hand manipulation.
\newblock \emph{The International Journal of Robotics Research}, 39\penalty0 (1):\penalty0 3--20, 2020.

\bibitem[Peng et~al.(2018{\natexlab{a}})Peng, Abbeel, Levine, and Van~de Panne]{peng2018deepmimic}
Xue~Bin Peng, Pieter Abbeel, Sergey Levine, and Michiel Van~de Panne.
\newblock Deepmimic: Example-guided deep reinforcement learning of physics-based character skills.
\newblock \emph{ACM Transactions on Graphics}, 37\penalty0 (4):\penalty0 1--14, 2018{\natexlab{a}}.

\bibitem[Peng et~al.(2018{\natexlab{b}})Peng, Kanazawa, Malik, Abbeel, and Levine]{peng2018sfv}
Xue~Bin Peng, Angjoo Kanazawa, Jitendra Malik, Pieter Abbeel, and Sergey Levine.
\newblock Sfv: Reinforcement learning of physical skills from videos.
\newblock \emph{ACM Transactions on Graphics}, 37\penalty0 (6):\penalty0 1--14, 2018{\natexlab{b}}.

\bibitem[Peng et~al.(2021)Peng, Ma, Abbeel, Levine, and Kanazawa]{peng2021amp}
Xue~Bin Peng, Ze~Ma, Pieter Abbeel, Sergey Levine, and Angjoo Kanazawa.
\newblock Amp: Adversarial motion priors for stylized physics-based character control.
\newblock \emph{ACM Transactions on Graphics}, 40\penalty0 (4):\penalty0 1--20, 2021.

\bibitem[Pertsch et~al.(2020)Pertsch, Lee, and Lim]{pertsch2020spirl}
Karl Pertsch, Youngwoon Lee, and Joseph~J. Lim.
\newblock Accelerating reinforcement learning with learned skill priors.
\newblock In \emph{Conference on Robot Learning}, 2020.

\bibitem[Plappert et~al.(2018)Plappert, Andrychowicz, Ray, McGrew, Baker, Powell, Schneider, Tobin, Chociej, Welinder, Kumar, and Zaremba]{plappert2018robotics}
Matthias Plappert, Marcin Andrychowicz, Alex Ray, Bob McGrew, Bowen Baker, Glenn Powell, Jonas Schneider, Josh Tobin, Maciek Chociej, Peter Welinder, Vikash Kumar, and Wojciech Zaremba.
\newblock Multi-goal reinforcement learning: Challenging robotics environments and request for research.
\newblock \emph{arXiv preprint arXiv:1802.09464}, 2018.

\bibitem[Radosavovic et~al.(2023)Radosavovic, Xiao, Zhang, Darrell, Malik, and Sreenath]{radosavovic2023learning}
Ilija Radosavovic, Tete Xiao, Bike Zhang, Trevor Darrell, Jitendra Malik, and Koushil Sreenath.
\newblock Learning humanoid locomotion with transformers.
\newblock \emph{arXiv preprint arXiv:2303.03381}, 2023.

\bibitem[Raffin et~al.(2019)Raffin, Hill, Ernestus, Gleave, Kanervisto, and Dormann]{raffin2019stable}
Antonin Raffin, Ashley Hill, Maximilian Ernestus, Adam Gleave, Anssi Kanervisto, and Noah Dormann.
\newblock Stable baselines3, 2019.

\bibitem[Schulman et~al.(2017)Schulman, Wolski, Dhariwal, Radford, and Klimov]{schulman2017ppo}
John Schulman, Filip Wolski, Prafulla Dhariwal, Alec Radford, and Oleg Klimov.
\newblock Proximal policy optimization algorithms.
\newblock \emph{arXiv preprint arXiv:1707.06347}, 2017.

\bibitem[Sferrazza and D'Andrea(2022)]{sferrazza2022sim}
Carmelo Sferrazza and Raffaello D'Andrea.
\newblock Sim-to-real for high-resolution optical tactile sensing: From images to three-dimensional contact force distributions.
\newblock \emph{Soft Robotics}, 9\penalty0 (5):\penalty0 926--937, 2022.

\bibitem[Sferrazza et~al.(2023)Sferrazza, Seo, Liu, Lee, and Abbeel]{sferrazza2023power}
Carmelo Sferrazza, Younggyo Seo, Hao Liu, Youngwoon Lee, and Pieter Abbeel.
\newblock The power of the senses: Generalizable manipulation from vision and touch through masked multimodal learning.
\newblock \emph{arXiv preprint arXiv:2311.00924}, 2023.

\bibitem[Srivastava et~al.(2021)Srivastava, Li, Lingelbach, Mart{\'\i}n-Mart{\'\i}n, Xia, Vainio, Lian, Gokmen, Buch, Liu, Savarese, Gweon, Wu, and Fei-Fei]{srivastava2021behavior}
Sanjana Srivastava, Chengshu Li, Michael Lingelbach, Roberto Mart{\'\i}n-Mart{\'\i}n, Fei Xia, Kent~Elliott Vainio, Zheng Lian, Cem Gokmen, Shyamal Buch, Karen Liu, Silvio Savarese, Hyowon Gweon, Jiajun Wu, and Li~Fei-Fei.
\newblock Behavior: Benchmark for everyday household activities in virtual, interactive, and ecological environments.
\newblock In \emph{Conference on Robot Learning}, 2021.

\bibitem[Sutton et~al.(1999)Sutton, Precup, and Singh]{sutton1999between}
Richard~S Sutton, Doina Precup, and Satinder Singh.
\newblock Between mdps and semi-mdps: A framework for temporal abstraction in reinforcement learning.
\newblock \emph{Artificial intelligence}, 112\penalty0 (1-2):\penalty0 181--211, 1999.

\bibitem[Szot et~al.(2021)Szot, Clegg, Undersander, Wijmans, Zhao, Turner, Maestre, Mukadam, Chaplot, Maksymets, Gokaslan, Vondrus, Dharur, Meier, Galuba, Chang, Kira, Koltun, Malik, Savva, and Batra]{szot2021habitat}
Andrew Szot, Alex Clegg, Eric Undersander, Erik Wijmans, Yili Zhao, John Turner, Noah Maestre, Mustafa Mukadam, Devendra Chaplot, Oleksandr Maksymets, Aaron Gokaslan, Vladimir Vondrus, Sameer Dharur, Franziska Meier, Wojciech Galuba, Angel Chang, Zsolt Kira, Vladlen Koltun, Jitendra Malik, Manolis Savva, and Dhruv Batra.
\newblock Habitat 2.0: Training home assistants to rearrange their habitat.
\newblock In \emph{Neural Information Processing Systems}, 2021.

\bibitem[Tassa et~al.(2018)Tassa, Doron, Muldal, Erez, Li, de~Las~Casas, Budden, Abdolmaleki, Merel, Lefrancq, Lillicrap, and Riedmiller]{tassa2018deepmind}
Yuval Tassa, Yotam Doron, Alistair Muldal, Tom Erez, Yazhe Li, Diego de~Las~Casas, David Budden, Abbas Abdolmaleki, Josh Merel, Andrew Lefrancq, Timothy~P. Lillicrap, and Martin~A. Riedmiller.
\newblock Deepmind control suite.
\newblock \emph{arXiv preprint arXiv:1801.00690}, 2018.

\bibitem[Tassa et~al.(2020)Tassa, Tunyasuvunakool, Muldal, Doron, Liu, Bohez, Merel, Erez, Lillicrap, and Heess]{tassa2020dm_control}
Yuval Tassa, Saran Tunyasuvunakool, Alistair Muldal, Yotam Doron, Siqi Liu, Steven Bohez, Josh Merel, Tom Erez, Timothy Lillicrap, and Nicolas Heess.
\newblock dm\_control: Software and tasks for continuous control.
\newblock \emph{arXiv preprint arXiv:2006.12983}, 2020.

\bibitem[Todorov et~al.(2012)Todorov, Erez, and Tassa]{todorov2012mujoco}
Emanuel Todorov, Tom Erez, and Yuval Tassa.
\newblock Mujoco: A physics engine for model-based control.
\newblock In \emph{IEEE/RSJ International Conference on Intelligent Robots and Systems}, pages 5026--5033, 2012.

\bibitem[Wang et~al.(2023)Wang, Lin, Zeng, Luo, Zhang, and Zhang]{wang2023physhoi}
Yinhuai Wang, Jing Lin, Ailing Zeng, Zhengyi Luo, Jian Zhang, and Lei Zhang.
\newblock Physhoi: Physics-based imitation of dynamic human-object interaction.
\newblock \emph{arXiv preprint arXiv:2312.04393}, 2023.

\bibitem[Wei et~al.(2022)Wei, Liu, Ling, and Su]{wei2022coacd}
Xinyue Wei, Minghua Liu, Zhan Ling, and Hao Su.
\newblock Approximate convex decomposition for 3d meshes with collision-aware concavity and tree search.
\newblock \emph{ACM Transactions on Graphics (TOG)}, 41\penalty0 (4):\penalty0 1--18, 2022.

\bibitem[Xie et~al.(2023)Xie, Tseng, Starke, van~de Panne, and Liu]{xie2023hierarchical}
Zhaoming Xie, Jonathan Tseng, Sebastian Starke, Michiel van~de Panne, and C~Karen Liu.
\newblock Hierarchical planning and control for box loco-manipulation.
\newblock \emph{Symposium on Computer Animation}, 2023.

\bibitem[Yu et~al.(2019)Yu, Quillen, He, Julian, Hausman, Finn, and Levine]{yu2019meta}
Tianhe Yu, Deirdre Quillen, Zhanpeng He, Ryan Julian, Karol Hausman, Chelsea Finn, and Sergey Levine.
\newblock Meta-world: A benchmark and evaluation for multi-task and meta reinforcement learning.
\newblock In \emph{Conference on Robot Learning}, 2019.

\bibitem[Yuan et~al.(2023)Yuan, Che, Qin, Huang, Yin, Lee, Wu, Lim, and Wang]{yuan2023robot}
Ying Yuan, Haichuan Che, Yuzhe Qin, Binghao Huang, Zhao-Heng Yin, Kang-Won Lee, Yi~Wu, Soo-Chul Lim, and Xiaolong Wang.
\newblock Robot synesthesia: In-hand manipulation with visuotactile sensing.
\newblock \emph{arXiv preprint arXiv:2312.01853}, 2023.

\bibitem[Zakka et~al.(2022)Zakka, Tassa, and {MuJoCo Menagerie Contributors}]{menagerie2022github}
Kevin Zakka, Yuval Tassa, and {MuJoCo Menagerie Contributors}.
\newblock {MuJoCo Menagerie: A collection of high-quality simulation models for MuJoCo}, 2022.
\newblock URL \url{http://github.com/google-deepmind/mujoco_menagerie}.

\bibitem[Zakka et~al.(2023)Zakka, Wu, Smith, Gileadi, Howell, Peng, Singh, Tassa, Florence, Zeng, and Abbeel]{zakka2023robopianist}
Kevin Zakka, Philipp Wu, Laura Smith, Nimrod Gileadi, Taylor Howell, Xue~Bin Peng, Sumeet Singh, Yuval Tassa, Pete Florence, Andy Zeng, and Pieter Abbeel.
\newblock Robopianist: Dexterous piano playing with deep reinforcement learning.
\newblock In \emph{Conference on Robot Learning}, pages 2975--2994. PMLR, 2023.

\bibitem[Zhang et~al.(2023)Zhang, Yuan, Makoviychuk, Guo, Fidler, Peng, and Fatahalian]{zhang2023learning}
Haotian Zhang, Ye~Yuan, Viktor Makoviychuk, Yunrong Guo, Sanja Fidler, Xue~Bin Peng, and Kayvon Fatahalian.
\newblock Learning physically simulated tennis skills from broadcast videos.
\newblock \emph{ACM Transactions on Graphics}, 42\penalty0 (4):\penalty0 1--14, 2023.

\bibitem[Zhao et~al.(2023)Zhao, Kumar, Levine, and Finn]{zhao2023learning}
Tony~Z Zhao, Vikash Kumar, Sergey Levine, and Chelsea Finn.
\newblock Learning fine-grained bimanual manipulation with low-cost hardware.
\newblock In \emph{Robotics: Science and Systems}, 2023.

\bibitem[Zhu et~al.(2020)Zhu, Wong, Mandlekar, and Mart\'{i}n-Mart\'{i}n]{zhu2020robosuite}
Yuke Zhu, Josiah Wong, Ajay Mandlekar, and Roberto Mart\'{i}n-Mart\'{i}n.
\newblock robosuite: A modular simulation framework and benchmark for robot learning.
\newblock \emph{arXiv preprint arXiv:2009.12293}, 2020.

\bibitem[Zhuang et~al.(2023)Zhuang, Fu, Wang, Atkeson, Schwertfeger, Finn, and Zhao]{zhuang2023robot}
Ziwen Zhuang, Zipeng Fu, Jianren Wang, Christopher~G Atkeson, S{\"o}ren Schwertfeger, Chelsea Finn, and Hang Zhao.
\newblock Robot parkour learning.
\newblock In \emph{Conference on Robot Learning}, 2023.

\end{thebibliography}

\clearpage
\appendices

\begin{center}
{\huge \textbf{Appendix}}    
\end{center}

\addtocontents{toc}{\protect\setcounter{tocdepth}{2}} 
\startcontents
\printcontents{}{1}{\textbf{Table of Contents}\vskip3pt\hrule\vskip5pt} \vskip3pt\hrule\vskip5pt
\section{Additional Components}

\label{sec:additional_tasks}

\label{sec:additional_tasks:digit}
\begin{figure}[t]
    \centering
    \begin{subfigure}{0.49\linewidth}
        \includegraphics[width=\textwidth]{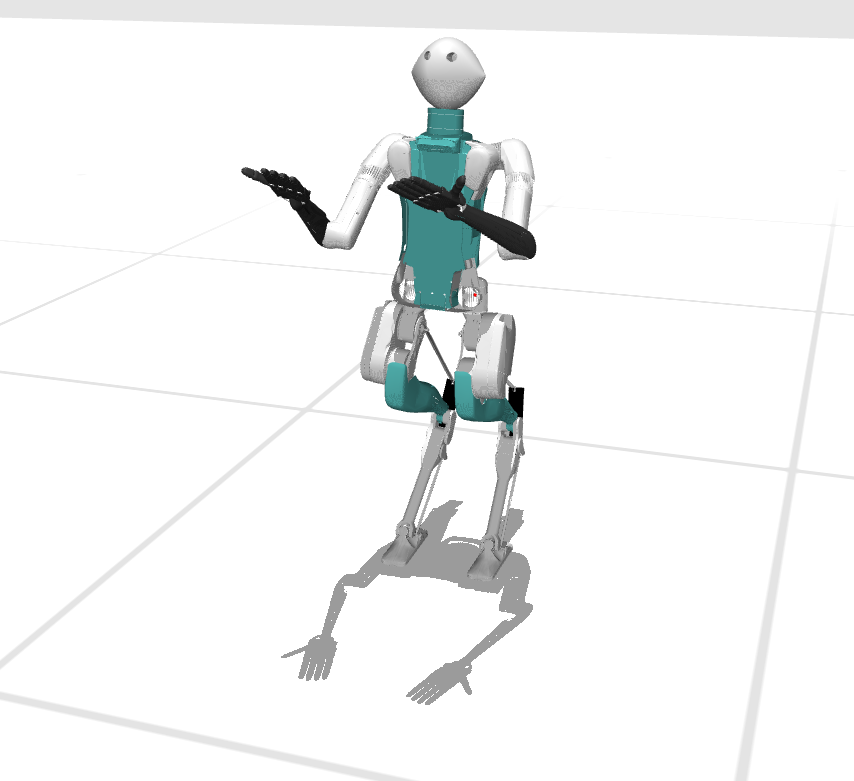}
        \caption{Digit w/ Shadow Hands}
    \end{subfigure}
    \begin{subfigure}{0.4\linewidth}\includegraphics[width=\textwidth]{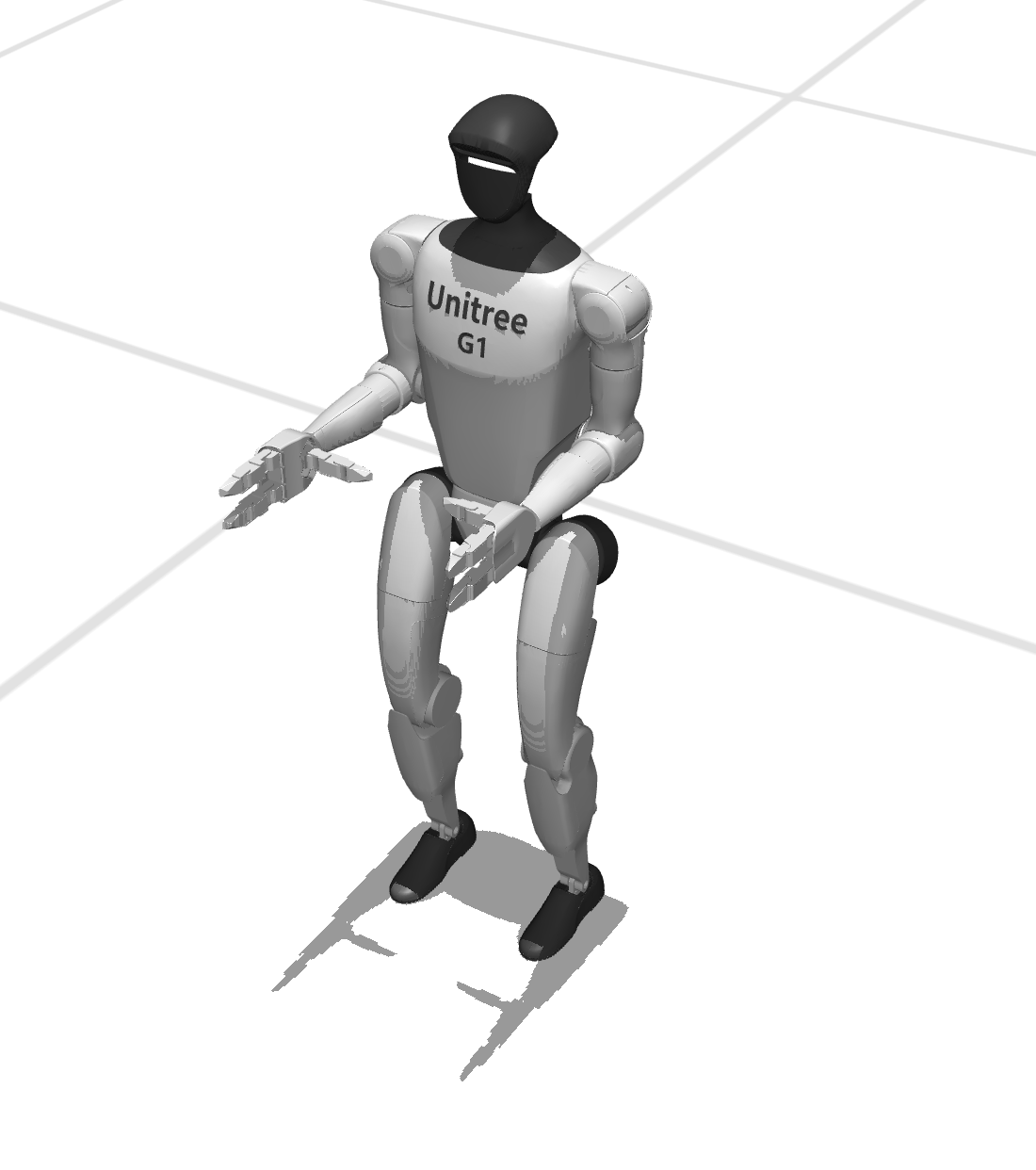}
        \caption{Unitree G1}
    \end{subfigure}
    \\
    \begin{subfigure}{0.45\linewidth}
        \includegraphics[width=\textwidth]{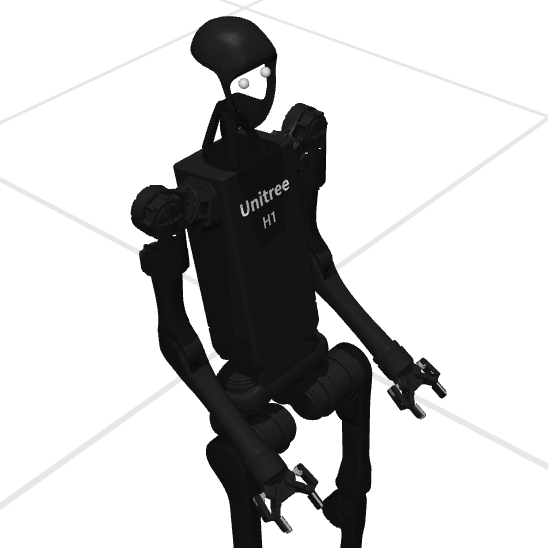}
        \caption{H1 w/ Robotiq grippers}
    \end{subfigure}
    \begin{subfigure}{0.45\linewidth}
        \includegraphics[width=\textwidth]{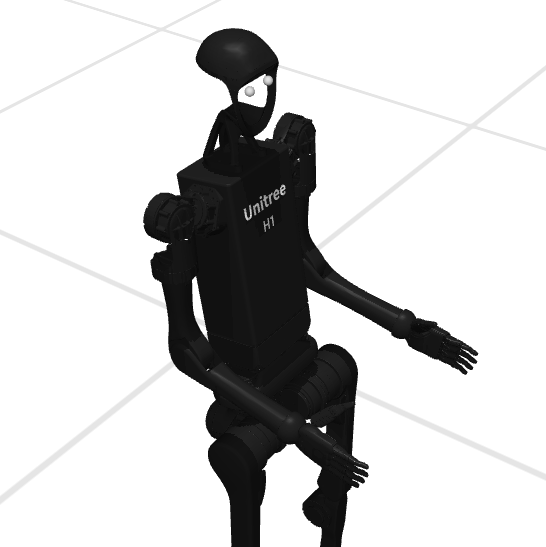}
        \caption{H1 w/ Unitree hands}
    \end{subfigure}
    \caption{Additional configurations available in HumanoidBench.}
    \label{fig:additional_robots}
\end{figure}
In addition to the Unitree H1 robot, we also include model files for the Agility Robotics Digit humanoid (see \Cref{fig:additional_robots}). We add a custom head, together with egocentric vision and whole-body tactile sensing, similarly to the H1. We provide a torque controlled version of this robot. We also provide a torque controlled version of the more recent Unitree G1 humanoid.

For the broader use of HumanoidBench, we also provide two simple end-effector options for the Unitree H1 robot: a $2$-DoF Robotiq 2F-45 parallel-jaw gripper (with a rotational wrist joint), and a $13$-DoF dexterous Unitree hand available in the Unitree model collection.\cref{footnote:h1_github}

Specifications for all these variants are detailed in \Cref{tab:add_robot_spec}.

\begin{table*}[t]
    \centering
    \begin{tabular}{ccccccc}
        \toprule
            & H1 w/o hands    & H1 w/ ShadowHand   & H1 w/ Robotiq gripper     & H1 w/ Unitree hand  & Digit w/ ShadowHand & Unitree G1  \\
        \midrule
        Observation space           & $51$      & $151$     & $55$  & $103$     & $221$ & $87$\\
        Action space                & $19$      & $61$      & $23$  & $45$      & $65$  & $37$\\
        DoF (body)                  & $25$      & $25$      & $25$  & $25$      & $57$ & $29$ \\
        DoF ($2$ end-effectors)     & $0$       & $50$      & $4$   & $26$      & $50$ & $14$ \\
        \bottomrule
    \end{tabular}
    \caption{\textbf{All supported robot specifications}. Note that the observation space in this table does not take into account any observations of the surrounding environment and solely comprises generalized positions and velocities. We use quaternions for the robot floating base orientation, as well as for ball joints, which add additional position coordinates compared to the velocity components (which match the DoFs).}
    \label{tab:add_robot_spec}
\end{table*}

\section{Simulated Environment Details}
\label{sec:environment_details}

\subsection{Observation Space}

The observation space for the benchmarked H1 robot state (joint positions and velocities) comprises $151$ dimensions, with $51$ dimensions representing the humanoid robot body and $50$ dimensions representing each hand. \Cref{tab:add_robot_spec} summarizes observation spaces for other robotic configurations supported in HumanoidBench.
The observation space can also include environment states for tasks interacting with objects, as described in detail in \Cref{sec:environment_details:task_specification}.

\subsection{Action Space}

The action space is the same across all environments. We normalize the action space to be $[-1,1]^{\lVert A \rVert}$, where $\lVert A \rVert = 61$ ($19$ for the humanoid body and $21$ for each hand). We use position control for the benchmarking results in this paper.

\subsection{Simulation Performance}

HumanoidBench is based on MuJoCo~\citep{todorov2012mujoco}, which provides fast and accurate physics simulation. We provide a diverse set of configurations, such as the full humanoid model with two hands, the humanoid model without hands, and the humanoid model with collision meshes only on its feet. We use the fastest model (i.e., collision meshes only on feet) to train the low-level reaching policies, described in \Cref{sec:training_details:reaching_policy}. 

We benchmark our HumanoidBench simulation using the Unitree H1 model and report its performance in \Cref{tab:simulation_performance}. Even with complex humanoid body and dexterous hands, HumanoidBench can run $1000$+ FPS on a single CPU with a simulation timestep of \SI{0.002}{s}.

\begin{table}[ht]
    \centering
    \begin{tabular}{lc}
        \toprule
        \textbf{Configuration} & \textbf{FPS} \\
        \midrule
        Without hands & $2450$ \\
        Simplified body collisions & $3600$ \\
        Collisions only for feet & $5100$ \\        
        \midrule
        \textbf{Default} & $1050$ \\
        \bottomrule
    \end{tabular}
    \caption{\textbf{HumanoidBench Simulation Performance.}}
    \label{tab:simulation_performance}
\end{table}

\subsection{Whole-body Tactile Sensing}
\label{sec:environment_details:tactile}
\begin{figure}[t]
    \centering
    \begin{subfigure}[t]{0.6\linewidth}
        \includegraphics[width=\textwidth]{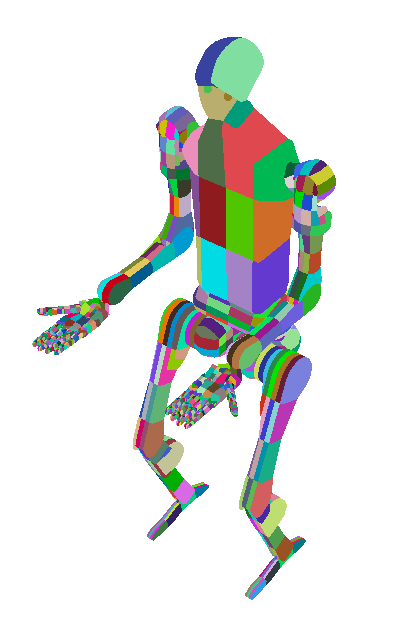}
        \vspace{-3.0em}
        \caption{Refined meshes}
    \end{subfigure}
    \\ \vspace{0.5cm}
    \begin{subfigure}[t]{0.42\linewidth}
        \includegraphics[width=\textwidth]{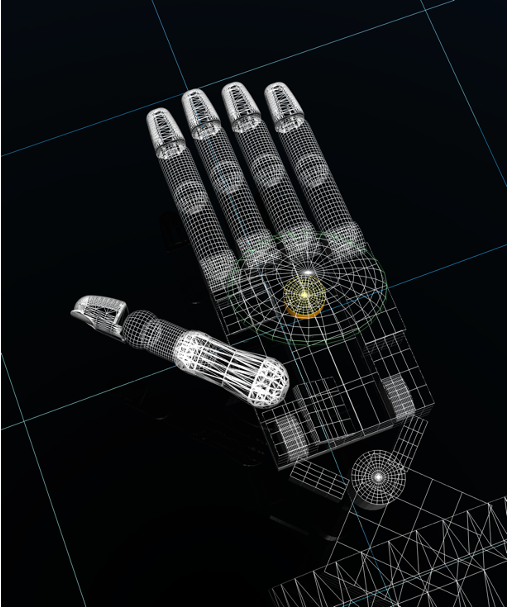}
        \vspace{-1.7em}
        \caption{Contact before refinement}
    \end{subfigure}
    \hspace{0.5cm}
    \begin{subfigure}[t]{0.42\linewidth}
        \includegraphics[width=\textwidth]{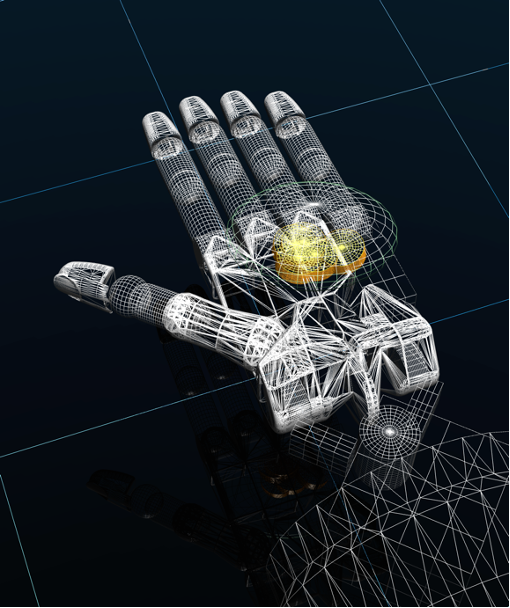}
        \vspace{-1.7em}
        \caption{Contact after refinement}
    \end{subfigure}
 
    \caption{\textbf{Refinement of collision meshes.} As a result of the mesh refinement (see (a), where each colored section indicates a different collision mesh), our model can detect a higher number of contact points, as shown by the yellow disks in the figure. This results in a better spatial discretization of the tactile readings.}
    \label{fig:convexification}
\end{figure}

We implement whole-body tactile sensing by employing MuJoCo touch grid, which aggregates pressure and shear contact forces into discrete bins. Similar distributed force readings have been captured on real-world systems both on humanoid bodies \cite{mittendorfer2011humanoid} and end-effectors \citep{sferrazza2022sim}. To increase the number of contact point candidates and fully exploit the spatial resolution of the touch grid, we subdivide the original meshes into many smaller meshes. Specifically, we build on top of CoACD \citep{wei2022coacd}, which in addition makes sure that the resulting meshes are all convex. This procedure results in finer contact resolution when the MuJoCo physics engine computes collisions. An example is depicted in \Cref{fig:convexification}, with a larger number of contact points generally resulting in a considerably better discretization of the tactile readings. The full model with refined meshes and tactile readings runs at 550 FPS.

\subsection{Task Specification}
\label{sec:environment_details:task_specification}

Before enumerating the environment details below, let us define auxiliary functions and variables that are employed in the reward functions of multiple environments:
\begin{itemize}
    \item $tol\left(x, (x_\text{lower},x_\text{upper}), m\right)$ is a function provided in the DeepMind Control Suite package \cite{tassa2020dm_control}. This is denoted there as a \textit{tolerance} function that returns $1$ when the evaluated value is within the bounds, i.e., $x \in (x_\text{lower},x_\text{upper})$, and between $0$ and $1$ otherwise. The margin $m$ regulates the slope of the function, i.e., how far from the bounds the function approaches $0$.
    \item $height\left( (x_\text{lower},x_\text{upper}), m\right)$ is a variable defined as $tol\left(z_\text{head}, (x_\text{lower},x_\text{upper}), m\right)$ that rewards the head height,  where $z_\text{head}$ is the vertical coordinate of the robot head position. Whenever the arguments are not indicated, we assume $x_\text{lower}=1.65$, $x_\text{upper} = +\infty$, and $m=0.4125$.
    \item $d(object_\text{A}, object_\text{B})$ is the distance between object A and object B.
    \item $upright\left((x_\text{lower},x_\text{upper}), m\right)$ is a variable defined as $tol\left(z_\text{proj}, (x_\text{lower},x_\text{upper}), m\right)$, which rewards the alignment of the robot torso with respect to the vertical axis, where $z_\text{proj}$ is the unit projection of the $z$-axis in the robot body frame onto the $z$-axis in the global frame. Whenever the arguments are not indicated, we assume $x_\text{lower}=0.9$, $x_\text{upper} = +\infty$, and $m=1.9$.
    \item $stand := height \times upright$ represents a standing posture reward.
    \item $e := 0.2 \cdot \left[4 + \frac{1}{|u|} \sum_i tol(u_i, (0, 0), 10)\right]$ rewards small control effort, where $u$ is the vector of actuation inputs.
    \item $stable = stand \times e$, densely rewards stable standing configurations.
    \item $v_x$ is the robot velocity in the $x$ direction (positive forward) of the robot body coordinate frame.
    \item $v_y$ is the robot velocity in the $y$ direction (positive left) of the robot body coordinate frame.
    \item $z_\text{item}$ is the vertical $z$ coordinate  of the indicated `item' in the global frame.
    \item $pos_\text{item}$ is the 3D position  of the indicated `item' in the global frame.
\end{itemize}

\subsubsection{\texttt{walk}}
\begin{center}
    \includegraphics[width=0.4\linewidth]{fig/tasks/0_h1hand-walk-v0-middle.png}
\end{center}

\textbf{\textit{Objective.}} Keep forward velocity close to \SI{1}{\meter/\second} without falling to the ground.

\textbf{\textit{Observation.}} Joint positions and velocities of the robot.

\textbf{\textit{Initialization.}} The robot is initialized to a standing position, with random noise added to all joint positions during each episode reset.

\textbf{\textit{Termination.}} The episode terminates after 1000 steps, or when $z_\text{pelvis} < 0.2$.

\textbf{\textit{Reward Implementation.}} \[R(s, a) = stable \times tol(v_x, (1, +\infty), 1).\]

\subsubsection{\texttt{stand}}
\begin{center}
    \includegraphics[width=0.4\linewidth]{fig/tasks/1_h1hand-stand-v0-solved.png}
\end{center}

\textbf{\textit{Objective.}} Maintain a standing pose.

\textbf{\textit{Observation.}} Joint positions and velocities of the robot.

\textbf{\textit{Initialization.}} The robot is initialized to a standing position, with random noise added to all joint positions during each episode reset.

\textbf{\textit{Reward Implementation.}} Let:
\begin{itemize}
    \item $still_x = tol(v_x, (0, 0), 2)$
    \item $still_y = tol(v_y, (0, 0), 2)$,
\end{itemize}
Then:
\[R(s, a) = stable \times mean(still_x, still_y)\]

\textbf{\textit{Termination.}} The episode terminates after 1000 steps, or when $z_\text{pelvis} < 0.2$.

\subsubsection{\texttt{run}}
\begin{center}
    \includegraphics[width=0.4\linewidth]{fig/tasks/2_h1hand-run-v0-middle.png}
\end{center}

\textbf{\textit{Objective.}} Keep forward velocity close to \SI{5}{\meter/\second} without falling to the ground.

\textbf{\textit{Observation.}} Joint positions and velocities of the robot.

\textbf{\textit{Initialization.}} The robot is initialized to a standing position, with random noise added to all joint positions during each episode reset.

\textbf{\textit{Reward Implementation.}} \[R(s, a) = stable \times tol(v_x, (5, \infty), 5)\]

\textbf{\textit{Termination.}} The episode terminates after 1000 steps, or when $z_\text{pelvis} < 0.2$.

\subsubsection{\texttt{sit}}
\begin{center}
    \includegraphics[width=0.4\linewidth]{fig/tasks/3_h1hand-sit_simple-v0-solved.png}
\end{center}

\textbf{\textit{Objective.}} Sit onto a chair situated closely behind the robot.

\textbf{\textit{Observation.}} In \texttt{sit\_simple}, the observation is a vector containing all joint positions on the robot unit. In \texttt{sit\_hard}, we allow the chair to be moved, and the robot initial position is randomized, so the observation includes position and orientation of the chair as well.

\textbf{\textit{Initialization.}} The robot is initialized to a standing position, with random noise added to all joint positions during each episode reset. Note that in \texttt{sit\_hard}, the robot is rotated at a random angle $\alpha \in [-1.8\ \rm{rad}, 1.8\ \rm{rad}]$ with position initialized at random values $x \in [0.2, 0.4], y \in [-0.15, 0.15]$

\textbf{\textit{Reward Implementation.}} Let
\begin{itemize}
    \item $sitting_x = tol(x_{robot} - x_{chair}, (-0.19, 0.19), 0.2$
    \item $sitting_y = tol(y_{robot} - y_{chair}, (0, 0), 0.1$
    \item $sitting_z = tol(z_{robot}, (0.68, 0.72), 0.2)$
    \item $posture = tol(z_\text{head} - z_\text{IMU}, (0.35, 0.45), 0.3)$
    \item $still_x = tol(v_x, (0, 0), 2)$
    \item $still_y = tol(v_y, (0, 0), 2)$
\end{itemize}
then, the reward of this task is
\begin{align*}
    R(s, a) =
    (&(0.5 \cdot sitting_z + 0.5 \cdot sitting_x \times sitting_y) \\
    &\times upright \times posture) \\
    &\times e \times mean(still_x, still_y)
\end{align*}

\textbf{\textit{Termination.}} The episode terminates after 1000 steps, or when $z_\text{pelvis} < 0.5$.

\subsubsection{\texttt{balance}}
\begin{center}
    \includegraphics[width=0.4\linewidth]{fig/tasks/4_h1hand-balance_hard-v0-middle.png}
\end{center}

\textbf{\textit{Objective.}} Balance on the unstable board. There are two variants to this environment: the \texttt{balance\_simple} variant's spherical pivot beneath the board does not move, while the \texttt{balance\_hard} variant's pivot does.

\textbf{\textit{Observation.}} The observation comprises the joint positions and velocities of the robot and those of the board.

\textbf{\textit{Initialization.}} The robot is initialized to a standing position on the unstable board, with random noise added to all joint positions during each episode reset.

\textbf{\textit{Reward Implementation.}} Let
\begin{itemize}
    \item $still_x = tol(v_x, (0, 0), 2)$
    \item $still_y = tol(v_y, (0, 0), 2)$
    \item $still = \frac{1}{2}(still_x + still_y)$
    \item $height_\text{robot} = height((2.15, +\infty))$
\end{itemize}
Then:
\[R(s, a) = (e \times still) \times (height_\text{robot} \times upright)\]

\textbf{\textit{Termination.}} The episode terminates after 1000 steps, or when one of the following condition satisfies:
\begin{itemize}
    \item $z_{robot} < 0.8$
    \item The sphere collides with anything other than the floor and the standing board.
    \item The standing board collides with the floor.
\end{itemize}

\subsubsection{\texttt{stair}}
\begin{center}
    \includegraphics[width=0.4\linewidth]{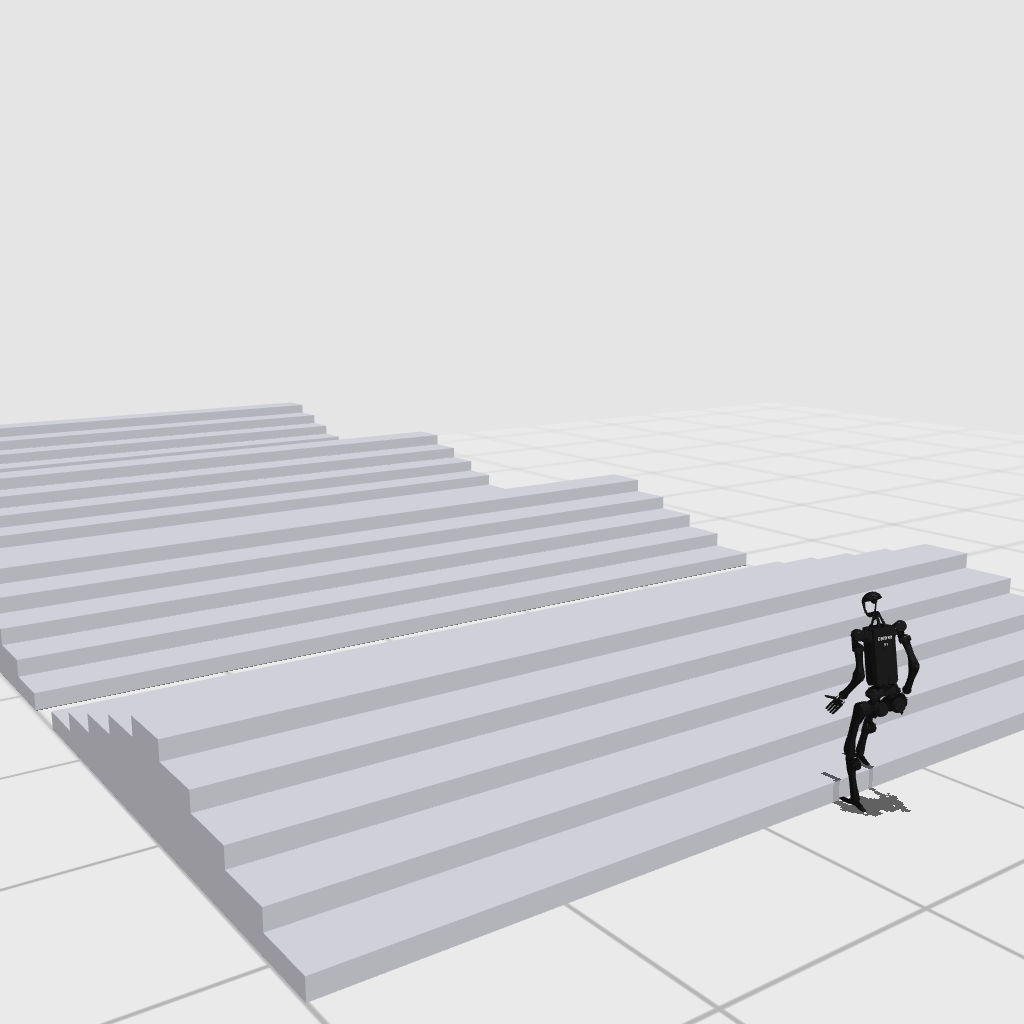}
\end{center}

\textbf{\textit{Objective.}}  Traverse an iterating sequence of upward and downward stairs at \SI{1}{\meter/\second}.

\textbf{\textit{Observation.}} Joint positions and velocities of the robot.

\textbf{\textit{Initialization.}} The robot is initialized to a standing position, with random noise added to all joint positions during each episode reset.

\textbf{\textit{Reward Implementation.}} Let
\begin{itemize}
    \item $vertical_\text{foot,left} = tol(z_\text{head} - z_\text{foot,left}, (1.2, +\infty), 0.45)$
    \item $vertical_\text{foot,right} = tol(z_\text{head} - z_\text{foot,right}, (1.2, +\infty), 0.45)$
\end{itemize}
Then:
\begin{align*}
    R(s, a)
    = e &\times tol(v_x, (1, +\infty), 1) \times upright((0.5, 1), 1.9) \\
    &\times (vertical_\text{foot,left} \times vertical_\text{foot,right})
\end{align*}

\textbf{\textit{Termination.}} The episode terminates after 1000 steps, or when $z_\text{proj} < 0.1$.

\subsubsection{\texttt{slide}}
\begin{center}
    \includegraphics[width=0.4\linewidth]{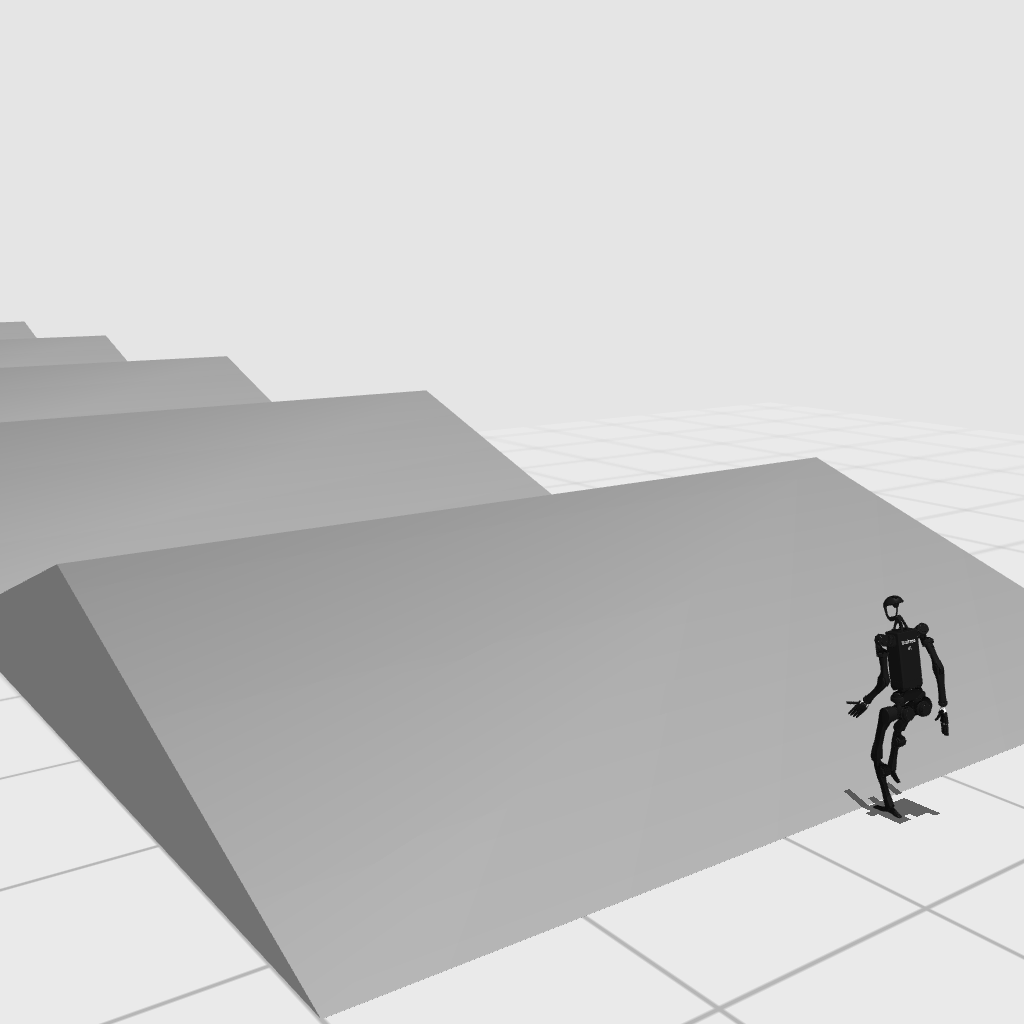}
\end{center}

\textbf{\textit{Objective.}} Walk over an iterating sequence of upward and downward slides at \SI{1}{\meter/\second}.

\textbf{\textit{Observation.}} Joint positions and velocities of the robot.

\textbf{\textit{Initialization.}} The robot is initialized to a standing position, with random noise added to all joint positions during each episode reset.

\textbf{\textit{Reward Implementation.}} Let
\begin{itemize}
    \item $vertical_\text{foot,left} = tol(z_\text{head} - z_\text{foot,left}, (1.2, +\infty), 0.45)$
    \item $vertical_\text{foot,right} = tol(z_\text{head} - z_\text{foot,right}, (1.2, +\infty), 0.45)$
\end{itemize}
Then:
\begin{align*}
    R(s, a)
    = e &\times tol(v_x, (1, +\infty), 1) \times upright((0.5, 1), 1.9) \\
    &\times (vertical_\text{foot,left} \times vertical_\text{foot,right})
\end{align*}

\textbf{\textit{Termination.}} The episode terminates after 1000 steps, or when $z_\text{proj} < 0.6$.

\subsubsection{\texttt{pole}}
\begin{center}
    \includegraphics[width=0.4\linewidth]{fig/tasks/7_h1hand-pole-v0-middle.png}
\end{center}

\textbf{\textit{Objective.}} Travel in forward direction over a dense forest of high thin poles, without colliding with them.

\textbf{\textit{Observation.}} Joint positions and velocities of the robot.

\textbf{\textit{Initialization.}} The robot is initialized to a standing position, with random noise added to all joint positions during each episode reset.

\textbf{\textit{Reward Implementation.}} Let
\begin{itemize}
    \item $\gamma_\text{collision} = \begin{cases}
        0.1, &,\text{robot collides with pole} \\
        1, & \text{otherwise}
    \end{cases}$
\end{itemize}
Then,
\[R(s, a) = \gamma_\text{collision} \times (0.5 \cdot stable + 0.5 \cdot tol(v_x, (1, +\infty), 1)).\] 

\textbf{\textit{Termination.}} The episode terminates after 1000 steps, or when $z_\text{pelvis} < 0.6$.

\subsubsection{\texttt{reach}}
\begin{center}
    \includegraphics[width=0.4\linewidth]{fig/tasks/8_h1hand-reach-v0-solved.png}
\end{center}

\textbf{\textit{Objective.}} Reach a randomly initialized 3D point with the left hand.

\textbf{\textit{Observation.}} Joint positions and velocities of the robot, left hand position of robot, and reaching target position.

\textbf{\textit{Initialization.}} The robot is initialized to a standing position, with random noise added to all joint positions during each episode reset.

\textbf{\textit{Reward Implementation.}} Let
\begin{itemize}
    \item $d_\text{hand} = d(hand_\text{left}, goal)$
    \item $health = 5 \cdot z_{pelvis,proj}$
    \item $penalty_{motion} = \| v_{robot} \|^2$
    \item $close = \begin{cases}
        5, &d_\text{hand} < 1 \\
        0, &\text{otherwise}
    \end{cases}$
    \item $success = \begin{cases}
        10, &d_\text{hand} < 0.05 \\
        0, &\text{otherwise}
    \end{cases}$
\end{itemize}
Then,
\[
    R(s, a) = - 10^{-4} \cdot penalty_{motion} + health + close + success
\]

\textbf{\textit{Termination.}} The episode terminates after 1000 steps.

\subsubsection{\texttt{hurdle}}
\begin{center}
    \includegraphics[width=0.4\linewidth]{fig/tasks/9_h1hand-hurdle-v0-solved.png}
\end{center}

\textbf{\textit{Objective.}} Keep forward velocity close to \SI{5}{\meter/\second} without falling to the ground.

\textbf{\textit{Observation.}} Joint positions and velocities of the robot.

\textbf{\textit{Initialization.}} The robot is initialized to a standing position, with random noise added to all joint positions during each episode reset.

\textbf{\textit{Reward Implementation.}} Let
\[
    \gamma_\text{collision} = \begin{cases} 0.1 &,\text{robot is colliding with wall} \\ 1 &,\text{otherwise}\end{cases}
\]
Then, we formulate the reward of this task as:
\[R(s, a) = stable \times tol(v_x, (5, \infty), 5) \times \gamma_\text{collision}\]

\textbf{\textit{Termination.}} The episode terminates after 1000 steps.

\subsubsection{\texttt{crawl}}
\begin{center}
    \includegraphics[width=0.4\linewidth]{fig/tasks/10_h1hand-crawl-v0-solved.png}
\end{center}

\textbf{\textit{Objective.}} Keep forward velocity close to \SI{1}{\meter/\second} while passing inside a tunnel.

\textbf{\textit{Observation.}} Joint positions and velocities of the robot.

\textbf{\textit{Initialization.}} The robot is initialized to a standing position, with random noise added to all joint positions during each episode reset.

\textbf{\textit{Reward Implementation.}} Let:
\begin{itemize}
    \item $height_\text{crawl} = height((0.6, 1), 1)$
    \item $height_\text{IMU} = tol(z_\text{IMU}, (0.6, 1), 1)$
    \item $quat_\text{crawl} = \begin{bmatrix} 0.75 & 0 & 0.65 & 0 \end{bmatrix}$ is the expected quaternion expected when the robot is crawling.
    \item $orientation = tol(\|quat_\text{pelvis} - quat_\text{crawl}\|, (0, 0), 1)$ rewards correct robot body orientation.
    \item $tunnel =tol(y_\text{IMU}, (-1, 1), 0)$ rewards the robot when its $y$ coordinate is inside the tunnel.
    \item $speed = tol(v_x, (1, +\infty), 1)$
\end{itemize}
Then, the reward is formulated as:
\begin{align*}
    R = tunnel \times (0.1 \cdot e &+ 0.25 \cdot \min(height_\text{crawl}, height_\text{IMU}) \\
    &+ 0.25 \cdot orientation + 0.4 \cdot speed)
\end{align*}

\textbf{\textit{Termination.}} The episode terminates after 1000 steps.

\subsubsection{\texttt{maze}}
\begin{center}
    \includegraphics[width=0.4\linewidth]{fig/tasks/11_h1hand-maze-v0-middle_opaque.png}
\end{center}

\textbf{\textit{Objective.}} Reach the goal position in a maze by taking multiple turns at the intersections.

\textbf{\textit{Observation.}} Joint positions and velocities of the robot.

\textbf{\textit{Initialization.}} The robot is initialized to a standing position, with random noise added to all joint positions during each episode reset.

\textbf{\textit{Reward Implementation.}} For each checkpoint in the maze, we assign $v_{target}$ to be the velocity prescribed to travel from the previous checkpoint towards next one. Then, let
\begin{itemize}
    \item $move =$
    \begin{align*}
        &tol(v_x - v_{target_x}, (0, 0), v_{target_x}) \\
        &\times tol(v_y - v_{target_y}, (0, 0), v_{target_y})
    \end{align*}
    \item $\gamma_\text{collision} = \begin{cases} 0.1 &,\text{robot is colliding with wall} \\ 1 &,\text{otherwise}\end{cases}$
    \item $proximity = tol(d(checkpoint, pos_{robot}), (0, 0), 1)$
\end{itemize}
We formulate the reward as
\begin{align*}
    R(s, a)
    = (&0.2 \cdot stable + 0.4 \cdot move  \\
    &+ 0.4 \cdot proximity) \times \gamma_\text{collision}
\end{align*}
We also provide a sparse reward of $i \times 100$ for arriving at the $i^{th}$ checkpoint.

\textbf{\textit{Termination.}} The episode terminates after 1000 steps, or when $z_\text{pelvis} < 0.2$.

\subsubsection{\texttt{push}}
\begin{center}
    \includegraphics[width=0.4\linewidth]{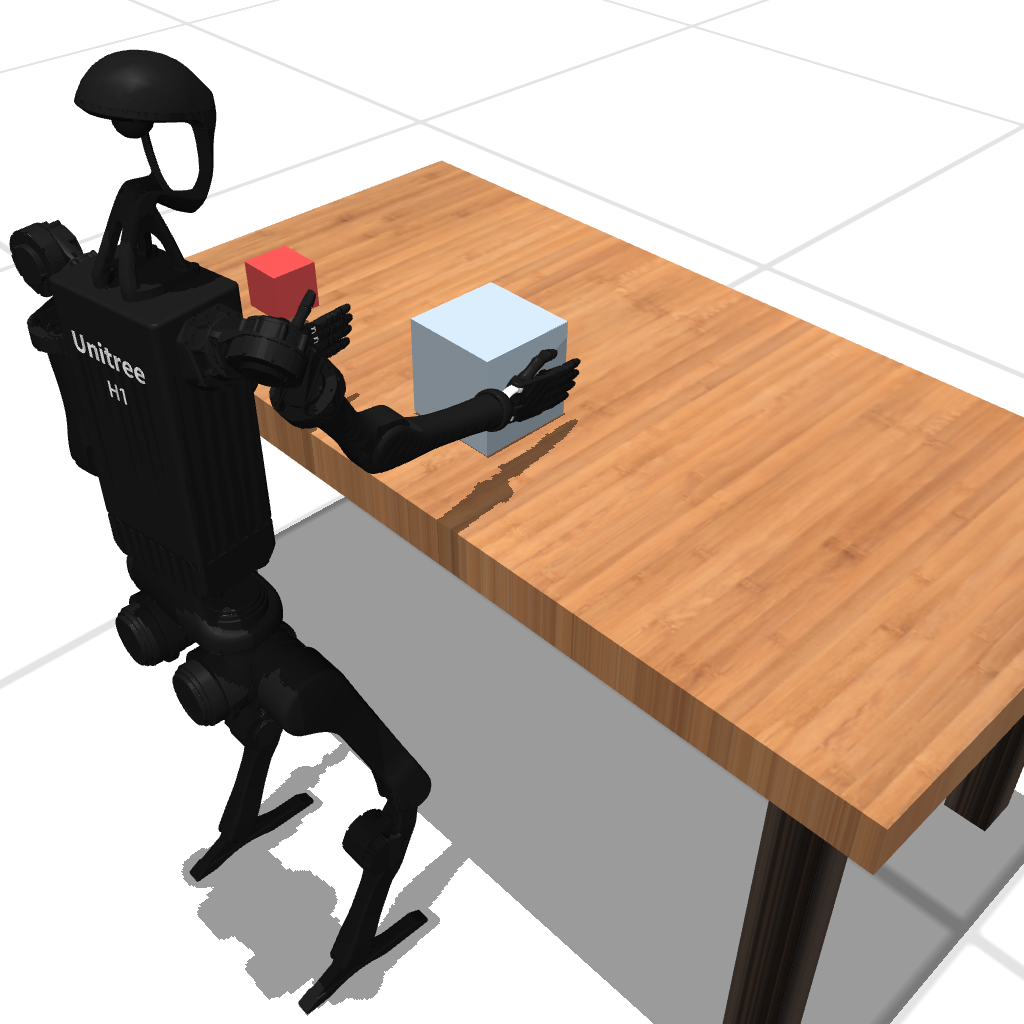}
\end{center}

\textbf{\textit{Objective.}} Move a box to a randomly initialized 3D point on a table.

\textbf{\textit{Observation.}} Joint positions and velocities of the robot, left hand position of the robot, box destination, box position, and box velocity.

\textbf{\textit{Initialization.}} The robot is initialized to a standing position. The box and its destination are initialized at a random location on the table. Random noise is added to all joint positions during each episode reset.

\textbf{\textit{Reward Implementation.}} Let:
\begin{itemize}
    \item $d_\text{goal} = d(box, destination)$
    \item $success = \mathbbm{1}_{d_\text{goal} < 0.05}$
    \item $d_\text{hand} = d(box,hand_\text{left})$
\end{itemize}
Then the reward is
\[
    R(s, a) = \alpha_s \cdot success - \alpha_t \cdot d_\text{goal} - \alpha_h \cdot d_\text{hand}
\]
where by default $\alpha_s=1000, \alpha_t=1, \alpha_h=0.1$.

\textbf{\textit{Termination.}} The episode terminates after 500 steps, or when $d_\text{goal} < 0.05$.

\subsubsection{\texttt{cabinets}}
\begin{center}
    \includegraphics[width=0.4\linewidth]{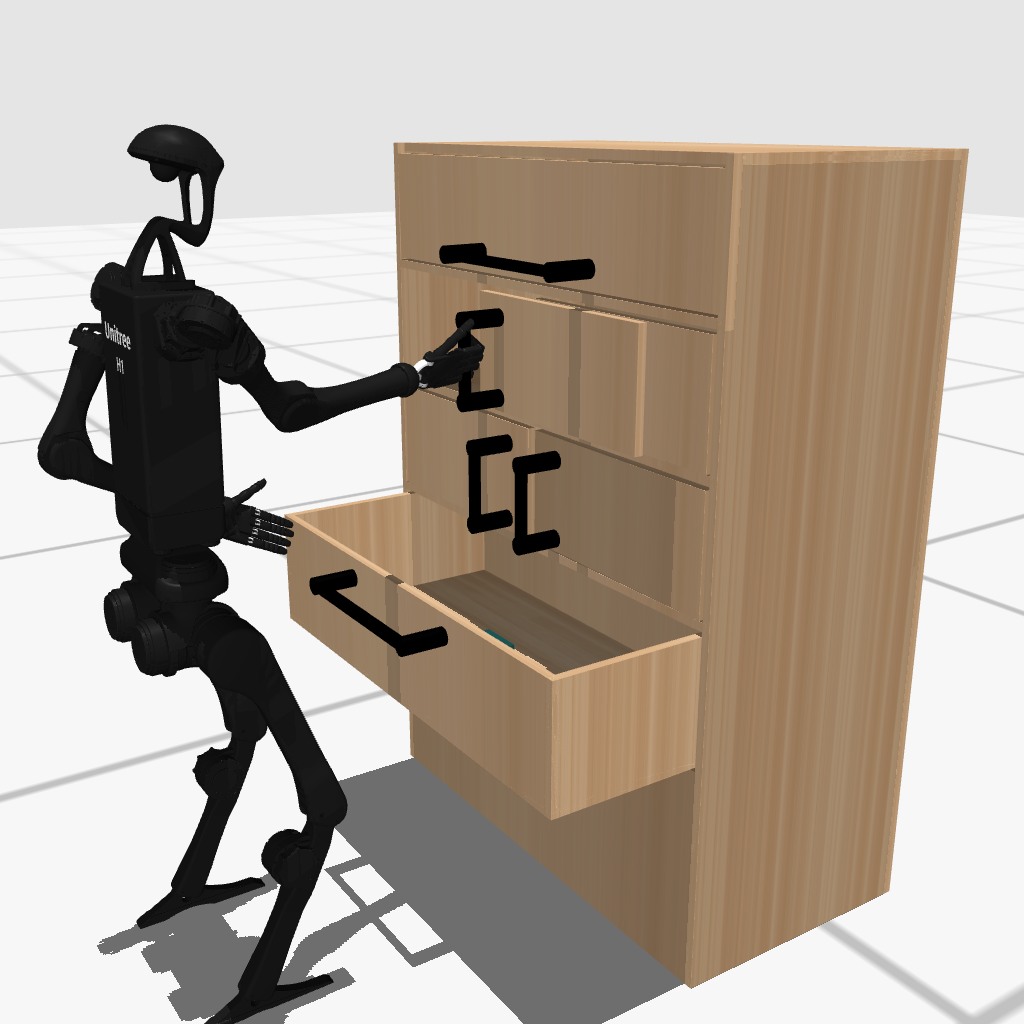}
\end{center}

\textbf{\textit{Objective.}} Open four different types of cabinet doors (e.g., hinge door, sliding door, drawer) and perform different manipulations (see subtasks below) for objects inside the cabinet.

\textbf{\textit{Observation.}} Positions and velocities of the robot's joints, the cabinets' joints, and the objects situated inside the cabinets.

\textbf{\textit{Initialization.}} The robot is initialized to a standing position, with random noise added to all joint positions during each episode reset.

\textbf{\textit{Reward Implementation.}} The reward of this task changes based on the occurring subtask.

Subtask 1 is to open the sliding door (second highest one), with reward
\[
    R_1 = 0.2 \cdot stable + 
    0.8 \cdot |l_\text{cabinet} / 0.4|
\]
where the cabinet joint position $l_\text{cabinet}$ is a value in $[0, 0.4]$.

Subtask 2 is to open the drawer (the lowest one), with reward
\[
    R_2 = 0.2 \cdot stable + 
    0.8 \cdot |l_\text{drawer} / 0.45|
\]
where the drawer joint position $l_\text{drawer}$ is a value in $[0, 0.45]$.

Subtask 3 is to put the cube from the drawer into the hinge-based cabinet (the third highest one). Both the left and right hinge-based cabinet doors' joint positions $\alpha_\text{door,left}, \alpha_\text{door,right}$ are values in $[0, 1.57]$. Let
\begin{itemize}
    \item $open_\text{door,left} = \min(1, |\alpha_\text{door,left}|)$
    \item $open_\text{door,right} = \min(1, |\alpha_\text{door,right}|)$
    \item $d_\text{destination,x} = tol(x_{cube} - 0.9, (-0.3, 0.3), 0.3)$
    \item $d_\text{destination,y} = tol(y_{cube}, (-0.6, 0.6), 0.3)$
    \item $d_\text{destination,z} = tol(z_{cube} - 0.94, (-0.15, 0.15), 0.3)$
    \item $r_\text{destination} = 0.3 \cdot mean(d_\text{destination,x}, d_\text{destination,y}) + 0.7 \cdot d_\text{destination,z}$
\end{itemize}
Then the reward of this subtask, $R_3$, is formulated as follows
\begin{align*}
    r_3 &= 0.5 \cdot \max(open_\text{door,left}, open_\text{door,right}) + 0.5 \cdot r_\text{destination} \\
    R_3 &= 0.2 \cdot stable + 0.8 \cdot r_3
\end{align*}

Subtask 4 is to put the original cube from the hinge-based left-right cabinet (third highest) into the pull-up cabinet (highest). The pull-up door has a joint position $\alpha_\text{pull}$ in $[0, 1.57]$. Let
\begin{itemize}
    \item $open_\text{pull} =  \min(1, |\alpha_\text{pull}|)$
    \item $d_\text{destination,x} = tol(x_{cube} - 0.9, (-0.3, 0.3), 0.3)$
    \item $d_\text{destination,y} = tol(y_{cube}, (-0.6, 0.6), 0.3)$
    \item $d_\text{destination,z} = tol(z_{cube} - 1.54, (-0.15, 0.15), 0.3)$
    \item $r_\text{destination} = 0.3 \cdot mean(d_\text{destination,x}, d_\text{destination,y}) + 0.7 \cdot d_\text{destination,z}$
\end{itemize}
Then the reward of this subtask, $R_4$, is formulated as follows
\begin{align*}
    r_4 = &0.5 \cdot open_\text{pull} + 0.5 \cdot r_\text{destination} \\
    R_4 = &0.2 \cdot stable + 0.8 \cdot r_4
\end{align*}

The reward during the occurrence of subtask $i$ is $R_i$. Upon the completion of subtask $i$, a sparse reward of $i * 100$ is offered. At the timestep where all subtasks are completed, a sparse reward of $1000$ is added to the total reward.

\textbf{\textit{Termination.}} The episode terminates after 1000 steps or whenever all subtasks are complete.

\subsubsection{\texttt{highbar}}
\begin{center}
    \includegraphics[width=0.4\linewidth]{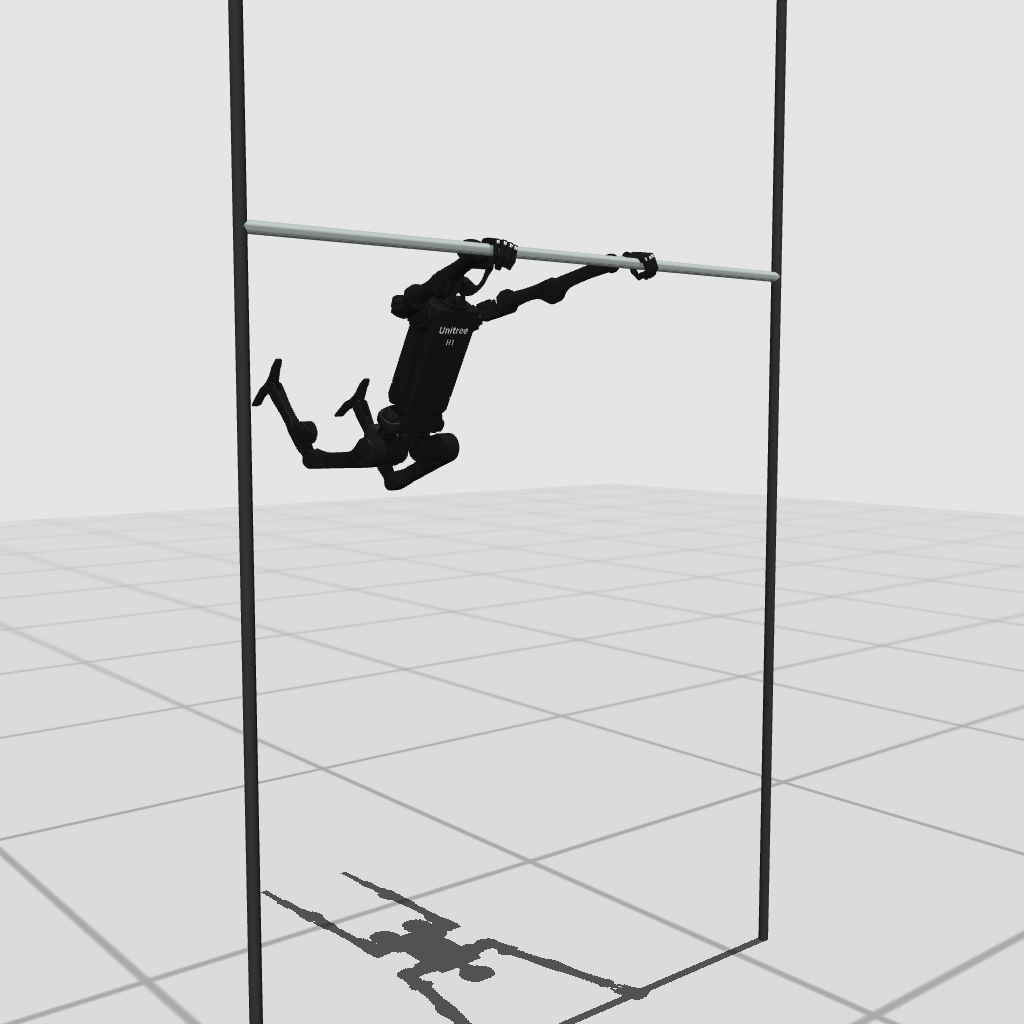}
\end{center}

\textbf{\textit{Objective.}} Athletically swing while staying attached to a horizontal high bar until reaching a vertical upside-down position.

\textbf{\textit{Observation.}} Joint positions and velocities of the robot.

\textbf{\textit{Initialization.}} The robot is initialized such that its hand is gripping the high bar, and its body is hanging from it.

\textbf{\textit{Reward Implementation.}} Let
\begin{itemize}
    \item $upright_{highbar} = upright((-\infty, -0.9), 1.9)$
    \item $feet = tol((z_\text{foot,left} + z_\text{foot,right}) / 2, (4.8, +\infty), 2)$
\end{itemize}
Then,
\[
    R(s, a) = upright_{highbar} \times feet \times e
\]

\textbf{\textit{Termination.}} The episode terminates after 1000 steps, or when $z_\text{head} < 2$.

\subsubsection{\texttt{door}}
\begin{center}
    \includegraphics[width=0.4\linewidth]{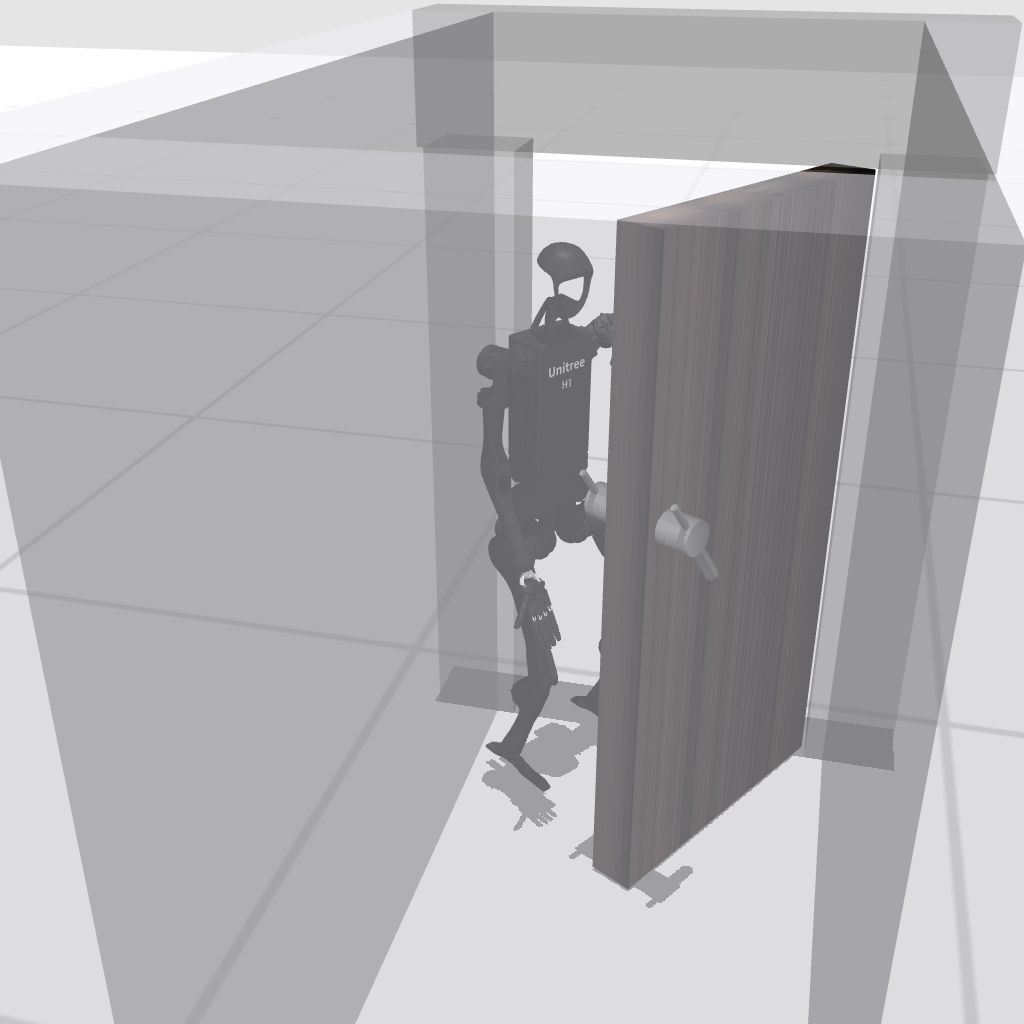}
\end{center}

\textbf{\textit{Objective.}} Pull a door open using its doorknob, and traverse through the doorpath while keeping the door open.

\textbf{\textit{Observation.}} Joint positions and velocities of the robot, door hinge, and door hatch.

\textbf{\textit{Initialization.}} The robot is initialized to a standing position. The door hinge joint and door latch are initialized such that the door is completely closed. For the current implementation, the door can only be pulled towards the robot, limiting its hinge joint position to be within $[0, 1.4]$. The door latch has a joint range of $[0, 2]$ (being pulled downwards until $2$ radians from positive x-axis, clockwise). Random noise is added to all joint positions during each episode reset.

\textbf{\textit{Reward Implementation.}} Let
\begin{itemize}
    \item $open_\text{door} = \min(1, q_\text{door}^2)$
    \item $open_\text{hatch} = tol(q_\text{hatch}, (0.75, 2), 0.75)$
    \item $proximity_\text{door} = $
    \begin{align*}
        tol(&\min(d(hand_\text{left}, door), d(hand_\text{right}, door)),\\
        &(0, 0.25), 1)
    \end{align*}
    \item $passage = tol(x_\text{IMU}, (1.2, +\infty), 1)$
\end{itemize}
Then, the reward of this task is:
\begin{align*}
    R &= 0.1 \cdot stable + 0.45 \cdot open_\text{door} + 0.05 \cdot open_\text{hatch} \\
    &+ 0.05 \cdot proximity_\text{door} + 0.35 \cdot passage
\end{align*}

\textbf{\textit{Termination.}} The episode terminates after 1000 steps, or when $z_\text{pelvis} < 0.58$.

\subsubsection{\texttt{truck}}
\begin{center}
    \includegraphics[width=0.4\linewidth]{fig/tasks/16_h1hand-truck_unload-v0-middle.png}
\end{center}

\textbf{\textit{Objective.}} Unload packages from a truck by moving them onto a platform.

\textbf{\textit{Observation.}} Joint positions and velocities of the robot, and positions and velocities of packages.

\textbf{\textit{Initialization.}} The robot is initialized to a standing position. Packages are initialized to be on the truck. Random noise is added to all joint positions during each episode reset.

\textbf{\textit{Reward Implementation.}} The reward of this task relies on the subsets of packages based on three categories: (1) being on truck ($p_\text{truck}$), (2) being picked up ($p_\text{picked}$), and (3) being on table ($p_\text{table}$). Let
\begin{itemize}
    \item $truck = tol(\min_{p \in p_\text{truck}} \| pos_{p} - pos_\text{pelvis} \|, (0, 0.2), 4)$
    \item $picked = tol(\min_{p \in p_\text{picked}} \| pos_{p} - pos_\text{pelvis} \|, (0, 0.2), 4)$
    \item $table = tol(\min_{p \in p_\text{table}} \| pos_{p} - pos_\text{table} \|, (0, 0.2), 4)$
    \item $r_{location} = 100 \cdot (p_{table} + p_{picked} - p_{truck})$
\end{itemize}
Then, the reward is:
\[
    R(s, a) = r_{location} + upright \times (1 + truck + picked + table)
\]
If all packages are picked up, an additional sparse reward of $1000$ is provided, and the episode terminates thereafter.

\textbf{\textit{Termination.}} The episode terminates after 1000 steps, or when all packages are delivered onto the table.

\subsubsection{\texttt{cube}}
\begin{center}
    \includegraphics[width=0.4\linewidth]{fig/tasks/17_h1hand-cubes-v0-solved.png}
\end{center}

\textbf{\textit{Objective.}} Manipulate two cubes, each cube in one hand, until they both correspond with a specific, randomly initialized target orientation.

\textbf{\textit{Observation.}} Joint positions and velocities of the robot and the two cubes to be manipulated on hand. The transparent cube in front of the robot is an indication of the target orientation for in-hand cubes, and its orientation is also in the state.

\textbf{\textit{Initialization.}} The robot is initialized to a standing position. The cubes are initialized at a random orientation right above the robot hands. Random noise is added to all joint positions during each episode reset.

\textbf{\textit{Reward Implementation.}} Let
\begin{itemize}
    \item $still_x = tol(v_x, (0, 0), 2)$
    \item $still_y = tol(v_y, (0, 0), 2)$
    \item $still = mean(still_x, still_y)$
    \item $quat_\text{target}$ denotes the target orientation for both in-hand cubes.
    \item $orientation=$ \[\frac{1}{2}[(quat_\text{cube,left} - quat_\text{target})^2 + (quat_\text{cube,right} - quat_\text{target})^2]\]
    \item $proximity_\text{cube} =$
    \begin{align*}
        \frac{1}{2}[&tol(d(cube_\text{left}, hand_\text{left}), (0,0), 0.5) \\
        &+ tol(d(cube_\text{right}, hand_\text{right}), (0,0), 0.5)]
    \end{align*}
\end{itemize}
Then the reward of this task is:
\begin{align*}
    R = &0.2 \cdot (stable \times still) \\
    &+ 0.5 \cdot orientation + 0.3 \cdot proximity_\text{cube}
\end{align*}

\textbf{\textit{Termination.}} The episode terminates after 500 steps, or when $z_\text{pelvis} < 0.5$, $z_\text{cube,left} < 0.5$, or $z_\text{cube,right} < 0.5$.

\subsubsection{\texttt{bookshelf}}
\begin{center}
    \includegraphics[width=0.4\linewidth]{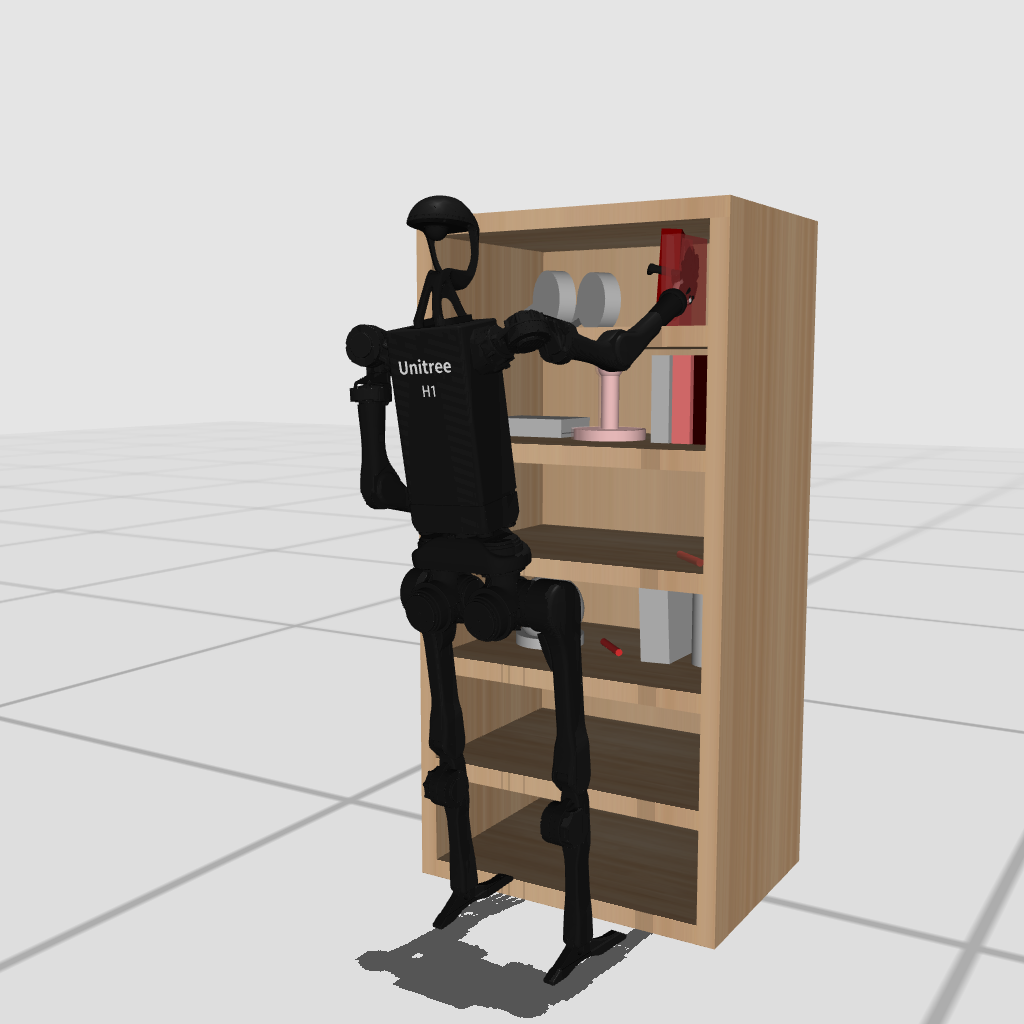}
\end{center}

\textbf{\textit{Objective.}} The \texttt{bookshelf} environment mainly concerns relocating five objects across the various shelves. There are five designated subtasks, resembling five different relocations (each for a different item and destination location). The items involved in the subtasks are colored from brightest to darkest in a red shade, where the brighter shade of red shows that the object's relocation is an earlier subtask to complete. The subtasks must be completed in order. The order is always the same in \texttt{bookshelf\_simple}, while it is randomized at every episode in \texttt{bookshelf\_hard}.

\textbf{\textit{Observation.}} Positions and velocities of the robot's joints and all objects on the bookshelf, including those that are not associated with any subtasks, as well as an index representing the next object to be relocated.

\textbf{\textit{Initialization.}} The robot is initialized to a standing position. All objects on the bookshelf are currently initialized at a default position. Random noise is added to all joint positions during each episode reset. In the simple version of this environment, the objects to move on the bookshelf and their destinations are fixed; in the hard version, both of them are randomized at the beginning of every episode instead.

\textbf{\textit{Reward Implementation.}} For each subtask $t_i$ where $i \in \{1, 2, \dots, 5\}$, its reward $r_i$ is formulated as follows. First, let
\begin{itemize}
    \item $proximity_\text{destination} =$
    \[tol(d(object, destination), (0, 0.15), 1)\]
    \item $d_\text{hand} = $
    \[\min(d(object,hand_\text{left}), d(object, hand_\text{right}))\]
    \item $proximity_\text{hand} = \exp(-d_\text{hand})$.
\end{itemize}
Then, the subtask reward is
\begin{align*}
    r_i = &0.4 \cdot proximity_\text{hand} \\
    &+ 0.2 \cdot stable + 0.4 \cdot proximity_\text{destination}
\end{align*}
and the reward for that step is the subtask reward per se. The current subtask is considered complete when the distance between the destination and object of current subtask is less than $0.15$.

An additional sparse reward of $100 * i$ is added to the timestep of subtask $i$'s completion.

\textbf{\textit{Termination.}} The episode terminates after 1000 steps, or when $z_\text{pelvis} < 0.58$, $z_\text{object} < 0.5$, or all subtasks succceeded.

\subsubsection{\texttt{basketball}}
\begin{center}
    \includegraphics[width=0.4\linewidth]{fig/tasks/19_h1hand-basketball-v0-solved.png}
\end{center}

\textbf{\textit{Objective.}} Catch a ball coming from random direction and throw it into the basket.

\textbf{\textit{Observation.}} Positions and velocities of the robot's joints and the basketball.

\textbf{\textit{Initialization.}} The robot is initialized to a standing position. The basketball is initialized such that it will be randomly spawned at a radius of \SI{1.5}{\meter} from the robot, at a random angle $\omega \in [-1.45\ \rm{rad}, 1.45\ \rm{rad}]$ from positive x-axis, and arrive right in front of the robot after \SI{0.2}{\second}. Random noise is added to all joint positions during each episode reset.

\textbf{\textit{Reward Implementation.}} The task is divided into two stages:
\[
    \begin{cases}
        \text{catch}: &\text{before the basketball collides with anything}\\
        \text{throw}: &\text{after the basketball experiences one collision}
    \end{cases}
\]
The rewards at different stages are formulated differently. Let
\begin{itemize}
    \item $proximity_\text{hand} = $
    \begin{align*}
         tol(&\max(d(ball,hand_\text{left}), d(ball,hand_\text{right})),\\
         &(0, 0.2), 1)
    \end{align*}
    \item $aim = tol(d(ball,basket), (0, 0), 7)$.
\end{itemize}
Then, the reward at each stage of the task is formulated as follows:

\begin{align*}
    R_\text{catch}(x, u) &= 0.5 \cdot proximity_\text{hand} +0.5 \cdot stable \\
    R_\text{throw}(x, u) &= 0.05 \cdot proximity_\text{hand} \\
    &+ 0.15 \cdot stable + 0.8 \cdot aim
\end{align*}

At the earliest timestep where $d(ball, basket) \leq 0.05$, the episode terminates, and a large sparse reward of $1000$ is given.

\textbf{\textit{Termination.}} The episode terminates after 500 steps, or when $z_\text{pelvis} < 0.5$, $z_\text{ball} < 0.5$, or when success has been achieved.

\subsubsection{\texttt{window}}
\begin{center}
    \includegraphics[width=0.4\linewidth]{fig/tasks/20_h1hand-window-v0-solved.png}
\end{center}

\textbf{\textit{Objective.}} Grab a window wiping tool and keep its tip parallel to a window by following a prescribed vertical speed (in absolute value). 

\textbf{\textit{Observation.}} Joint positions and velocities of the robot and the window wiping tool (position and velocities of the entire tool, in addition to those of the rotational joint attached between the wipe and the rod of the tool).

\textbf{\textit{Initialization.}} The robot is initialized to a standing position. The window wiping tool is initialized above the robot unit's hand in a parallel direction to the window frames. Random noise is added to all joint positions during each episode reset.

\textbf{\textit{Reward Implementation.}} Let
\begin{itemize}
    \item $proximity_\text{tool} =$
    \begin{align*}
        &\frac{1}{2}[tol(d(tool,hand_\text{left}), (0,0), 0.5) \\
        &+ tol(d(tool,hand_\text{right}), (0,0), 0.5)]
    \end{align*}
    \item $d_\text{window} = tol(d(head, window), (0.4, 0.4), 0.1)$
    \item $move_\text{wipe} = tol(|v_\text{wipe,z}|, (0.5, 0.5), 0.5)$
\end{itemize}
The manipulation reward is defined as:
\begin{align*}
    r_\text{manipulation} =
    &0.4 \cdot move_\text{wipe} + 0.4 \cdot proximity_\text{tool} \\
    &+ 0.2 \cdot (stable \times d_\text{window})
\end{align*}
Meanwhile, five sites are put on the four corners and center of the wipe to detect coverage of contact. Let these sites be $s_1, s_2, \dots, s_5$, then the reward for contact between the tool and window is defined as:
\begin{align*}
    r_\text{contact}
    &= \frac{1}{5} \sum_{s_i} tol(x_{s_i}, (0.92, 0.92), 0.4)
\end{align*}
Then, the reward of this task is:
\[
    R = 0.5 \cdot r_{\rm manipulation} + 0.5 \cdot r_{\rm contact}
\]

\textbf{\textit{Termination.}} The episode terminates after 1000 steps, or when $z_\text{pelvis} < 0.58$ or $z_\text{tool} < 0.58$.

\subsubsection{\texttt{spoon}}
\begin{center}
    \includegraphics[width=0.4\linewidth]{fig/tasks/21_h1hand-spoon-v0-middle.png}
\end{center}

\textbf{\textit{Objective.}} Grab a spoon and use it to follow a circular pattern inside a pot.

\textbf{\textit{Observation.}} Joint positions and velocities of the robot and the position and velocity of the spoon, as well as the target position that the spoon should be at the current timestep.

\textbf{\textit{Initialization.}} The robot is initialized to a standing position. The spoon is initialized at a specified position on the table, leftwards of the pot. Random noise is added to all joint positions during each episode reset.

\textbf{\textit{Reward Implementation.}} Let
\begin{itemize}
    \item $t:=$ current timestep
    \item $proximity_\text{tool} =$
    \begin{align*}
        &\frac{1}{2}[tol(d(tool,hand_\text{left}), (0,0), 0.5) \\
        &+ tol((tool,hand_\text{right}), (0,0), 0.5)]
    \end{align*}
    \item $destination = \begin{bmatrix}
            x_\text{pot} + 0.06 \cos(\frac{t \pi}{20}) \\
            y_\text{pot} + 0.06 \sin(\frac{t \pi}{20}) \\
            z_\text{pot}
        \end{bmatrix}$
    \item $r_\text{trajectory} = tol(d(spoon,destination), (0, 0), 0.15)$
    \item $r_\text{destination} = $
    \[
        \frac{1}{3} \sum_{i \in \{x, y, z\}} \mathbbm{1}_{i_\text{spoon} \text{in the pot}}
    \]
\end{itemize}
Then:
\begin{align*}
    R(s, a) &= 0.15 \cdot stable + 0.25 \cdot proximity_{tool} \\
    &+ 0.25 \cdot r_\text{destination} + 0.35 \cdot r_\text{trajectory}
\end{align*}

\textbf{\textit{Termination.}} The episode terminates after 1000 steps, or when $z_\text{pelvis} < 0.58$.

\subsubsection{\texttt{kitchen}}
\begin{center}
    \includegraphics[width=0.4\linewidth]{fig/tasks/22_h1hand-kitchen-v0-solved.png}
\end{center}

\textbf{\textit{Objective.}} Execute a sequence of actions in a kitchen environment, namely, open a microwave door, move a kettle, and turn burner and light switches.

\textbf{\textit{Observation.}} Joint positions and velocities of the robot, and positions and velocities of kitchenware.

\textbf{\textit{Initialization.}} The robot is initialized to a standing position. Kitchenware is initialized at specified positions. Random noise is added to all joint positions during each episode reset.

\textbf{\textit{Reward Implementation.}} A subtask in \texttt{kitchen} is considered complete if the distance between an object and its goal position is lower than a specified threshold. The sparse reward is the number of subtasks completed.

\textbf{\textit{Termination.}} The episode terminates after 500 steps.

\subsubsection{\texttt{package}}
\begin{center}
    \includegraphics[width=0.4\linewidth]{fig/tasks/23_h1hand-package-v0-middle.png}
\end{center}

\textbf{\textit{Objective.}} Move a box to a randomly initialized target position, also tested in the ablation in \Cref{fig:ablation_hierarchy}.

\textbf{\textit{Observation.}} Joint positions and velocities of the robot, both hand positions of the robot, package destination, package position, and package velocity.

\textbf{\textit{Initialization.}} The robot is initialized to a standing position. The package position and its destination are randomly initialized within a specific area. Random noise is added to all joint positions during each episode reset.

\textbf{\textit{Reward Implementation.}} Let
\begin{itemize}
    \item $height_\text{package} = \min(1, z_\text{package})$
    \item $success = \mathbbm{1}_{d(package, destination) < 0.1}$
    \item $d_\text{hand} = d(package, hand_\text{left}) + d(package, hand_\text{right})$.
\end{itemize}
Then, the reward is:
\begin{align*}
    R(s, a)
    = &- 3 \cdot d(package, destination) - 0.1 \cdot d_\text{hand} \\
    &+ stable + height_\text{package} + 1000 \cdot success
\end{align*}

\textbf{\textit{Termination.}} The episode terminates after 1000 steps, or when $d(package, destination) < 0.1$.

\subsubsection{\texttt{powerlift}}
\begin{center}
    \includegraphics[width=0.4\linewidth]{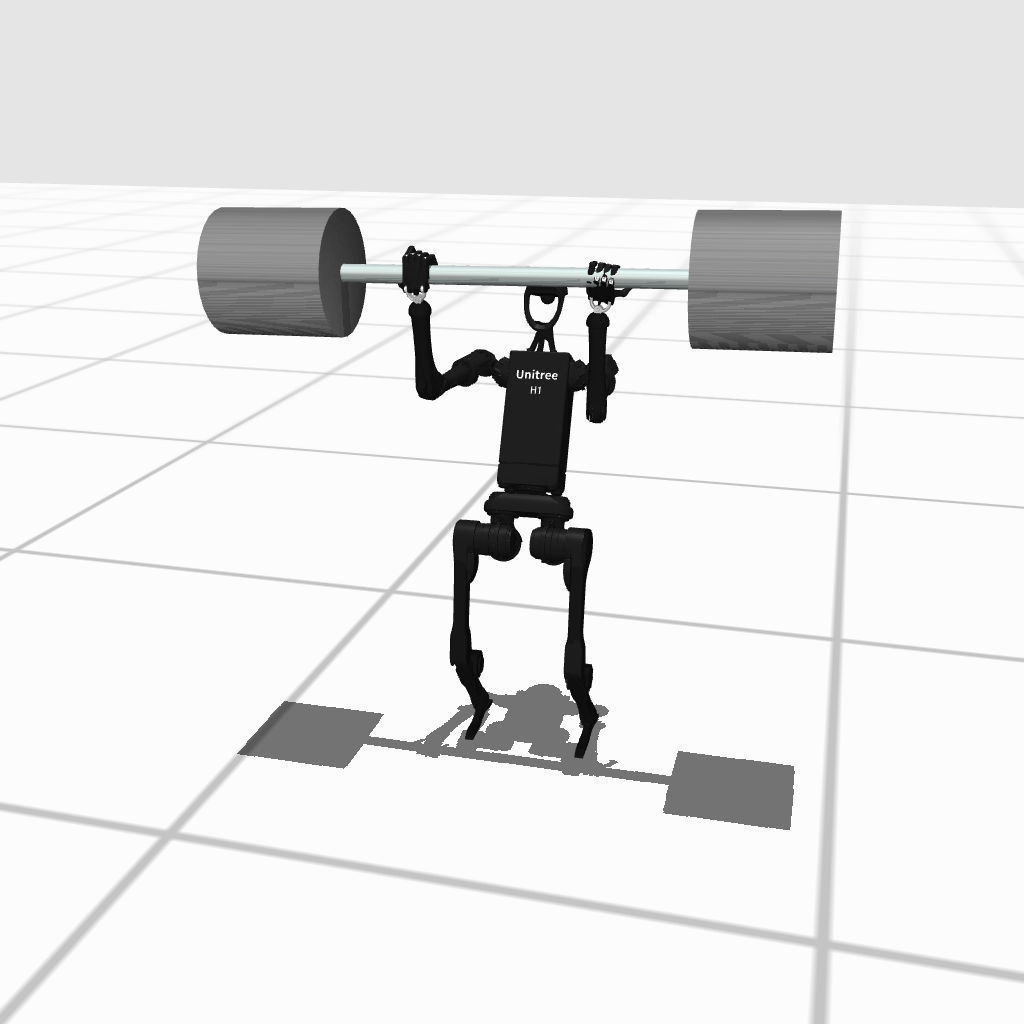}
\end{center}

\textbf{\textit{Objective.}} Lift a barbell of a designated mass.

\textbf{\textit{Observation.}} Joint positions and velocities of the robot and barbell.

\textbf{\textit{Initialization.}} The robot is initialized to a standing position. The barbell is initialized on the ground. Random noise is added to all joint positions during each episode reset.

\textbf{\textit{Reward Implementation.}} Let
\begin{itemize}
    \item $height_{barbell} = tol(z_{barbell}, (1.9, 2.1), 2)$.
\end{itemize}
Then,
\[
    R(s, a) = 0.2 \cdot stable + 0.8 \cdot height_{barbell}.
\]

\textbf{\textit{Termination.}} The episode terminates after 1000 steps, or when $z_\text{pelvis} < 0.2$.

\subsubsection{\texttt{room}}
\begin{center}
    \includegraphics[width=0.4\linewidth]{fig/tasks/25_h1hand-room-v0-middle.png}
\end{center}

\textbf{\textit{Objective.}} Organize a \SI{5}{\meter} by \SI{5}{\meter} space populated with randomly scattered object to minimize the variance of scattered objects' locations in $x$, $y$-axis directions.

\textbf{\textit{Observation.}} Joint positions and velocities of the robot and scattered objects.

\textbf{\textit{Initialization.}} The robot is initialized to a standing position. The scattered objects' positions are randomly initialized within a specific area. Random noise is added to all joint positions during each episode reset.

\textbf{\textit{Reward Implementation.}} Let
\begin{itemize}
    \item $X$ be a matrix containing the location of all scattered objects in 3D coordinates.
    \item $cleanness = tol(\max(Var(X_{:,0}), Var(X_{:,1})), (0, 0), 3)$, where $Var(X_{:,0})$ is the variance of x-coordinates of all objects' locations, and $Var(X_{:,1})$ is such variance for y-coordinates.
\end{itemize}
Then,
\[
    R(s, a) = 0.2 \cdot stable + 0.8 \cdot cleanness
\]

\textbf{\textit{Termination.}} The episode terminates after 1000 steps, or when $z_\text{pelvis} < 0.3$.

\subsubsection{\texttt{insert}}
\begin{center}
    \includegraphics[width=0.4\linewidth]{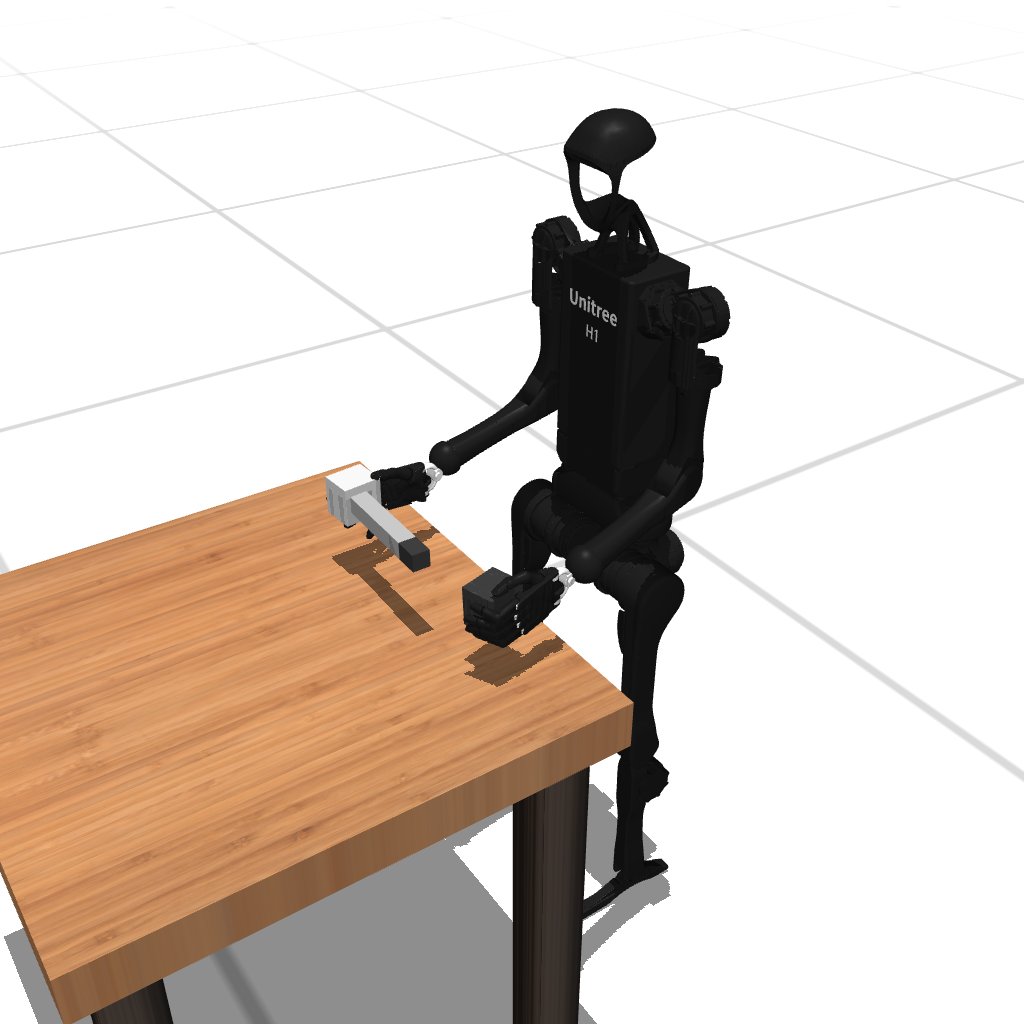}
\end{center}

\textbf{\textit{Objective.}} Insert the ends of a rectangular block into two small pegs. Two versions, \texttt{insert\_small} and \texttt{insert\_normal} present different object sizes.

\textbf{\textit{Observation.}} Joint positions and velocities of the robot, rectangular block to insert, and two provided small pegs.

\textbf{\textit{Initialization.}} The robot is initialized to a standing position. The rectangular block and two pegs are initialized on the table at specific orientation and position.

\textbf{\textit{Reward Implementation.}} Let the rectangular blocks be formualted to have two ends $end_a$, $end_b$, which must be respectively inserted to the pegs $peg_a$, $peg_b$.
\begin{itemize}
    \item $proximity_{peg, site} = tol(d(peg, site), (0, 0), 0.5)$
    \item $proximity_{block} =$ \[mean(proximity(peg_a, end_a), proximity(peg_b, end_b))\]
    \item $height(peg) = tol(z_{peg} - 1.1, (0, 0), 0.15)$
    \item $height_{pegs} = mean(height(peg_a), height(peg_b))$
    \item $proximity_{hands} =$
    \begin{align*}
        mean(&proximity(peg_a, hand_{left}), \\
            &proximity(peg_b, hand_{right}))
    \end{align*}
\end{itemize}
The reward of this task is phrased as:
\begin{align*}
    R(s, a) =
    &(0.5 \cdot stable + 0.5 \cdot proximity_{block}) \\
    &\times (0.5 \cdot height_{pegs} + 0.5 \cdot proximity_{hands})
\end{align*}

\textbf{\textit{Termination.}} The episode terminates after 1000 steps, or when any of the blocks or pegs are at a height lower than $0.5$ from floor.

\section{Training Details}
\label{sec:training_details}

\subsection{Baseline Implementation Details}
\label{sec:training_details:baselines}

For SAC, we use the implementation from JaxRL Minimal~\citep{jaxrl_minimal}. For PPO, we use the Stable-Baselines3 \citep{raffin2019stable} implementation with 4 parallel environments. For DreamerV3 and TD-MPC2, we use their official code. For DreamerV3, we use the `medium' configuration, with an update-to-data ratio of 64. For TD-MPC2, we use the 5M configuration. Unless specified, we use their default hyperparameters.


\subsection{Reaching Policy Implementation Details}
\label{sec:training_details:reaching_policy}
We train the low-level reaching policies described in \Cref{sec:benchmark:ablation_hierarchy} using PPO, largely parallelized on GPU using MuJoCo MJX. We employ the PureJaxRL \citep{lu2022discovered} implementation, using their default parameters, except a higher number of environments as detailed in \Cref{sec:benchmark:ablation_hierarchy}, 16 steps per environment, and an entropy coefficient of 0.001.

\subsection{Benchmarking Results}
\label{sec:training_details:results}

We summarize our benchmarking results in \Cref{tab:returns_mean} and \Cref{tab:returns_max}. We report the mean and standard deviation of maximum episode returns over three seeds. DreamerV3 and SAC are trained for $10$M, while TD-MPC2 is trained for $2$M environment steps, which roughly corresponds to \SI{48}{\hour}.

\begin{table}[ht]
    \centering
    \resizebox{\linewidth}{!}{
    \begin{tabular}{c|ccc|c}
        \toprule
          & DreamerV3 & TD-MPC2 & SAC & Target \\
         \midrule
\texttt{walk}& $   800.2 \pm    158.7$& $   782.0 \pm    109.2$& $    31.7 \pm     24.0$& $   700.0$    \\
\texttt{stand}& $   622.7 \pm    404.8$& $   809.0 \pm    137.1$& $   208.3 \pm    105.6$& $   800.0$    \\
\texttt{run}& $   633.8 \pm    222.4$& $    93.3 \pm     14.3$& $     5.0 \pm      2.1$& $   700.0$    \\
\texttt{reach}& $  7580.9 \pm   1951.0$& $  7316.1 \pm   2112.1$& $  4565.1 \pm    212.8$& $ 12000.0$    \\
\texttt{hurdle}& $   126.2 \pm     59.4$& $    46.4 \pm     10.8$& $    13.2 \pm      8.8$& $   700.0$    \\
\texttt{crawl}& $   878.8 \pm    122.7$& $   957.4 \pm     17.5$& $   330.0 \pm    111.9$& $   700.0$    \\
\texttt{maze}& $   272.3 \pm    116.6$& $   244.3 \pm     97.7$& $   144.8 \pm     17.8$& $  1200.0$    \\
\texttt{sit\_simple}& $   891.4 \pm     38.4$& $   411.1 \pm    368.0$& $   148.3 \pm    103.8$& $   750.0$    \\
\texttt{sit\_hard}& $   433.4 \pm    355.9$& $   343.0 \pm    381.7$& $    55.0 \pm     18.2$& $   750.0$    \\
\texttt{balance\_simple}& $    19.8 \pm      7.0$& $    40.5 \pm     23.9$& $    61.5 \pm      1.1$& $   800.0$    \\
\texttt{balance\_hard}& $    45.9 \pm     27.4$& $    48.2 \pm     28.5$& $    42.5 \pm     22.6$& $   800.0$    \\
\texttt{stair}& $   131.1 \pm     43.6$& $    70.4 \pm      7.1$& $    14.1 \pm      6.8$& $   700.0$    \\
\texttt{slide}& $   436.5 \pm    200.1$& $   119.0 \pm     35.9$& $     6.3 \pm      2.8$& $   700.0$    \\
\texttt{pole}& $   658.3 \pm    343.3$& $   226.3 \pm    116.1$& $    46.3 \pm     26.4$& $   700.0$    \\
\texttt{push}& $ -1251.9 \pm    659.8$& $  -258.7 \pm     66.5$& $   -97.9 \pm    147.0$& $   700.0$    \\
\texttt{cabinet}& $    57.3 \pm     66.3$& $   112.8 \pm    142.9$& $   211.8 \pm     33.8$& $  2500.0$    \\
\texttt{highbar}& $     8.9 \pm      5.8$& $     0.3 \pm      0.0$& $     9.4 \pm      3.7$& $   750.0$    \\
\texttt{door}& $   213.0 \pm    149.3$& $   274.7 \pm     12.5$& $    39.4 \pm     25.2$& $   600.0$    \\
\texttt{truck}& $  1103.8 \pm    232.9$& $  1132.6 \pm     72.1$& $  1077.5 \pm     95.0$& $  3000.0$    \\
\texttt{cube}& $   111.2 \pm     59.9$& $    54.7 \pm     33.1$& $   130.7 \pm     30.5$& $   370.0$    \\
\texttt{bookshelf\_simple}& $   840.4 \pm      5.6$& $   136.2 \pm     71.6$& $   346.9 \pm    231.5$& $  2000.0$    \\
\texttt{bookshelf\_hard}& $   530.2 \pm    302.5$& $    37.0 \pm      1.3$& $   293.9 \pm    121.6$& $  2000.0$    \\
\texttt{basketball}& $    19.3 \pm      2.5$& $    42.0 \pm     14.8$& $    22.1 \pm      3.2$& $  1200.0$    \\
\texttt{window}& $   461.0 \pm    252.8$& $    87.1 \pm     37.5$& $    62.9 \pm     83.8$& $   650.0$    \\
\texttt{spoon}& $   349.7 \pm     46.2$& $    77.9 \pm     80.6$& $    87.7 \pm     80.5$& $   650.0$    \\
\texttt{kitchen}& $     0.0 \pm      0.0$& $     0.0 \pm      0.0$& $     0.0 \pm      0.0$& $     4.0$    \\
\texttt{package}& $-18015.2 \pm   9477.7$& $ -3655.6 \pm   1055.0$& $ -6718.3 \pm    607.0$& $  1500.0$    \\
\texttt{powerlift}& $   315.9 \pm     16.9$& $    99.1 \pm     47.3$& $    81.8 \pm     46.7$& $   800.0$    \\
\texttt{room}& $   120.5 \pm     71.4$& $   131.4 \pm     56.7$& $    12.0 \pm      4.9$& $   400.0$    \\
\texttt{insert\_small}& $   184.8 \pm     26.3$& $   129.8 \pm     51.9$& $    10.8 \pm     13.4$& $   350.0$    \\
\texttt{insert\_normal}& $   171.5 \pm     33.2$& $   237.6 \pm      9.1$& $    46.3 \pm     63.1$& $   350.0$    \\
         \bottomrule
    \end{tabular}
    }
    \caption{\textbf{Average returns for HumanoidBench.} Each number represents average return@10M (return@2M) with the standard deviation for DreamerV3 and SAC (TD-MPC2).}
    \label{tab:returns_mean}
\end{table}

\begin{table}[ht]
    \centering
    \resizebox{\linewidth}{!}{
    \begin{tabular}{c|ccc|c}
        \toprule
          & DreamerV3 & TD-MPC2 & SAC & Target \\
         \midrule
\texttt{walk}& $   932.4 \pm      0.3$& $   900.3 \pm     47.6$& $    68.7 \pm     27.0$& $   700.0$    \\
\texttt{stand}& $   932.9 \pm      1.1$& $   925.7 \pm      2.5$& $   809.9 \pm    194.5$& $   800.0$    \\
\texttt{run}& $   895.9 \pm      6.0$& $   226.7 \pm     23.3$& $   104.8 \pm      7.4$& $   700.0$    \\
\texttt{reach}& $  9831.6 \pm    115.9$& $  9727.6 \pm     48.9$& $  7169.9 \pm    874.1$& $ 12000.0$    \\
\texttt{hurdle}& $   396.8 \pm     39.7$& $   196.7 \pm     30.5$& $    78.6 \pm     32.8$& $   700.0$    \\
\texttt{crawl}& $   985.3 \pm      0.6$& $   985.2 \pm      0.4$& $   626.5 \pm     29.1$& $   700.0$    \\
\texttt{maze}& $   592.5 \pm     49.0$& $   444.9 \pm     22.1$& $   269.9 \pm     37.8$& $  1200.0$    \\
\texttt{sit\_simple}& $   935.7 \pm      5.5$& $   928.4 \pm      1.5$& $   842.7 \pm     50.8$& $   750.0$    \\
\texttt{sit\_hard}& $   914.6 \pm      1.5$& $   906.3 \pm      6.0$& $   214.0 \pm     47.9$& $   750.0$    \\
\texttt{balance\_simple}& $    95.4 \pm      8.3$& $    95.3 \pm      3.1$& $    80.7 \pm      3.7$& $   800.0$    \\
\texttt{balance\_hard}& $   114.0 \pm     12.3$& $   122.2 \pm     13.1$& $    71.0 \pm      7.9$& $   800.0$    \\
\texttt{stair}& $   411.4 \pm      9.7$& $   251.9 \pm      9.1$& $    42.8 \pm      0.7$& $   700.0$    \\
\texttt{slide}& $   928.4 \pm      2.0$& $   311.9 \pm     15.1$& $    41.4 \pm      2.4$& $   700.0$    \\
\texttt{pole}& $   952.2 \pm     10.3$& $   644.9 \pm     21.2$& $   440.0 \pm     88.4$& $   700.0$    \\
\texttt{push}& $  1000.0 \pm      0.0$& $  1000.0 \pm      0.0$& $   352.8 \pm     31.5$& $   700.0$    \\
\texttt{cabinet}& $   722.6 \pm      7.3$& $   721.6 \pm     25.9$& $   485.9 \pm    137.2$& $  2500.0$    \\
\texttt{highbar}& $    83.1 \pm      4.6$& $     0.9 \pm      0.4$& $    40.8 \pm     41.7$& $   750.0$    \\
\texttt{door}& $   335.7 \pm     14.8$& $   310.6 \pm     10.6$& $   251.2 \pm      9.0$& $   600.0$    \\
\texttt{truck}& $  1674.3 \pm     52.6$& $  1457.2 \pm     24.3$& $  1387.5 \pm     10.0$& $  3000.0$    \\
\texttt{cube}& $   237.9 \pm      3.4$& $   241.1 \pm      1.2$& $   203.5 \pm     27.2$& $   370.0$    \\
\texttt{bookshelf\_simple}& $   849.6 \pm      0.3$& $   825.0 \pm      9.6$& $   766.5 \pm     10.3$& $  2000.0$    \\
\texttt{bookshelf\_hard}& $   867.8 \pm      8.4$& $   320.4 \pm     58.9$& $   681.5 \pm     10.3$& $  2000.0$    \\
\texttt{basketball}& $   808.8 \pm    340.5$& $  1055.3 \pm      4.1$& $   192.3 \pm     45.6$& $  1200.0$    \\
\texttt{window}& $   765.6 \pm     38.4$& $   201.1 \pm     91.5$& $   128.6 \pm    170.8$& $   650.0$    \\
\texttt{spoon}& $   421.5 \pm      2.5$& $   403.5 \pm      3.2$& $   297.5 \pm     26.5$& $   650.0$    \\
\texttt{kitchen}& $     0.0 \pm      0.0$& $     0.0 \pm      0.0$& $     0.0 \pm      0.0$& $     4.0$    \\
\texttt{package}& $  1009.2 \pm      4.1$& $  1003.3 \pm      3.4$& $ -3552.8 \pm    361.3$& $  1500.0$    \\
\texttt{powerlift}& $   338.6 \pm      0.1$& $   264.7 \pm     23.9$& $   171.2 \pm      3.6$& $   800.0$    \\
\texttt{room}& $   420.8 \pm     51.1$& $   353.3 \pm     41.5$& $    52.2 \pm      3.9$& $   400.0$    \\
\texttt{insert\_small}& $   239.8 \pm      4.6$& $   226.4 \pm     10.8$& $    72.7 \pm     21.9$& $   350.0$    \\
\texttt{insert\_normal}& $   279.9 \pm      9.9$& $   273.0 \pm      5.7$& $   135.3 \pm     51.5$& $   350.0$    \\
         \bottomrule
    \end{tabular}
    }
    \caption{\textbf{Maximum returns for HumanoidBench.} Each number represents maximum return@10M (return@2M) with the standard deviation for DreamerV3 and SAC (TD-MPC2).}
    \label{tab:returns_max}
\end{table}

\end{document}